%% file: fnet2.tex
\documentclass{article}

\input{preamble.tex}
\title{$\sutranets$\@: Sub-series Autoregressive Networks for
  Long-Sequence, Probabilistic Forecasting}

\author{%
Shane Bergsma \quad Timothy Zeyl \quad Lei Guo\\
Huawei Cloud, Alkaid Lab Canada\\
\texttt{\{shane.bergsma,timothy.zeyl,leiguo\}@huawei.com}
}

\begin{document}

\maketitle

\vspace{-2mm}
\begin{abstract}
\input{abstract.tex}
\end{abstract}

\section{Introduction}\label{sec:intro}

\input{intro.tex}

\section{Background and related work}\label{sec:background}

\input{related.tex}

\section{$\sutranets$: Sub-series autoregressive networks}\label{sec:method}

\input{method.tex}

\section{Experiments}\label{sec:experiments}

\input{experiments.tex}

\input{results.tex}

\section{Conclusion}\label{sec:conclusion}

\input{conclusion.tex}

\bibliographystyle{ACM-Reference-Format}
\bibliography{fnet2}

\include{supplemental.tex}

\end{document}

%% file: preamble.tex
\usepackage[utf8]{inputenc} % allow utf-8 input
\usepackage[T1]{fontenc}    % use 8-bit T1 fonts
\usepackage[hidelinks]{hyperref}       % hyperlinks
\usepackage{url}            % simple URL typesetting
\usepackage{booktabs}       % professional-quality tables
\usepackage{amsfonts}       % blackboard math symbols
\usepackage{nicefrac}       % compact symbols for 1/2, etc.
\usepackage{microtype}      % microtypography
\usepackage[table,xcdraw]{xcolor}
\usepackage{amsmath}
\usepackage[inline]{enumitem}
\usepackage{caption}
\usepackage{subcaption}
\usepackage{graphicx}
\usepackage{floatrow}
\usepackage{multirow}
\usepackage{soul}
\usepackage{tcolorbox}
\usepackage[normalem]{ulem}
\usepackage{mathdots}
\usepackage[multiple]{footmisc}

\usepackage[final]{neurips_2023}

\setcitestyle{numbers,square,comma}

\usepackage{float}
\floatstyle{plaintop}
\restylefloat{table}

\usepackage{xr}

\newcommand{\naive}{\mbox{Na\"{\i}ve}}
\newcommand{\seasonalnaive}{\mbox{SeasNa\"{\i}ve}}
\newcommand{\seasonalnaived}{\mbox{SeasNa\"{\i}ve1d}}
\newcommand{\seasonalnaivew}{\mbox{SeasNa\"{\i}ve1w}}
\newcommand{\deeparbinned}{\mbox{DeepAR-binned}}
\newcommand{\ctofar}{\mbox{C2FAR}}
\newcommand{\lags}{\mbox{C2FAR+lags}}
\newcommand{\dropout}{\mbox{C2FAR+dropout}}
\newcommand{\freqhier}{\mbox{Low2HighFreq}}
\newcommand{\regularprevs}{\mbox{Regular-alt}}
\newcommand{\regularnoprevs}{\mbox{Regular-non}}
\newcommand{\backfillprevs}{\mbox{Backfill-alt}}
\newcommand{\backfillnoprevs}{\mbox{Backfill-non}}
\newcommand{\backfillstandard}{\mbox{Backfill-standard}}
\newcommand{\sutranets}{\mbox{SutraNets}}
\newcommand{\sutranet}{\mbox{SutraNet}}
\newcommand{\hatbigt}{\tilde{T}}
\newcommand{\hatt}{\tilde{t}}
\newcommand{\hatn}{\tilde{N}}
\newcommand{\resourcesshrink}{0.75}

\newcommand{\flowchartshrink}{0.69}
\newcommand{\rolloutshrink}{0.69}
\newcommand{\shrinkimg}{0.96}
\newcommand{\shrinkfigthree}{0.35}
\newcommand{\shrinkfigthreeb}{0.34}
\newcommand{\shrinkfigtwo}{0.49}
\newcommand{\shrinkfigfour}{0.24}
\newcommand{\ltk}{<\! k}

\newcommand{\gtk}{>\! k}
\newcommand{\yhat}{\hat{y}}
\newcommand{\xvec}{\mathbf{x}}

\newcommand{\rnn}{\mbox{rnn}}
\newcommand{\elec}{\mbox{\emph{elec}}}
\newcommand{\traffic}{\mbox{\emph{traff}}}
\newcommand{\wiki}{\mbox{\emph{wiki}}}
\newcommand{\azure}{\mbox{\emph{azure}}}
\newcommand{\mnist}{\mbox{\emph{mnist}}}
\newcommand{\mnists}{\mbox{\emph{mnist$^\pi$}}}
\definecolor{lightlightgray}{gray}{0.9}
\newcommand{\hlc}[2][yellow]{{%
    \colorlet{foo}{#1}%
    \sethlcolor{foo}\hl{#2}}%
}
\newcounter{fcounter}
\newcommand\result[1]{
        \refstepcounter{fcounter}\vspace{2pt}
        \begin{tcolorbox}[boxsep=1pt,left=2pt,right=2pt,top=1pt,bottom=1pt]\noindent\emph{\textbf{Finding \arabic{fcounter}}: #1}
        \end{tcolorbox}\vspace{0pt}}

\input{tikz_figures/paper_tikz_preamble.tex}
\definecolor{lightlightgray}{gray}{0.9}
\tikzstyle{halo} = [rectangle, draw=none, text centered]
\tikzstyle{main2} = [rectangle, draw, text width=5cm, text height=5cm, fill=lightlightgray, fill opacity=.6]
\tikzstyle{gblock2} = [rectangle, draw, fill=lightlightgray, text centered, text width=3cm]

%% file: tikz_figures/paper_tikz_preamble.tex
\usepackage{tikz}
\usetikzlibrary{positioning,shapes,arrows,fit,arrows.meta,decorations.pathreplacing,calligraphy}
\usepackage{pgfplots}
\tikzstyle{rbox} = [circle, draw, inner sep=0pt, minimum width=0.5cm]
\tikzstyle{hbox} = [rbox, draw=none, fill=none]
\tikzstyle{tbox} = [rbox, line width=0.6mm]
\tikzstyle{fbox} = [tbox, dashed]
\tikzstyle{hfbox} = [hbox, fill=none]
\tikzstyle{algstep} = [draw=none, fill=none]

%% file: abstract.tex
We propose $\sutranets$, a novel method for neural probabilistic
forecasting of long-sequence time series.
$\sutranets$ use an autoregressive generative model to factorize the
likelihood of long sequences into products of conditional
probabilities.
When generating long sequences, most autoregressive approaches suffer
from harmful error accumulation, as well as challenges in modeling
long-distance dependencies.
$\sutranets$ treat long, univariate prediction as
\emph{multivariate} prediction over lower-frequency \emph{sub-series}.
Autoregression proceeds across time \emph{and} across sub-series in
order to ensure coherent multivariate (and, hence, high-frequency
univariate) outputs.  Since sub-series can be generated using fewer
steps, $\sutranets$ effectively reduce error accumulation and signal
path distances.
We find $\sutranets$ to significantly improve forecasting accuracy
over competitive alternatives on six real-world datasets, including
when we vary the number of sub-series and scale up the depth and width
of the underlying sequence models.

%% file: intro.tex
As a cloud provider, we rely on long-range forecasts to allocate
resources and plan capacity.
In particular, we need to predict the \emph{fine-grained} pattern of
demand (e.g., hourly) for a large number of products, over a long time
horizon (e.g., months).
Since purchase decisions are large-scale (e.g., supplying regions with
100K machines), and made weeks-to-months before delivery, cloud
providers have a profound need for methods that ``capture the
uncertainty of the future''~\cite{li2023eigen}.
Fine granularity, long histories, long forecast horizons, and
\emph{quantification of uncertainty} can be formalized as the long-sequence
(many-step) \emph{probabilistic} forecasting problem.

Recent work applies Transformers~\cite{vaswani2017attention} to
long-sequence forecasting.
In order to circumvent the Transformer's
quadratically-scaling \emph{attention} mechanism, long sequences
necessitate trading off increased signal path for reduced
computational
complexity~\cite{li2019enhancing,zhou2021informer,wu2021autoformer,liu2021pyraformer}.
Unfortunately, the effectiveness and even validity of these approaches
have been
questioned~\cite{zeng2022transformers,hewamalage2023forecast}.
Most of these approaches do not provide probabilistic outputs (rather
best-guess point predictions), and all condition on fixed-size
windows, typically of very short duration.  Indeed, many recent
papers on ``long-term''
forecasting~\cite{wu2021autoformer,wu2022timesnet,shabani2022scaleformer,zhou2022fedformer}
condition on a maximum of 96 inputs --- only 4 days of history at
1-hour granularity.
Such restricted context is a serious limitation given that many time
series (including in standard evaluation datasets based on electricity
and traffic) have strong \emph{weekly} seasonality.

Our production forecasting system, like many in industrial
settings~\cite{salinas2020deepar,mukherjee2018armdn,smyl2020hybrid}
and cloud services~\cite{aws2023deepar}, is based on RNNs.
The RNN's recurrent architecture forces the compression and
consolidation of long-range information, and RNNs work well on smaller
datasets as they are not as ``data hungry'' as
Transformers~\cite{didolkar2022temporal}.
RNNs can theoretically condition on context of any length; in this
paper, we condition RNN forecasts on length-2016 contexts without
difficulties in scaling.

RNNs are often used to generate probabilistic forecasts via an
autoregressive
approach~\cite{gasthaus2019probabilistic,salinas2020deepar,rabanser2020effectiveness,bergsma2022c2far},
whereby the probability over future timesteps is factorized into a
product of one-step-ahead
conditionals~\cite{bengio2003neural,graves2013generating}.
When generating many steps, however, two key problems arise:
\begin{enumerate*}[label=(\arabic*)]
\item a discrepancy grows between training, where we
condition on true previous values, versus inference, where we
condition on sampled values ({the \emph{discrepancy} problem}), and
\item long-term dependencies may
be challenging to exploit ({the \emph{signal path} problem}).
\end{enumerate*}
The discrepancy problem leads to error accumulation during
inference~\cite{bengio2015scheduled}, but also a kind
of \emph{informational asymmetry}: because highly-informative true
values are available during training, the model may ignore other
useful info, e.g., seasonally-lagged inputs provided as extra features.
The signal path problem, meanwhile, is especially acute for RNNs,
where distant information must be preserved over many steps in the RNN
hidden state.

\input{flowchart_fig.tex}

We propose $\sutranets$,\footnote{In Sanskrit, a \emph{s\={u}tra} is a
thread or ``that which like a thread runs through or holds everything
together''~\cite{monier1964sanskrit}.
A $\sutranet$ prediction is woven from threads of sub-series
predictions; Fig.~\ref{fig:rollouts} shows different stitching
patterns.}
a general method for probabilistic prediction of long time series. We
focus on the application of SutraNets to RNN-based forecasting, where
they offer major benefits.  $\sutranets$ transform a length-$N$ series
into $K$ sub-series of length $\nicefrac{N}{K}$.  Sub-series forecasts
are generated autoregressively, sequentially conditional on each
other, enabling coherent outputs (Fig.~\ref{fig:flowchart}).
$\sutranets$ address both the discrepancy problem (by forcing the
network to take larger \emph{generative strides}, i.e., predicting
without access to immediately-preceding true values) and the signal
path problem (by reducing the distance between historical and output
values by a factor of $K$).
Training of $\sutranets$ \emph{decomposes} over sub-series; i.e.,
sub-series RNNs can be trained in \emph{parallel}, enabling a $K$-fold
improvement in training parallelism over standard RNNs.

Experimentally, SutraNets demonstrate superior forecasting across six
datasets, improving accuracy by an average of 15\% over
$\ctofar$~\cite{bergsma2022c2far} and over strong seasonality-aware
baselines.  We evaluate a variety of other strategies for improving
long-sequence forecasting, including lags, input dropout, deeper and
wider networks, and multi-rate hierarchical approaches. Through these
evaluations, we gain insights into how to succeed at long-sequence
generation, in time series and beyond.

%% file: flowchart_fig.tex
\begin{figure}
  \centering
  \scalebox{\flowchartshrink}{
    {\input{tikz_figures/paper_sutranet_flowchart.3.12.tex}}
  }
  \mbox{}
  \vspace{-1mm}
  \mbox{}
  \caption{$\sutranets$: a long, univariate history is converted to
    lower-frequency sub-series. Multivariate autoregressive
    prediction, across time and sub-series, ensures coherent
    univariate output samples.\label{fig:flowchart}}
\end{figure}
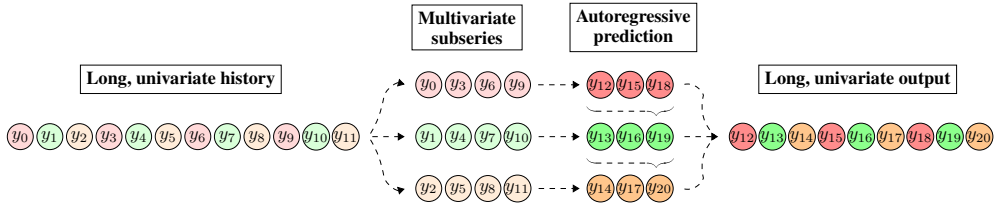

%% file: tikz_figures/paper_sutranet_flowchart.3.12.tex
\begin{tikzpicture}
\node [rbox, fill=red!15] (box0_0) {$y_{0}$};
\node [rbox, right of=box0_0, xshift=-4.3mm, fill=green!15] (box0_1) {$y_{1}$};
\node [rbox, right of=box0_1, xshift=-4.3mm, fill=orange!15] (box0_2) {$y_{2}$};
\node [rbox, right of=box0_2, xshift=-4.3mm, fill=red!15] (box0_3) {$y_{3}$};
\node [rbox, right of=box0_3, xshift=-4.3mm, fill=green!15] (box0_4) {$y_{4}$};
\node [rbox, right of=box0_4, xshift=-4.3mm, fill=orange!15] (box0_5) {$y_{5}$};
\node [rbox, right of=box0_5, xshift=-4.3mm, fill=red!15] (box0_6) {$y_{6}$};
\node [rbox, right of=box0_6, xshift=-4.3mm, fill=green!15] (box0_7) {$y_{7}$};
\node [rbox, right of=box0_7, xshift=-4.3mm, fill=orange!15] (box0_8) {$y_{8}$};
\node [rbox, right of=box0_8, xshift=-4.3mm, fill=red!15] (box0_9) {$y_{9}$};
\node [rbox, right of=box0_9, xshift=-4.3mm, fill=green!15] (box0_10) {$y_{10}$};
\node [rbox, right of=box0_10, xshift=-4.3mm, fill=orange!15] (box0_11) {$y_{11}$};
\node [draw, align=center, above of=box0_0, xshift=88, yshift=3] (ser_label) {\bf Long, univariate history};
\node [rbox, right of=box0_11, xshift=6mm, fill=green!15] (box0_10100) {$y_{1}$};
\node [rbox, right of=box0_10100, xshift=-4.3mm, fill=green!15] (box0_10101) {$y_{4}$};
\node [rbox, right of=box0_10101, xshift=-4.3mm, fill=green!15] (box0_10102) {$y_{7}$};
\node [rbox, right of=box0_10102, xshift=-4.3mm, fill=green!15] (box0_10103) {$y_{10}$};
\node [hbox, above of=box0_11, xshift=0mm] (box961_961) {\phantom{$y$}};
\node [hbox, below of=box0_11, xshift=0mm] (box962_961) {\phantom{$y$}};
\node [rbox, right of=box961_961, xshift=6mm, fill=red!15] (box0_10000) {$y_{0}$};
\node [rbox, right of=box0_10000, xshift=-4.3mm, fill=red!15] (box0_10001) {$y_{3}$};
\node [rbox, right of=box0_10001, xshift=-4.3mm, fill=red!15] (box0_10002) {$y_{6}$};
\node [rbox, right of=box0_10002, xshift=-4.3mm, fill=red!15] (box0_10003) {$y_{9}$};
\node [rbox, right of=box962_961, xshift=6mm, fill=orange!15] (box0_10200) {$y_{2}$};
\node [rbox, right of=box0_10200, xshift=-4.3mm, fill=orange!15] (box0_10201) {$y_{5}$};
\node [rbox, right of=box0_10201, xshift=-4.3mm, fill=orange!15] (box0_10202) {$y_{8}$};
\node [rbox, right of=box0_10202, xshift=-4.3mm, fill=orange!15] (box0_10203) {$y_{11}$};
\draw[-Triangle, dashed, shorten >= 1pt, shorten <= 2pt] ([xshift=5] box0_11.east) to[bend right=10] +(0.3cm,0.8cm) to[bend left=40] ([xshift=-5] box0_10000.west);
\draw[-Triangle, dashed, shorten >= 1pt, shorten <= 2pt] ([xshift=5] box0_11.east) -- ([xshift=-5] box0_10100.west);
\draw[-Triangle, dashed, shorten >= 1pt, shorten <= 2pt] ([xshift=5] box0_11.east) to[bend left=10] +(0.3cm,-0.8cm) to[bend right=40] ([xshift=-5] box0_10200.west);
\node [draw, align=center, above of=box0_10000, xshift=21, yshift=3] (ser_label) {\bf Multivariate \\ \bf subseries};
\node [rbox, right of=box0_10003, xshift=6mm, fill=red!45] (box0_20000) {$y_{12}$};
\node [rbox, right of=box0_20000, xshift=-4.3mm, fill=red!45] (box0_20001) {$y_{15}$};
\node [rbox, right of=box0_20001, xshift=-4.3mm, fill=red!45] (box0_20002) {$y_{18}$};
\node [rbox, right of=box0_10103, xshift=6mm, fill=green!45] (box0_20100) {$y_{13}$};
\node [rbox, right of=box0_20100, xshift=-4.3mm, fill=green!45] (box0_20101) {$y_{16}$};
\node [rbox, right of=box0_20101, xshift=-4.3mm, fill=green!45] (box0_20102) {$y_{19}$};
\node [rbox, right of=box0_10203, xshift=6mm, fill=orange!45] (box0_20200) {$y_{14}$};
\node [rbox, right of=box0_20200, xshift=-4.3mm, fill=orange!45] (box0_20201) {$y_{17}$};
\node [rbox, right of=box0_20201, xshift=-4.3mm, fill=orange!45] (box0_20202) {$y_{20}$};
\draw[-Triangle, dashed, shorten >= 4pt, shorten <= 4pt] (box0_10003) -- (box0_20000);
\draw[-Triangle, dashed, shorten >= 4pt, shorten <= 4pt] (box0_10103) -- (box0_20100);
\draw[-Triangle, dashed, shorten >= 4pt, shorten <= 4pt] (box0_10203) -- (box0_20200);
\draw[dashed, decorate, decoration={calligraphic brace,mirror,amplitude=5pt,aspect=0.8}] ([yshift=-12] box0_20000.west) --  ([yshift=-12] box0_20002.east);
\draw[dashed, decorate, decoration={calligraphic brace,mirror,amplitude=5pt,aspect=0.8}] ([yshift=-12] box0_20100.west) --  ([yshift=-12] box0_20102.east);
\node [draw, align=center, above of=box0_20000, xshift=18, yshift=3] (ser_label) {\bf Autoregressive \\ \bf prediction};
\node [rbox, right of=box0_20102, xshift=6mm, fill=red!45] (box0_0) {$y_{12}$};
\node [rbox, right of=box0_0, xshift=-4.3mm, fill=green!45] (box0_1) {$y_{13}$};
\node [rbox, right of=box0_1, xshift=-4.3mm, fill=orange!45] (box0_2) {$y_{14}$};
\node [rbox, right of=box0_2, xshift=-4.3mm, fill=red!45] (box0_3) {$y_{15}$};
\node [rbox, right of=box0_3, xshift=-4.3mm, fill=green!45] (box0_4) {$y_{16}$};
\node [rbox, right of=box0_4, xshift=-4.3mm, fill=orange!45] (box0_5) {$y_{17}$};
\node [rbox, right of=box0_5, xshift=-4.3mm, fill=red!45] (box0_6) {$y_{18}$};
\node [rbox, right of=box0_6, xshift=-4.3mm, fill=green!45] (box0_7) {$y_{19}$};
\node [rbox, right of=box0_7, xshift=-4.3mm, fill=orange!45] (box0_8) {$y_{20}$};
\draw[-, dashed, shorten >= 1pt, shorten <= 1pt] ([xshift=5] box0_20002.east) to[bend left=40] +(0.4cm,-0.2cm) to[bend right=30] ([xshift=-7] box0_0.west);
\draw[-Triangle, dashed, shorten >= 1pt, shorten <= 1pt] ([xshift=5] box0_20102.east) -- ([xshift=-5] box0_0.west);
\draw[-, dashed, shorten >= 1pt, shorten <= 1pt] ([xshift=5] box0_20202.east) to[bend right=40] +(0.4cm,0.2cm) to[bend left=30] ([xshift=-7] box0_0.west);
\node [draw, align=center, above of=box0_0, xshift=63, yshift=3] (ser_label) {\bf Long, univariate output};
\end{tikzpicture}

%% file: related.tex
\paragraph{Probabilistic autoregressive forecasting}
Autoregressive forecasting models typically combine a backbone
sequence model with a technique to estimate an output probability
distribution at each step.
In this paper, we apply SutraNets to
$\ctofar$~\cite{bergsma2022c2far}, a state-of-the-art distribution
estimator that was shown to improve over
DeepAR~\cite{salinas2020deepar},
$\deeparbinned$~\cite{rabanser2020effectiveness},
SQF-RNN~\cite{gasthaus2019probabilistic}, and
IQN-RNN~\cite{gouttes2021probabilistic}.
SutraNets can serve as the sequential backbone for any of these
estimators, as well as for other recent approaches such as the
denoising diffusion models~\cite{rasul2021autoregressive}, or
conditioned normalizing flows~\cite{rasul2020multi}.

\paragraph{Multi-rate forecasting}
It is sometimes acceptable to \emph{aggregate} high-frequency series
or event data to a lower frequency prior to sequence modeling, e.g.,
aggregating patient events over 6-month
intervals~\cite{lu2021recurrent}, or network traffic over 10
subframes~\cite{trinh2018mobile}.
When high-frequency predictions are required, lower-frequency
forecasts can provide guidance.  Techniques for \emph{temporal
  disaggregation} convert low-frequency series to higher frequencies
via related high-frequency \emph{indicator}
series~\cite{chow1971best,sax2013temporal}.
\citet{athanasopoulos2017forecasting} simultaneously forecast at
multiple frequencies and then reconcile the predictions.
Such methods do not provide uncertainty estimates nor are they based
on neural networks.

Scaleformer~\cite{shabani2022scaleformer} aggregates series, forecasts
at the lower frequency, and upsamples the lower-freq predictions in
order to provide additional inputs for the original forecast model.
We evaluate a similar approach, $\freqhier$ (\S\ref{subsec:freqhier}),
as a baseline.
Other neural approaches (in forecasting and beyond) implicitly
encourage learning of lower-frequency dynamics through hierarchical
attention~\cite{fan2019multi,liu2021pyraformer,didolkar2022temporal},
pooling
blocks~\cite{zhou2021informer,wu2021autoformer,madhusudhanan2021yformer,zhou2022fedformer},
and deeply-stacked~\cite{graves2013speech,van2016pixel} and
multi-rate~\cite{schmidhuber1992learning,hihi1995hierarchical,koutnik2014clockwork,sordoni2015hierarchical,chung2016hierarchical,chen2021time}
recurrence layers.
Unlike these approaches, $\sutranets$ \emph{explicitly} generate
\emph{low-frequency} sub-series conditional on other low-frequency
sub-series, without directly forecasting at a high frequency at all.

\paragraph{Reducing training/inference discrepancy}
If we know \emph{a~priori} how far to forecast, it is possible to
avoid error accumulation by directly forecasting all horizons in one
step~\cite{taieb2015bias}, or via non-autoregressive
decoding~\cite{fan2019multi}.  Such systems usually generate point
predictions~\cite{zhou2021informer,wu2021autoformer,challu2022n},
although predicting certain fixed forecast quantiles is also
possible~\cite{wen2017multi,fan2019multi}.
ProTran~\cite{tang2021probabilistic} is a probabilistic but
non-autoregressive approach.
Without modeling local dependencies among outputs, such models prevent
generation of \emph{coherent} samples, and preclude computing robust
likelihoods for \emph{given} output sequences --- where error
accumulation is not an issue (e.g., for anomaly detection, missing
value interpolation, compression, etc.).
$\sutranets$ are more in-step with flexible autoregressive models like
GPT3~\cite{brown2020language}.

For autoregressive models, \emph{scheduled
sampling}~\cite{bengio2015scheduled} aims to handle errors better
during inference by sampling (incorrect) inputs
during \emph{training}.  This approach was effective in tasks such as
human motion synthesis~\cite{li2017auto}, but results have been mixed
in autoregressive
forecasting~\cite{salinas2020deepar,sangiorgio2020robustness}.
Noting that scheduled sampling produces a biased
estimator, \citet{lamb2016professor} instead used adversarial
techniques to reduce training/inference discrepancy.
\citet{wu2020adversarial} applied this method to forecasting,
using a GAN~\cite{goodfellow2014generative} with an autoregressive
sparse transformer.
However, adversarial training showed insignificant gains over
the sparse transformer alone on 7-day electricity and traffic predictions.

Like scheduled sampling, \emph{dropout}~\cite{srivastava2014dropout}
is a kind of stochastic regularization that prevents overfitting by
injecting noise into networks.  Dropout rates of around 20\% are
commonly used in input layers of autoencoders and feed-forward
nets~\cite{baldi2013understanding,srivastava2014dropout}.  Dropout is
also common in autoregressive
models~\cite{vaswani2017attention,radford2018improving}, where, in
theory, it may encourage networks to rely less on (now noisy) previous
values~\cite{neill2018analysing}.

\paragraph{Modeling long-term dependencies}
When forecasting long series, seasonally \emph{lagged} values ---
historical values from, e.g., one year ago --- can be used as extra
features in the sequence model, acting as a kind of residual
input~\cite{he2016deep}.
A recent competition featured long time series (e.g., 700 days of
history)~\cite{anava2017web}; the winning model found lag to be more
effective than fixed attention weights~\cite{suilin2017how}. Some
researchers have even declared that, when it comes to attention, ``lag
is all you need''~\cite{faloutsos2019forecasting}.

Learned \emph{attention} can capture long-range
dependencies~\cite{vaswani2017attention}, but is infeasible for very
long sequences due to memory and time complexity quadratic in sequence
length.  Complexity also scales quadratically for forecasting via
fully-connected layers~\cite{oreshkin2019n,challu2022n}.  Recent work
has explored sparsified or hierarchical attention in
forecasting~\cite{li2019enhancing,zhou2021informer,liu2021pyraformer,wu2021autoformer}.
TimesNet~\cite{wu2022timesnet} transforms univariate time series into
2D tensors of multiple periods, which are then processed via
Inception-style 2D kernels to generate point predictions.
As mentioned earlier, most of these systems fail to condition on more
than very short windows --- a few days of hourly data.
Ultimately, we are interested in
\emph{probabilistic} forecasting using \emph{years} of fine-grained
historical data.
It is unclear whether lag is really all we need, or even whether
seasonal-naive baselines~\cite{hyndman2018forecasting} are already
more effective than Transformers~\cite{hewamalage2023forecast}.

Unlike Transformers, RNNs can be applied to long
sequences \emph{out-of-the-box}.
However, since RNN signal path is linear in sequence length,
mechanisms are required to preserve long-range info, such as the
special gated units of LSTMs~\cite{hochreiter1997long}.
Multi-dimensional RNNs~\cite{graves2008offline} propagate distant info
via row and column connections across
images~\cite{theis2015generative,van2016pixel}.
\citet{didolkar2022temporal} combine RNNs for long-range signals with
Transformers that attend over shorter chunks.
$\sutranets$, in contrast, do not require architectural
changes \emph{within} RNNs; indeed, they work with RNNs or
Transformers.  In $\sutranets$, low-frequency information is
both \emph{implicitly} captured in network state and \emph{explicitly}
and autoregressively communicated through generated low-frequency
forecasts.

\paragraph{Comparison to other RNNs}

\input{tab_prior_rnns.tex}

Table~\ref{tab:prior_rnns} compares SutraNets to prior RNNs (cf.\@
Table~1 in~\cite{vaswani2017attention}).
PatchTST~\cite{nie2022time} groups consecutive values into input
patches.  While patching was proposed as a method to reduce the
attentional complexity of Transformers, patching was also recently
used with RNNs in SegRNN~\cite{lin2023segrnn}, where it effectively
reduces the signal path by a factor of $K$ (taking in $K$ inputs each
step --- like SutraNets --- but with $K$ times fewer overall input
steps).
However, neither PatchTST nor SegRNN provides a mechanism to
probabilistically \emph{generate} patches, and thus generate coherent
\emph{probabilistic} outputs.
Dilated LSTMs~\cite{vezhnevets2017feudal} and Dilated
RNNs~\cite{chang2017dilated} with minimum dilation amounts ${>}1$ do
facilitate error reduction and parallel training, but at the cost of
sacrificing coherency, as the model is then ``equivalent to multiple
shared-weight networks, each working on partial
inputs''\cite[\S4.4]{chang2017dilated}.
With such DilatedRNNs, if the time series were to spike to a high
level at the very final conditioning input, only a subset of the
outputs would be generated conditional on this spike, resulting in
highly incoherent output.
This can be mitigated for point predictions by adding a
final \emph{fusion layer}~\cite{chang2017dilated}, but such an
approach is not compatible with probabilistic forecasts.

Another related approach is the Subscale WaveRNN model for audio
generation~\cite{kalchbrenner2018efficient} (conditioned on input
text).  For Subscale WaveRNN, the motivation is not to reduce signal
path nor improve error accumulation, but to have a $K$-fold increase
in inference parallelism when generating very, very long audio
sequences.
Generation proceeds somewhat similar to our regular, non-alternating
version (\S\ref{sec:method}) but conditioning on past and some
\emph{future} values, and with a shared neural network for all
sub-series.

%% file: tab_prior_rnns.tex
\begin{table}[]
\caption{Comparison of SutraNets to prior RNNs. Like prior methods,
  SutraNets reduce signal path, but also improve error accumulation
  (via a larger generative stride), while enabling greater training
  parallelism (via fewer sequential operations), without sacrificing
  coherent \emph{probabilistic} output.~\label{tab:prior_rnns}}
\footnotesize
\begin{tabular}{@{}cccccc@{}}
\toprule
\multirow{2}{*}{RNN method} & \multirow{2}{*}{Meaning of K}   & Signal & Generative & Sequential & Coherent \\
                    &                                          & path & max stride & operations & outputs \\ \midrule
Standard RNN, e.g., LSTM~\cite{hochreiter1997long} & N/A & $\mathcal{O}(N)$ & $\mathcal{O}(1)$ & $\mathcal{O}(N)$ & Yes \\
Skips~\cite{zhang2016architectural}, lags/residuals~\cite{he2016deep} & Skip/lag amount & $\mathcal{O}(N/K)$ & $\mathcal{O}(1)$ & $\mathcal{O}(N)$ & Yes \\
PatchTST~\cite{nie2022time}, SegRNN~\cite{lin2023segrnn} & Patch size/stride & $\mathcal{O}(N/K)$ & N/A & $\mathcal{O}(N/K)$ & N/A \\
DilatedRNN~\cite{chang2017dilated}, dilations $C{\ldots}K$ & Largest dilation & $\mathcal{O}(N/K)$ & $\mathcal{O}(C)$ & $\mathcal{O}(N/C)$ & If $C$=$1$ \\
SutraNets & Num.\@ sub-series & $\mathcal{O}(N/K)$ & $\mathcal{O}(K)$ & $\mathcal{O}(N/K)$ & Yes \\ \bottomrule
\end{tabular}
\end{table}

%% file: method.tex
Let $y_t \in \mathbb{R}$ be the value of a time series at time $t$,
and $\xvec_t$ be a vector of time-varying features or
\emph{covariates}.  Let $y_{i:j}$ denote sequence $y_i \ldots y_j$.
We seek a model of the distribution of $N$ future values of $y_t$ (the
\emph{prediction range}) given $T$ historical values (the
\emph{conditioning range}).  We can formulate this conditional
distribution as an autoregressive (AR) generative model as
in~\cite{graves2013generating}:
\begin{align}
p(y_{T+1:T+N}|y_{1:T}, \xvec_{1:T+N}) =& \prod_{t=T+1}^{T+N} p(y_t |y_{1:{t-1}}, \xvec_{1:{T+N}})
\label{eqn:standard}
\end{align}

Following DeepAR~\cite{salinas2020deepar}, forecasting approaches have
modeled the one-step-ahead distributions using
RNNs~\cite{gasthaus2019probabilistic,salinas2020deepar,rabanser2020effectiveness,bergsma2022c2far}
or Transformers~\cite{li2019enhancing}.  In these approaches, a global
sequence model is trained by slicing many training series into many
\emph{windows}, i.e., conditioning+prediction ranges at different
start points, and normalizing the windows using their conditioning
ranges.
Model parameters are fit via gradient descent, minimizing NLL of
prediction-range outputs.
At inference time, forecasts for a given conditioning range are
generated by sampling $\yhat_{T+1} \ldots \yhat_{T+N}$ sequentially,
autoregressively conditioning on samples generated at previous
timesteps.  By repeating this procedure many times, a Monte Carlo
estimate of Eq.~(\ref{eqn:standard}) is obtained, from which desired
forecast quantiles can be derived.

\input{rollouts_fig.tex}

$\sutranets$ consider each univariate window to be comprised of an
ordered collection of $K$ lower-frequency sub-series, each
$\nicefrac{1}{K}$ of the original length (Fig.~\ref{fig:flowchart}).
The $k$th sub-series is obtained by selecting every $K$th value in the
original window, each sub-series starting at a unique offset in
$1 \ldots K$.\footnote{Since the $k$th sub-series is assigned after
slicing training/testing windows, it will vary in terms of which
offset it corresponds to in the \emph{original} time series, e.g., it
may start at 1pm, 2pm, 3pm, etc., in different windows.\label{footnote:offsets}}
Let $y^k$ be the $k$th sub-series.  Let $\hatn=\nicefrac{N}{K}$ and
$\hatbigt=\nicefrac{T}{K}$ be prediction and conditioning lengths
corresponding to each sub-series, and let $\ltk$ and $\gtk$ denote
indices from $1 \ldots K$ that are less than and greater than $k$.
Through another application of the chain rule, now across sub-series,
Eq.~(\ref{eqn:standard}) becomes:
\begin{align}
p(y_{T+1:T+N}|y_{1:T}, \xvec_{1:T+N}) =& \prod_{\hatt=\hatbigt+1}^{\hatbigt+\hatn}
\prod_{k=1}^{K} p(y_{\hatt}^k | {\color{teal} y_{1:\hatt}^{<k} }, {\color{blue} y_{1:\hatt-1}^k }, {\color{violet} y_{1:\hatt-1}^{>k} }, \xvec_{1:{T+N}})
\label{eqn:sutranet}
\end{align}
where $\hatt$ iterates over timesteps in each sub-series.
Log-likelihoods of sub-series prediction ranges can be summed to
compute overall NLL, which is minimized during $\sutranet$ training.
Rather than modeling these conditionals using a single RNN,
$\sutranets$ use a separate RNN for each sub-series.  Each RNN has
its own parameters and hidden state, but $\sutranets$ are trained and
tuned collectively as a single network.
The $k$th RNN determines the $k$th conditional probability at each
timestep:
\begin{align}
p(y_{T+1:T+N}|y_{1:T}, \xvec_{1:T+N}) =& \prod_{\hatt=\hatbigt+1}^{\hatbigt+\hatn} \prod_{k=1}^{K} p(y_{\hatt}^k | \theta_{t}^k=f(h_{\hatt}^k))\label{eqn:sutranetrnn}
\end{align}
where $h_{\hatt}^k = \rnn^k(h_{\hatt-1}^k, {\color{teal} y_{\hatt}^{<k}}, {\color{blue} y_{\hatt-1}^k}, {\color{violet} y_{\hatt-1}^{>k}}, \xvec^k_{\hatt})$
and $f(\cdot)$ is a function mapping the $k$th RNN state to parameters
of a parametric output distribution,
$p(y_{\hatt}^{k}|\theta_t^k)$.\footnote{Note we can input any subset
of the covariates to the RNN at each timestep.  Here we include those
specific to the given timestep of the given sub-series,
$\xvec^k_{\hatt}$, as in practice most covariates are either static
across the prediction range (e.g., a product ID), or provide
timestep-specific info (e.g., the
hour-of-the-day)~\cite{salinas2020deepar}.}
So while prior models condition $y_t$ only on $y_{t-1}$, $\sutranets$
autoregressively condition the generation of $y_{\hatt}^k$ on:
\begin{itemize}[leftmargin=7em,noitemsep]
\item[1.~~${\color{teal} y_{\hatt}^{<k} }$:] \emph{Current} values (at $\hatt$) of sub-series with indices
$\ltk$ (always)
\item[2. ${\color{blue} y_{\hatt-1}^k }$:] The \emph{previous} value (at $\hatt-1$) of the current
sub-series, $k$ (always)
\item[3. ${\color{violet} y_{\hatt-1}^{>k} }$:] \emph{Previous} values (at
$\hatt-1$) of sub-series with indices $\gtk$ (optional)
\end{itemize}
Conditioning $y_{\hatt}^k$ on ${\color{teal} y_{\hatt}^{<k} }$ makes
the model autoregressive across sub-series; this ensures sub-series
are generated conditional on each other, and thus (recombined) samples
of the original high-frequency series will be coherent.  Conditioning
on ${\color{violet} y_{\hatt-1}^{>k}}$ is optional; when using
${\color{violet} y_{\hatt-1}^{>k}}$, $\sutranets$ must generate in an
\emph{alternating} manner, generating one value for each sub-series at
each timestep, $\hatt$ (Fig.~\ref{fig:rollouts:regularprevs}).
Without ${\color{violet} y_{\hatt-1}^{>k} }$, $\sutranets$ can
generate the complete prediction range of the $k$th sub-series before
generating any predictions for $\gtk$ ones
(Fig.~\ref{fig:rollouts:regularnoprevs}); in other words, we can
exchange ordering of the products in Eq.~(\ref{eqn:sutranet}), remove
${\color{violet} y_{\hatt-1}^{>k} }$ from the equation,
and proceed in a \emph{non-alternating} manner.

$\sutranets$ must also specify how the sub-series index, $k$, relates
to the offset in the original series.
Figs.~\ref{fig:rollouts:regularprevs}
and~\ref{fig:rollouts:regularnoprevs} illustrate \emph{regular}
ordering, where sub-series $k$ begins at offset $k$ in the original
series.  An alternative is
\emph{backfill} ordering, where sub-series $k$ starts at
offset $K$-$k$+$1$
(Figs.~\ref{fig:rollouts:backfillnoprevs},~\ref{fig:rollouts:backfillprevs}).
We must also specify the number of sub-series, $K$. We pick $K$ so it
divides into the primary seasonal period (e.g., $K$=$6$ for
24-hour seasons), as each sub-series then exhibits seasonality; signal
processing techniques can reveal seasonality when it is not
known \emph{a priori}.
Sub-series with seasonality may be more predictable, which will be
especially useful for non-alternating models, as these models generate
initial sub-series while conditioning on only $\nicefrac{1}{K}$ of
historical values.  Success for these models depends on whether there
is \emph{redundancy} in the series, or whether values from other
sub-series are sufficiently important that non-alternating generation
creates a \emph{missing information} problem.

\paragraph{Reducing training/inference discrepancy, signal path}

Non-alternating and backfill approaches force the network to predict
$K$ steps ahead, without access to true previous values.
Non-alternating also generates far horizons in fewer steps (e.g., for
the $k$=$1$ sub-series), reducing error accumulation.
In contrast, $\regularprevs$ generates in the same order, and
conditional on the same values, as standard RNN models
(Fig.~\ref{fig:rollouts:standard}), and so should suffer the same
effects from training/inference discrepancy.

All $\sutranets$ reduce RNN signal path by a factor of $K$, as can be
seen by tracing paths between features and outputs in
Fig.~\ref{fig:rollouts}.
Note $\regularprevs$ has low signal path but standard discrepancy (as
noted above).  The relative effectiveness of $\regularprevs$ can
thereby provide a diagnostic for whether discrepancy or signal path
affects a given forecasting task (\S\ref{sec:experiments}).

\paragraph{Form of output distribution and input encoding}

Unlike sampling text from a softmax, forecasting requires a parametric
output distribution, $p(y_{\hatt}^{k}|\theta_t^k)$
(Eq.~(\ref{eqn:sutranetrnn})), that can account for mixed discrete and
continuous outputs, potentially of unbounded dynamic range.
$\ctofar$~\cite{bergsma2022c2far} proposes such an output
distribution, via an autoregressive generative model over a
hierarchical coarse-to-fine discretization of time series amplitudes
(with Pareto-distributed tails).
We adopt $\ctofar$-LSTMs for our sub-series sequence models.  We use a
3-level C2F model with 12 bins at each level, discretizing normalized
values to a precision of $12^3$ but requiring only $12 \times 3$
output dimensions at each timestep.
Previous values from each sub-series and covariate features from other
sub-series are also C2F-encoded and provided as inputs at each step.
Here the efficiency of $\ctofar$ proves helpful; at each step we can
encode $K$ covariate values (i.e., from $K$ other sub-series) using
only $12 \times 3 \times K$ input dimensions, as opposed to, e.g., the
$12^3 \times K$ dimensions that would be required with equivalent flat
binnings~\cite{rabanser2020effectiveness}.

\paragraph{$\freqhier$}\label{subsec:freqhier}

Instead of generating all lower-frequency sub-series, we experiment
with generating a single low-freq sub-series and conditioning on it
within a high-freq model.  This $\freqhier$ approach
(Fig.~\ref{fig:rollouts:freqhier}) is similar to multi-rate approaches
(\S\ref{sec:background}), except here low-freq samples provide not
only features, but hard constraints on every corresponding $K$th value
in the high-freq output.  During inference, when we use Monte Carlo
sampling to estimate $p(y_{T+1:T+N}|\ldots)$, we can
actually \emph{sample} high-freq values \emph{conditional} on low-freq
ones.  As such, techniques such as rejection
sampling~\cite{murphy2012machine} are applicable; we experimented with
importance sampling via likelihood weighting (fixing the low-freq
values and sampling the rest), but ultimately found the likelihood
weights had such high variance that it was more effective to use
uniform weights and accept inconsistent estimation.

\paragraph{Complexity}

Let $H$ be the number of RNN hidden units and $K$ the number of
sub-series.  The complexity-per-timestep of each $\sutranet$ RNN
scales with the sum of the RNN's recurrence, $H \times H$, and
projection of (at most) $K$ inputs (from other sub-series) to the RNN
hidden state, $K \times H$.  Even though $\sutranets$ have $K$
separate RNNs, only one runs at each original timestep (i.e., each
$\sutranet$ RNN runs $\nicefrac{1}{K}$ of total timesteps).  In
consequence, $\sutranet$ complexity-per-timestep scales with $H \times
H + K \times H$, whereas standard RNNs scale $H \times H$.  In
practice, $H >> K$, and we find empirically $\sutranets$ also scale
$H \times H$, resulting in similar training/inference speeds and
memory requirements compared to standard models, given the same
individual LSTM depth and width (\S\ref{sec:experiments}).

One key advantage of $\sutranets$ is that training \emph{decomposes}
over sub-series.
Sub-series RNNs do not condition on hidden states from other
sub-series RNNs, rather only on \emph{generated} values.
During training, when we have access to the true values of other
sub-series, all inputs are known in advance and sub-series RNNs can be
trained in \emph{parallel}.
The only sequential action is evolution of each RNN's hidden state ---
over $\nicefrac{1}{K}$ fewer steps than standard or multidimensional
RNNs~\cite{theis2015generative,van2016pixel}.  In this way,
$\sutranet$-RNNs provide middle-ground between RNNs and standard
Transformers (with no sequential computation in training).  Inference
in $\sutranets$, like all AR models, is
sequential~\cite{katharopoulos2020transformers}.

\paragraph{Application to Transformers}

Rather than using $K$ RNNs to parameterize the conditional
probabilities as in Eq.~(\ref{eqn:sutranetrnn}), we could instead use
$K$ autoregressive Transformers. Compared to a standard Transformer,
backfill and non-alternating sub-series Transformers would have the
advantage of reducing \emph{error accumulation} (by increasing the
generative stride), but not of \emph{signal path} (already
$\mathcal{O}(1)$ for
Transformers~\cite{vaswani2017attention}). Moreover, at each timestep,
a sub-series Transformer could attend to essentially all prior values,
limited only by the generative order. In a naive implementation,
SutraNet Transformers would therefore have an asymptotic complexity of
$\mathcal{O}(N^2)$ --- the same as standard Transformers. However, we
can attain $\mathcal{O}(N^2/K)$ complexity by restricting each
sub-series Transformer to only attend to values from its own
sub-series, plus a few proximal values from other sub-series (similar
to strided or banded
self-attention~\cite{child2019generating,brown2020language}). Although
asymptotically larger than the
$\mathcal{O}(N{(\log N)}^2)$ of LogSparse
attention~\cite{li2019enhancing}, it merits empirical investigation to
determine which approach achieves the most favorable accuracy vs.\@
complexity trade-off.

\paragraph{Limitations and broader impact}\label{subsec:limitations}

\input{limitations.tex}

%% file: rollouts_fig.tex
\setul{}{1.6pt}
\begin{figure}
  \centering
      {\makebox[\textwidth][c]{
          \begin{subfigure}{\shrinkfigthree\textwidth}
            \centering
            \scalebox{\rolloutshrink}{
              {\input{tikz_figures/paper_hf_multiv.vanilla_c2far.3.9.tex}}
            }
            \mbox{}
            \vspace{-4mm}
            \mbox{}
            \subcaption{Standard RNN, e.g., C2FAR\label{fig:rollouts:standard}}
          \end{subfigure}
      \hspace{-3mm}
          \begin{subfigure}{\shrinkfigthree\textwidth}
            \centering
            \scalebox{\rolloutshrink}{
              {\input{tikz_figures/paper_hf_freqhier.3.9.tex}}
            }
            \mbox{}
            \vspace{-4mm}
            \mbox{}
            \subcaption{$\freqhier$\label{fig:rollouts:freqhier}}
          \end{subfigure}
      \hspace{-3mm}
          \begin{subfigure}{\shrinkfigthree\textwidth}
            \centering
            \scalebox{\rolloutshrink}{
              {\input{tikz_figures/paper_hf_multiv.regular.interleaved.3.9.tex}}
            }
            \mbox{}
            \vspace{-4mm}
            \mbox{}
            \subcaption{$\regularprevs$ $\sutranet$\label{fig:rollouts:regularprevs}}
          \end{subfigure}
      }}
      \mbox{}
      \vspace{2mm}
      \mbox{}
      {\makebox[\textwidth][c]{
          \begin{subfigure}{\shrinkfigthree\textwidth}
            \centering
            \scalebox{\rolloutshrink}{
              {\input{tikz_figures/paper_hf_multiv.regular.wall2wall.3.9.tex}}
            }
            \mbox{}
            \vspace{-4mm}
            \mbox{}
            \subcaption{$\regularnoprevs$ $\sutranet$\label{fig:rollouts:regularnoprevs}}
          \end{subfigure}
      \hspace{-3mm}
          \begin{subfigure}{\shrinkfigthree\textwidth}
            \centering
            \scalebox{\rolloutshrink}{
              {\input{tikz_figures/paper_hf_multiv.backfill.wall2wall.3.9.tex}}
            }
            \mbox{}
            \vspace{-4mm}
            \mbox{}
            \subcaption{$\backfillnoprevs$ $\sutranet$\label{fig:rollouts:backfillnoprevs}}
          \end{subfigure}
      \hspace{-3mm}
          \begin{subfigure}{\shrinkfigthree\textwidth}
            \centering
            \scalebox{\rolloutshrink}{
              {\input{tikz_figures/paper_hf_multiv.backfill.interleaved.3.9.tex}}
            }
            \mbox{}
            \vspace{-4mm}
            \mbox{}
            \subcaption{$\backfillprevs$ $\sutranet$\label{fig:rollouts:backfillprevs}}
          \end{subfigure}
      }}
      \mbox{}
      \vspace{-1mm}
      \mbox{}
      \caption{Ordering of generative steps in $\sutranets$. One
        output value (\ul{bold border}) is generated in each row,
        conditional on (1)~feature nodes (\dashuline{dashed border})
        \emph{in that row}, and on (2)~state from previous steps
        ($\Longrightarrow$ arrow connections).  $\sutranet$ RNNs
        (\subref{fig:rollouts:regularprevs}-\subref{fig:rollouts:backfillprevs})
        use features generated in earlier steps by both themselves and
        other RNNs; these values can even be from future timesteps
        (backfill cases).\label{fig:rollouts}}
\end{figure}

%% file: tikz_figures/paper_hf_multiv.vanilla_c2far.3.9.tex
\begin{tikzpicture}
\draw[fill=gray!7,draw=none] (-0.9900000000000001cm,-0.35) rectangle ++(6.6,0.7);
\node [algstep,xshift=-0.81cm,yshift=-0.0cm] {\footnotesize 0};
\draw[fill=gray!22,draw=none] (-0.9900000000000001cm,-1.0499999999999998) rectangle ++(6.6,0.7);
\node [algstep,xshift=-0.81cm,yshift=-0.7cm] {\footnotesize 1};
\draw[fill=gray!7,draw=none] (-0.9900000000000001cm,-1.75) rectangle ++(6.6,0.7);
\node [algstep,xshift=-0.81cm,yshift=-1.4cm] {\footnotesize 2};
\draw[fill=gray!22,draw=none] (-0.9900000000000001cm,-2.4499999999999997) rectangle ++(6.6,0.7);
\node [algstep,xshift=-0.81cm,yshift=-2.0999999999999996cm] {\footnotesize 3};
\draw[fill=gray!7,draw=none] (-0.9900000000000001cm,-3.15) rectangle ++(6.6,0.7);
\node [algstep,xshift=-0.81cm,yshift=-2.8cm] {\footnotesize 4};
\draw[fill=gray!22,draw=none] (-0.9900000000000001cm,-3.85) rectangle ++(6.6,0.7);
\node [algstep,xshift=-0.81cm,yshift=-3.5cm] {\footnotesize 5};
\draw[fill=gray!7,draw=none] (-0.9900000000000001cm,-4.549999999999999) rectangle ++(6.6,0.7);
\node [algstep,xshift=-0.81cm,yshift=-4.199999999999999cm] {\footnotesize 6};
\draw[fill=gray!22,draw=none] (-0.9900000000000001cm,-5.249999999999999) rectangle ++(6.6,0.7);
\node [algstep,xshift=-0.81cm,yshift=-4.8999999999999995cm] {\footnotesize 7};
\draw[fill=gray!7,draw=none] (-0.9900000000000001cm,-5.949999999999999) rectangle ++(6.6,0.7);
\node [algstep,xshift=-0.81cm,yshift=-5.6cm] {\footnotesize 8};
\node [tbox, fill=red!45] (box0_0) {$y_{0}$};
\node [hbox, right of=box0_0, xshift=-3.5mm] (box0_1) {\phantom{$y$}};
\node [hbox, right of=box0_1, xshift=-3.5mm] (box0_2) {\phantom{$y$}};
\node [hbox, right of=box0_2, xshift=-3.5mm] (box0_3) {\phantom{$y$}};
\node [hbox, right of=box0_3, xshift=-3.5mm] (box0_4) {\phantom{$y$}};
\node [hbox, right of=box0_4, xshift=-3.5mm] (box0_5) {\phantom{$y$}};
\node [hbox, right of=box0_5, xshift=-3.5mm] (box0_6) {\phantom{$y$}};
\node [hbox, right of=box0_6, xshift=-3.5mm] (box0_7) {\phantom{$y$}};
\node [hbox, right of=box0_7, xshift=-3.5mm] (box0_8) {\phantom{$y$}};
\node [fbox, yshift=-0.7cm, fill=red!45] (box1_0) {$y_{0}$};
\node [tbox, right of=box1_0, xshift=-3.5mm, fill=red!45] (box1_1) {$y_{1}$};
\node [hbox, right of=box1_1, xshift=-3.5mm] (box1_2) {\phantom{$y$}};
\node [hbox, right of=box1_2, xshift=-3.5mm] (box1_3) {\phantom{$y$}};
\node [hbox, right of=box1_3, xshift=-3.5mm] (box1_4) {\phantom{$y$}};
\node [hbox, right of=box1_4, xshift=-3.5mm] (box1_5) {\phantom{$y$}};
\node [hbox, right of=box1_5, xshift=-3.5mm] (box1_6) {\phantom{$y$}};
\node [hbox, right of=box1_6, xshift=-3.5mm] (box1_7) {\phantom{$y$}};
\node [hbox, right of=box1_7, xshift=-3.5mm] (box1_8) {\phantom{$y$}};
\node [rbox, yshift=-1.4cm, fill=red!15] (box2_0) {$y_{0}$};
\node [fbox, right of=box2_0, xshift=-3.5mm, fill=red!45] (box2_1) {$y_{1}$};
\node [tbox, right of=box2_1, xshift=-3.5mm, fill=red!45] (box2_2) {$y_{2}$};
\node [hbox, right of=box2_2, xshift=-3.5mm] (box2_3) {\phantom{$y$}};
\node [hbox, right of=box2_3, xshift=-3.5mm] (box2_4) {\phantom{$y$}};
\node [hbox, right of=box2_4, xshift=-3.5mm] (box2_5) {\phantom{$y$}};
\node [hbox, right of=box2_5, xshift=-3.5mm] (box2_6) {\phantom{$y$}};
\node [hbox, right of=box2_6, xshift=-3.5mm] (box2_7) {\phantom{$y$}};
\node [hbox, right of=box2_7, xshift=-3.5mm] (box2_8) {\phantom{$y$}};
\node [rbox, yshift=-2.0999999999999996cm, fill=red!15] (box3_0) {$y_{0}$};
\node [rbox, right of=box3_0, xshift=-3.5mm, fill=red!15] (box3_1) {$y_{1}$};
\node [fbox, right of=box3_1, xshift=-3.5mm, fill=red!45] (box3_2) {$y_{2}$};
\node [tbox, right of=box3_2, xshift=-3.5mm, fill=red!45] (box3_3) {$y_{3}$};
\node [hbox, right of=box3_3, xshift=-3.5mm] (box3_4) {\phantom{$y$}};
\node [hbox, right of=box3_4, xshift=-3.5mm] (box3_5) {\phantom{$y$}};
\node [hbox, right of=box3_5, xshift=-3.5mm] (box3_6) {\phantom{$y$}};
\node [hbox, right of=box3_6, xshift=-3.5mm] (box3_7) {\phantom{$y$}};
\node [hbox, right of=box3_7, xshift=-3.5mm] (box3_8) {\phantom{$y$}};
\node [rbox, yshift=-2.8cm, fill=red!15] (box4_0) {$y_{0}$};
\node [rbox, right of=box4_0, xshift=-3.5mm, fill=red!15] (box4_1) {$y_{1}$};
\node [rbox, right of=box4_1, xshift=-3.5mm, fill=red!15] (box4_2) {$y_{2}$};
\node [fbox, right of=box4_2, xshift=-3.5mm, fill=red!45] (box4_3) {$y_{3}$};
\node [tbox, right of=box4_3, xshift=-3.5mm, fill=red!45] (box4_4) {$y_{4}$};
\node [hbox, right of=box4_4, xshift=-3.5mm] (box4_5) {\phantom{$y$}};
\node [hbox, right of=box4_5, xshift=-3.5mm] (box4_6) {\phantom{$y$}};
\node [hbox, right of=box4_6, xshift=-3.5mm] (box4_7) {\phantom{$y$}};
\node [hbox, right of=box4_7, xshift=-3.5mm] (box4_8) {\phantom{$y$}};
\node [rbox, yshift=-3.5cm, fill=red!15] (box5_0) {$y_{0}$};
\node [rbox, right of=box5_0, xshift=-3.5mm, fill=red!15] (box5_1) {$y_{1}$};
\node [rbox, right of=box5_1, xshift=-3.5mm, fill=red!15] (box5_2) {$y_{2}$};
\node [rbox, right of=box5_2, xshift=-3.5mm, fill=red!15] (box5_3) {$y_{3}$};
\node [fbox, right of=box5_3, xshift=-3.5mm, fill=red!45] (box5_4) {$y_{4}$};
\node [tbox, right of=box5_4, xshift=-3.5mm, fill=red!45] (box5_5) {$y_{5}$};
\node [hbox, right of=box5_5, xshift=-3.5mm] (box5_6) {\phantom{$y$}};
\node [hbox, right of=box5_6, xshift=-3.5mm] (box5_7) {\phantom{$y$}};
\node [hbox, right of=box5_7, xshift=-3.5mm] (box5_8) {\phantom{$y$}};
\node [rbox, yshift=-4.199999999999999cm, fill=red!15] (box6_0) {$y_{0}$};
\node [rbox, right of=box6_0, xshift=-3.5mm, fill=red!15] (box6_1) {$y_{1}$};
\node [rbox, right of=box6_1, xshift=-3.5mm, fill=red!15] (box6_2) {$y_{2}$};
\node [rbox, right of=box6_2, xshift=-3.5mm, fill=red!15] (box6_3) {$y_{3}$};
\node [rbox, right of=box6_3, xshift=-3.5mm, fill=red!15] (box6_4) {$y_{4}$};
\node [fbox, right of=box6_4, xshift=-3.5mm, fill=red!45] (box6_5) {$y_{5}$};
\node [tbox, right of=box6_5, xshift=-3.5mm, fill=red!45] (box6_6) {$y_{6}$};
\node [hbox, right of=box6_6, xshift=-3.5mm] (box6_7) {\phantom{$y$}};
\node [hbox, right of=box6_7, xshift=-3.5mm] (box6_8) {\phantom{$y$}};
\node [rbox, yshift=-4.8999999999999995cm, fill=red!15] (box7_0) {$y_{0}$};
\node [rbox, right of=box7_0, xshift=-3.5mm, fill=red!15] (box7_1) {$y_{1}$};
\node [rbox, right of=box7_1, xshift=-3.5mm, fill=red!15] (box7_2) {$y_{2}$};
\node [rbox, right of=box7_2, xshift=-3.5mm, fill=red!15] (box7_3) {$y_{3}$};
\node [rbox, right of=box7_3, xshift=-3.5mm, fill=red!15] (box7_4) {$y_{4}$};
\node [rbox, right of=box7_4, xshift=-3.5mm, fill=red!15] (box7_5) {$y_{5}$};
\node [fbox, right of=box7_5, xshift=-3.5mm, fill=red!45] (box7_6) {$y_{6}$};
\node [tbox, right of=box7_6, xshift=-3.5mm, fill=red!45] (box7_7) {$y_{7}$};
\node [hbox, right of=box7_7, xshift=-3.5mm] (box7_8) {\phantom{$y$}};
\node [rbox, yshift=-5.6cm, fill=red!15] (box8_0) {$y_{0}$};
\node [rbox, right of=box8_0, xshift=-3.5mm, fill=red!15] (box8_1) {$y_{1}$};
\node [rbox, right of=box8_1, xshift=-3.5mm, fill=red!15] (box8_2) {$y_{2}$};
\node [rbox, right of=box8_2, xshift=-3.5mm, fill=red!15] (box8_3) {$y_{3}$};
\node [rbox, right of=box8_3, xshift=-3.5mm, fill=red!15] (box8_4) {$y_{4}$};
\node [rbox, right of=box8_4, xshift=-3.5mm, fill=red!15] (box8_5) {$y_{5}$};
\node [rbox, right of=box8_5, xshift=-3.5mm, fill=red!15] (box8_6) {$y_{6}$};
\node [fbox, right of=box8_6, xshift=-3.5mm, fill=red!45] (box8_7) {$y_{7}$};
\node [tbox, right of=box8_7, xshift=-3.5mm, fill=red!45] (box8_8) {$y_{8}$};
\draw[red, bend left,->, line width=0.9mm,opacity=0.5]  (box0_0) to node [auto] {} (box1_1);
\draw[red, bend left,->, line width=0.9mm,opacity=0.5]  (box1_1) to node [auto] {} (box2_2);
\draw[red, bend left,->, line width=0.9mm,opacity=0.5]  (box2_2) to node [auto] {} (box3_3);
\draw[red, bend left,->, line width=0.9mm,opacity=0.5]  (box3_3) to node [auto] {} (box4_4);
\draw[red, bend left,->, line width=0.9mm,opacity=0.5]  (box4_4) to node [auto] {} (box5_5);
\draw[red, bend left,->, line width=0.9mm,opacity=0.5]  (box5_5) to node [auto] {} (box6_6);
\draw[red, bend left,->, line width=0.9mm,opacity=0.5]  (box6_6) to node [auto] {} (box7_7);
\draw[red, bend left,->, line width=0.9mm,opacity=0.5]  (box7_7) to node [auto] {} (box8_8);
\matrix[draw,thick,below left,fill=white,inner sep=1.5pt] at ([xshift=-0.18000000000000002cm,yshift=-3pt]current bounding box.north east) {
  \draw[->,bend left, line width=0.9mm,opacity=0.5,color=red] (0,0) -- ++ (0.35,0); \node[right,xshift=0.3cm]{\textsc{rnn}$_0$};\\
};
\end{tikzpicture}

%% file: tikz_figures/paper_hf_freqhier.3.9.tex
\begin{tikzpicture}
\draw[fill=gray!7,draw=none] (-0.9900000000000001cm,-0.35) rectangle ++(6.6,0.7);
\node [algstep,xshift=-0.81cm,yshift=-0.0cm] {\footnotesize 0};
\draw[fill=gray!22,draw=none] (-0.9900000000000001cm,-1.0499999999999998) rectangle ++(6.6,0.7);
\node [algstep,xshift=-0.81cm,yshift=-0.7cm] {\footnotesize 1};
\draw[fill=gray!7,draw=none] (-0.9900000000000001cm,-1.75) rectangle ++(6.6,0.7);
\node [algstep,xshift=-0.81cm,yshift=-1.4cm] {\footnotesize 2};
\draw[fill=gray!22,draw=none] (-0.9900000000000001cm,-2.4499999999999997) rectangle ++(6.6,0.7);
\node [algstep,xshift=-0.81cm,yshift=-2.0999999999999996cm] {\footnotesize 3};
\draw[fill=gray!7,draw=none] (-0.9900000000000001cm,-3.15) rectangle ++(6.6,0.7);
\node [algstep,xshift=-0.81cm,yshift=-2.8cm] {\footnotesize 4};
\draw[fill=gray!22,draw=none] (-0.9900000000000001cm,-3.85) rectangle ++(6.6,0.7);
\node [algstep,xshift=-0.81cm,yshift=-3.5cm] {\footnotesize 5};
\draw[fill=gray!7,draw=none] (-0.9900000000000001cm,-4.549999999999999) rectangle ++(6.6,0.7);
\node [algstep,xshift=-0.81cm,yshift=-4.199999999999999cm] {\footnotesize 6};
\draw[fill=gray!22,draw=none] (-0.9900000000000001cm,-5.249999999999999) rectangle ++(6.6,0.7);
\node [algstep,xshift=-0.81cm,yshift=-4.8999999999999995cm] {\footnotesize 7};
\draw[fill=gray!7,draw=none] (-0.9900000000000001cm,-5.949999999999999) rectangle ++(6.6,0.7);
\node [algstep,xshift=-0.81cm,yshift=-5.6cm] {\footnotesize 8};
\node [hbox] (box0_0) {\phantom{$y$}};
\node [hbox, right of=box0_0, xshift=-4mm] (box0_1) {\phantom{$y$}};
\node [tbox, right of=box0_1, xshift=-4mm, fill=orange!45] (box0_2) {$y_{2}$};
\node [hbox, right of=box0_2, xshift=-2.0mm] (box0_3) {\phantom{$y$}};
\node [hbox, right of=box0_3, xshift=-4mm] (box0_4) {\phantom{$y$}};
\node [hbox, right of=box0_4, xshift=-4mm] (box0_5) {\phantom{$y$}};
\node [hbox, right of=box0_5, xshift=-2.0mm] (box0_6) {\phantom{$y$}};
\node [hbox, right of=box0_6, xshift=-4mm] (box0_7) {\phantom{$y$}};
\node [hbox, right of=box0_7, xshift=-4mm] (box0_8) {\phantom{$y$}};
\node [hbox, yshift=-0.7cm] (box1_0) {\phantom{$y$}};
\node [hbox, right of=box1_0, xshift=-4mm] (box1_1) {\phantom{$y$}};
\node [fbox, right of=box1_1, xshift=-4mm, fill=orange!45] (box1_2) {$y_{2}$};
\node [hbox, right of=box1_2, xshift=-2.0mm] (box1_3) {\phantom{$y$}};
\node [hbox, right of=box1_3, xshift=-4mm] (box1_4) {\phantom{$y$}};
\node [tbox, right of=box1_4, xshift=-4mm, fill=orange!45] (box1_5) {$y_{5}$};
\node [hbox, right of=box1_5, xshift=-2.0mm] (box1_6) {\phantom{$y$}};
\node [hbox, right of=box1_6, xshift=-4mm] (box1_7) {\phantom{$y$}};
\node [hbox, right of=box1_7, xshift=-4mm] (box1_8) {\phantom{$y$}};
\node [hbox, yshift=-1.4cm] (box2_0) {\phantom{$y$}};
\node [hbox, right of=box2_0, xshift=-4mm] (box2_1) {\phantom{$y$}};
\node [rbox, right of=box2_1, xshift=-4mm, fill=orange!15] (box2_2) {$y_{2}$};
\node [hbox, right of=box2_2, xshift=-2.0mm] (box2_3) {\phantom{$y$}};
\node [hbox, right of=box2_3, xshift=-4mm] (box2_4) {\phantom{$y$}};
\node [fbox, right of=box2_4, xshift=-4mm, fill=orange!45] (box2_5) {$y_{5}$};
\node [hbox, right of=box2_5, xshift=-2.0mm] (box2_6) {\phantom{$y$}};
\node [hbox, right of=box2_6, xshift=-4mm] (box2_7) {\phantom{$y$}};
\node [tbox, right of=box2_7, xshift=-4mm, fill=orange!45] (box2_8) {$y_{8}$};
\node [tbox, yshift=-2.0999999999999996cm, fill=red!45] (box3_0) {$y_{0}$};
\node [hbox, right of=box3_0, xshift=-4mm] (box3_1) {\phantom{$y$}};
\node [fbox, right of=box3_1, xshift=-4mm, fill=orange!45] (box3_2) {$y_{2}$};
\node [hbox, right of=box3_2, xshift=-2.0mm] (box3_3) {\phantom{$y$}};
\node [hbox, right of=box3_3, xshift=-4mm] (box3_4) {\phantom{$y$}};
\node [rbox, right of=box3_4, xshift=-4mm, fill=orange!15] (box3_5) {$y_{5}$};
\node [hbox, right of=box3_5, xshift=-2.0mm] (box3_6) {\phantom{$y$}};
\node [hbox, right of=box3_6, xshift=-4mm] (box3_7) {\phantom{$y$}};
\node [rbox, right of=box3_7, xshift=-4mm, fill=orange!15] (box3_8) {$y_{8}$};
\node [fbox, yshift=-2.8cm, fill=red!45] (box4_0) {$y_{0}$};
\node [tbox, right of=box4_0, xshift=-4mm, fill=red!45] (box4_1) {$y_{1}$};
\node [fbox, right of=box4_1, xshift=-4mm, fill=orange!45] (box4_2) {$y_{2}$};
\node [hbox, right of=box4_2, xshift=-2.0mm] (box4_3) {\phantom{$y$}};
\node [hbox, right of=box4_3, xshift=-4mm] (box4_4) {\phantom{$y$}};
\node [rbox, right of=box4_4, xshift=-4mm, fill=orange!15] (box4_5) {$y_{5}$};
\node [hbox, right of=box4_5, xshift=-2.0mm] (box4_6) {\phantom{$y$}};
\node [hbox, right of=box4_6, xshift=-4mm] (box4_7) {\phantom{$y$}};
\node [rbox, right of=box4_7, xshift=-4mm, fill=orange!15] (box4_8) {$y_{8}$};
\node [rbox, yshift=-3.5cm, fill=red!15] (box5_0) {$y_{0}$};
\node [fbox, right of=box5_0, xshift=-4mm, fill=red!45] (box5_1) {$y_{1}$};
\node [tbox, right of=box5_1, xshift=-4mm, fill=orange!45] (box5_2) {$y_{2}$};
\node [hbox, right of=box5_2, xshift=-2.0mm] (box5_3) {\phantom{$y$}};
\node [hbox, right of=box5_3, xshift=-4mm] (box5_4) {\phantom{$y$}};
\node [rbox, right of=box5_4, xshift=-4mm, fill=orange!15] (box5_5) {$y_{5}$};
\node [hbox, right of=box5_5, xshift=-2.0mm] (box5_6) {\phantom{$y$}};
\node [hbox, right of=box5_6, xshift=-4mm] (box5_7) {\phantom{$y$}};
\node [rbox, right of=box5_7, xshift=-4mm, fill=orange!15] (box5_8) {$y_{8}$};
\node [rbox, yshift=-4.199999999999999cm, fill=red!15] (box6_0) {$y_{0}$};
\node [rbox, right of=box6_0, xshift=-4mm, fill=red!15] (box6_1) {$y_{1}$};
\node [fbox, right of=box6_1, xshift=-4mm, fill=orange!45] (box6_2) {$y_{2}$};
\node [tbox, right of=box6_2, xshift=-2.0mm, fill=red!45] (box6_3) {$y_{3}$};
\node [hbox, right of=box6_3, xshift=-4mm] (box6_4) {\phantom{$y$}};
\node [fbox, right of=box6_4, xshift=-4mm, fill=orange!45] (box6_5) {$y_{5}$};
\node [hbox, right of=box6_5, xshift=-2.0mm] (box6_6) {\phantom{$y$}};
\node [hbox, right of=box6_6, xshift=-4mm] (box6_7) {\phantom{$y$}};
\node [rbox, right of=box6_7, xshift=-4mm, fill=orange!15] (box6_8) {$y_{8}$};
\node [rbox, yshift=-4.8999999999999995cm, fill=red!15] (box7_0) {$y_{0}$};
\node [rbox, right of=box7_0, xshift=-4mm, fill=red!15] (box7_1) {$y_{1}$};
\node [rbox, right of=box7_1, xshift=-4mm, fill=orange!15] (box7_2) {$y_{2}$};
\node [fbox, right of=box7_2, xshift=-2.0mm, fill=red!45] (box7_3) {$y_{3}$};
\node [tbox, right of=box7_3, xshift=-4mm, fill=red!45] (box7_4) {$y_{4}$};
\node [fbox, right of=box7_4, xshift=-4mm, fill=orange!45] (box7_5) {$y_{5}$};
\node [hbox, right of=box7_5, xshift=-2.0mm] (box7_6) {\phantom{$y$}};
\node [hbox, right of=box7_6, xshift=-4mm] (box7_7) {\phantom{$y$}};
\node [rbox, right of=box7_7, xshift=-4mm, fill=orange!15] (box7_8) {$y_{8}$};
\node [rbox, yshift=-5.6cm, fill=red!15] (box8_0) {$y_{0}$};
\node [rbox, right of=box8_0, xshift=-4mm, fill=red!15] (box8_1) {$y_{1}$};
\node [rbox, right of=box8_1, xshift=-4mm, fill=orange!15] (box8_2) {$y_{2}$};
\node [rbox, right of=box8_2, xshift=-2.0mm, fill=red!15] (box8_3) {$y_{3}$};
\node [fbox, right of=box8_3, xshift=-4mm, fill=red!45] (box8_4) {$y_{4}$};
\node [tbox, right of=box8_4, xshift=-4mm, fill=orange!45] (box8_5) {$y_{5}$};
\node [hbox, right of=box8_5, xshift=-2.0mm] (box8_6) {\phantom{$y$}};
\node [hbox, right of=box8_6, xshift=-4mm] (box8_7) {\phantom{$y$}};
\node [rbox, right of=box8_7, xshift=-4mm, fill=orange!15] (box8_8) {$y_{8}$};
\draw[orange, bend left,->, line width=0.9mm,opacity=0.5]  (box0_2) to node [auto] {} (box1_5);
\draw[orange, bend left,->, line width=0.9mm,opacity=0.5]  (box1_5) to node [auto] {} (box2_8);
\draw[red, bend left,->, line width=0.9mm,opacity=0.5]  (box3_0) to node [auto] {} (box4_1);
\draw[red, bend left,->, line width=0.9mm,opacity=0.5]  (box4_1) to node [auto] {} (box5_2);
\draw[red, bend left,->, line width=0.9mm,opacity=0.5]  (box5_2) to node [auto] {} (box6_3);
\draw[red, bend left,->, line width=0.9mm,opacity=0.5]  (box6_3) to node [auto] {} (box7_4);
\draw[red, bend left,->, line width=0.9mm,opacity=0.5]  (box7_4) to node [auto] {} (box8_5);
\matrix[draw,thick,below left,fill=white,inner sep=1.5pt] at ([xshift=1.845cm,yshift=-3pt]current bounding box.north west) {
  \draw[->,bend left, line width=0.9mm,opacity=0.5,color=orange] (0,0) -- ++ (0.35,0); \node[right,xshift=0.3cm]{\textsc{rnn}$_2$};\\
  \draw[->,bend left, line width=0.9mm,opacity=0.5,color=red] (0,0) -- ++ (0.35,0); \node[right,xshift=0.3cm]{\textsc{rnn}$_0$};\\
};
\end{tikzpicture}

%% file: tikz_figures/paper_hf_multiv.regular.interleaved.3.9.tex
\begin{tikzpicture}
\draw[fill=gray!7,draw=none] (-0.9900000000000001cm,-0.35) rectangle ++(6.6,0.7);
\node [algstep,xshift=-0.81cm,yshift=-0.0cm] {\footnotesize 0};
\draw[fill=gray!22,draw=none] (-0.9900000000000001cm,-1.0499999999999998) rectangle ++(6.6,0.7);
\node [algstep,xshift=-0.81cm,yshift=-0.7cm] {\footnotesize 1};
\draw[fill=gray!7,draw=none] (-0.9900000000000001cm,-1.75) rectangle ++(6.6,0.7);
\node [algstep,xshift=-0.81cm,yshift=-1.4cm] {\footnotesize 2};
\draw[fill=gray!22,draw=none] (-0.9900000000000001cm,-2.4499999999999997) rectangle ++(6.6,0.7);
\node [algstep,xshift=-0.81cm,yshift=-2.0999999999999996cm] {\footnotesize 3};
\draw[fill=gray!7,draw=none] (-0.9900000000000001cm,-3.15) rectangle ++(6.6,0.7);
\node [algstep,xshift=-0.81cm,yshift=-2.8cm] {\footnotesize 4};
\draw[fill=gray!22,draw=none] (-0.9900000000000001cm,-3.85) rectangle ++(6.6,0.7);
\node [algstep,xshift=-0.81cm,yshift=-3.5cm] {\footnotesize 5};
\draw[fill=gray!7,draw=none] (-0.9900000000000001cm,-4.549999999999999) rectangle ++(6.6,0.7);
\node [algstep,xshift=-0.81cm,yshift=-4.199999999999999cm] {\footnotesize 6};
\draw[fill=gray!22,draw=none] (-0.9900000000000001cm,-5.249999999999999) rectangle ++(6.6,0.7);
\node [algstep,xshift=-0.81cm,yshift=-4.8999999999999995cm] {\footnotesize 7};
\draw[fill=gray!7,draw=none] (-0.9900000000000001cm,-5.949999999999999) rectangle ++(6.6,0.7);
\node [algstep,xshift=-0.81cm,yshift=-5.6cm] {\footnotesize 8};
\node [tbox, fill=red!45] (box0_0) {$y_{0}$};
\node [hbox, right of=box0_0, xshift=-4mm] (box0_1) {\phantom{$y$}};
\node [hbox, right of=box0_1, xshift=-4mm] (box0_2) {\phantom{$y$}};
\node [hbox, right of=box0_2, xshift=-2.0mm] (box0_3) {\phantom{$y$}};
\node [hbox, right of=box0_3, xshift=-4mm] (box0_4) {\phantom{$y$}};
\node [hbox, right of=box0_4, xshift=-4mm] (box0_5) {\phantom{$y$}};
\node [hbox, right of=box0_5, xshift=-2.0mm] (box0_6) {\phantom{$y$}};
\node [hbox, right of=box0_6, xshift=-4mm] (box0_7) {\phantom{$y$}};
\node [hbox, right of=box0_7, xshift=-4mm] (box0_8) {\phantom{$y$}};
\node [fbox, yshift=-0.7cm, fill=red!45] (box1_0) {$y_{0}$};
\node [tbox, right of=box1_0, xshift=-4mm, fill=green!45] (box1_1) {$y_{1}$};
\node [hbox, right of=box1_1, xshift=-4mm] (box1_2) {\phantom{$y$}};
\node [hbox, right of=box1_2, xshift=-2.0mm] (box1_3) {\phantom{$y$}};
\node [hbox, right of=box1_3, xshift=-4mm] (box1_4) {\phantom{$y$}};
\node [hbox, right of=box1_4, xshift=-4mm] (box1_5) {\phantom{$y$}};
\node [hbox, right of=box1_5, xshift=-2.0mm] (box1_6) {\phantom{$y$}};
\node [hbox, right of=box1_6, xshift=-4mm] (box1_7) {\phantom{$y$}};
\node [hbox, right of=box1_7, xshift=-4mm] (box1_8) {\phantom{$y$}};
\node [fbox, yshift=-1.4cm, fill=red!45] (box2_0) {$y_{0}$};
\node [fbox, right of=box2_0, xshift=-4mm, fill=green!45] (box2_1) {$y_{1}$};
\node [tbox, right of=box2_1, xshift=-4mm, fill=orange!45] (box2_2) {$y_{2}$};
\node [hbox, right of=box2_2, xshift=-2.0mm] (box2_3) {\phantom{$y$}};
\node [hbox, right of=box2_3, xshift=-4mm] (box2_4) {\phantom{$y$}};
\node [hbox, right of=box2_4, xshift=-4mm] (box2_5) {\phantom{$y$}};
\node [hbox, right of=box2_5, xshift=-2.0mm] (box2_6) {\phantom{$y$}};
\node [hbox, right of=box2_6, xshift=-4mm] (box2_7) {\phantom{$y$}};
\node [hbox, right of=box2_7, xshift=-4mm] (box2_8) {\phantom{$y$}};
\node [fbox, yshift=-2.0999999999999996cm, fill=red!45] (box3_0) {$y_{0}$};
\node [fbox, right of=box3_0, xshift=-4mm, fill=green!45] (box3_1) {$y_{1}$};
\node [fbox, right of=box3_1, xshift=-4mm, fill=orange!45] (box3_2) {$y_{2}$};
\node [tbox, right of=box3_2, xshift=-2.0mm, fill=red!45] (box3_3) {$y_{3}$};
\node [hbox, right of=box3_3, xshift=-4mm] (box3_4) {\phantom{$y$}};
\node [hbox, right of=box3_4, xshift=-4mm] (box3_5) {\phantom{$y$}};
\node [hbox, right of=box3_5, xshift=-2.0mm] (box3_6) {\phantom{$y$}};
\node [hbox, right of=box3_6, xshift=-4mm] (box3_7) {\phantom{$y$}};
\node [hbox, right of=box3_7, xshift=-4mm] (box3_8) {\phantom{$y$}};
\node [rbox, yshift=-2.8cm, fill=red!15] (box4_0) {$y_{0}$};
\node [fbox, right of=box4_0, xshift=-4mm, fill=green!45] (box4_1) {$y_{1}$};
\node [fbox, right of=box4_1, xshift=-4mm, fill=orange!45] (box4_2) {$y_{2}$};
\node [fbox, right of=box4_2, xshift=-2.0mm, fill=red!45] (box4_3) {$y_{3}$};
\node [tbox, right of=box4_3, xshift=-4mm, fill=green!45] (box4_4) {$y_{4}$};
\node [hbox, right of=box4_4, xshift=-4mm] (box4_5) {\phantom{$y$}};
\node [hbox, right of=box4_5, xshift=-2.0mm] (box4_6) {\phantom{$y$}};
\node [hbox, right of=box4_6, xshift=-4mm] (box4_7) {\phantom{$y$}};
\node [hbox, right of=box4_7, xshift=-4mm] (box4_8) {\phantom{$y$}};
\node [rbox, yshift=-3.5cm, fill=red!15] (box5_0) {$y_{0}$};
\node [rbox, right of=box5_0, xshift=-4mm, fill=green!15] (box5_1) {$y_{1}$};
\node [fbox, right of=box5_1, xshift=-4mm, fill=orange!45] (box5_2) {$y_{2}$};
\node [fbox, right of=box5_2, xshift=-2.0mm, fill=red!45] (box5_3) {$y_{3}$};
\node [fbox, right of=box5_3, xshift=-4mm, fill=green!45] (box5_4) {$y_{4}$};
\node [tbox, right of=box5_4, xshift=-4mm, fill=orange!45] (box5_5) {$y_{5}$};
\node [hbox, right of=box5_5, xshift=-2.0mm] (box5_6) {\phantom{$y$}};
\node [hbox, right of=box5_6, xshift=-4mm] (box5_7) {\phantom{$y$}};
\node [hbox, right of=box5_7, xshift=-4mm] (box5_8) {\phantom{$y$}};
\node [rbox, yshift=-4.199999999999999cm, fill=red!15] (box6_0) {$y_{0}$};
\node [rbox, right of=box6_0, xshift=-4mm, fill=green!15] (box6_1) {$y_{1}$};
\node [rbox, right of=box6_1, xshift=-4mm, fill=orange!15] (box6_2) {$y_{2}$};
\node [fbox, right of=box6_2, xshift=-2.0mm, fill=red!45] (box6_3) {$y_{3}$};
\node [fbox, right of=box6_3, xshift=-4mm, fill=green!45] (box6_4) {$y_{4}$};
\node [fbox, right of=box6_4, xshift=-4mm, fill=orange!45] (box6_5) {$y_{5}$};
\node [tbox, right of=box6_5, xshift=-2.0mm, fill=red!45] (box6_6) {$y_{6}$};
\node [hbox, right of=box6_6, xshift=-4mm] (box6_7) {\phantom{$y$}};
\node [hbox, right of=box6_7, xshift=-4mm] (box6_8) {\phantom{$y$}};
\node [rbox, yshift=-4.8999999999999995cm, fill=red!15] (box7_0) {$y_{0}$};
\node [rbox, right of=box7_0, xshift=-4mm, fill=green!15] (box7_1) {$y_{1}$};
\node [rbox, right of=box7_1, xshift=-4mm, fill=orange!15] (box7_2) {$y_{2}$};
\node [rbox, right of=box7_2, xshift=-2.0mm, fill=red!15] (box7_3) {$y_{3}$};
\node [fbox, right of=box7_3, xshift=-4mm, fill=green!45] (box7_4) {$y_{4}$};
\node [fbox, right of=box7_4, xshift=-4mm, fill=orange!45] (box7_5) {$y_{5}$};
\node [fbox, right of=box7_5, xshift=-2.0mm, fill=red!45] (box7_6) {$y_{6}$};
\node [tbox, right of=box7_6, xshift=-4mm, fill=green!45] (box7_7) {$y_{7}$};
\node [hbox, right of=box7_7, xshift=-4mm] (box7_8) {\phantom{$y$}};
\node [rbox, yshift=-5.6cm, fill=red!15] (box8_0) {$y_{0}$};
\node [rbox, right of=box8_0, xshift=-4mm, fill=green!15] (box8_1) {$y_{1}$};
\node [rbox, right of=box8_1, xshift=-4mm, fill=orange!15] (box8_2) {$y_{2}$};
\node [rbox, right of=box8_2, xshift=-2.0mm, fill=red!15] (box8_3) {$y_{3}$};
\node [rbox, right of=box8_3, xshift=-4mm, fill=green!15] (box8_4) {$y_{4}$};
\node [fbox, right of=box8_4, xshift=-4mm, fill=orange!45] (box8_5) {$y_{5}$};
\node [fbox, right of=box8_5, xshift=-2.0mm, fill=red!45] (box8_6) {$y_{6}$};
\node [fbox, right of=box8_6, xshift=-4mm, fill=green!45] (box8_7) {$y_{7}$};
\node [tbox, right of=box8_7, xshift=-4mm, fill=orange!45] (box8_8) {$y_{8}$};
\draw[red, bend left,->, line width=0.9mm,opacity=0.5]  (box0_0) to node [auto] {} (box3_3);
\draw[green, bend left,->, line width=0.9mm,opacity=0.5]  (box1_1) to node [auto] {} (box4_4);
\draw[orange, bend left,->, line width=0.9mm,opacity=0.5]  (box2_2) to node [auto] {} (box5_5);
\draw[red, bend left,->, line width=0.9mm,opacity=0.5]  (box3_3) to node [auto] {} (box6_6);
\draw[green, bend left,->, line width=0.9mm,opacity=0.5]  (box4_4) to node [auto] {} (box7_7);
\draw[orange, bend left,->, line width=0.9mm,opacity=0.5]  (box5_5) to node [auto] {} (box8_8);
\matrix[draw,thick,below left,fill=white,inner sep=1.5pt] at ([xshift=-0.18000000000000002cm,yshift=-3pt]current bounding box.north east) {
  \draw[->,bend left, line width=0.9mm,opacity=0.5,color=red] (0,0) -- ++ (0.35,0); \node[right,xshift=0.3cm]{\textsc{rnn}$_0$};\\
  \draw[->,bend left, line width=0.9mm,opacity=0.5,color=green] (0,0) -- ++ (0.35,0); \node[right,xshift=0.3cm]{\textsc{rnn}$_1$};\\
  \draw[->,bend left, line width=0.9mm,opacity=0.5,color=orange] (0,0) -- ++ (0.35,0); \node[right,xshift=0.3cm]{\textsc{rnn}$_2$};\\
};
\end{tikzpicture}

%% file: tikz_figures/paper_hf_multiv.regular.wall2wall.3.9.tex
\begin{tikzpicture}
\draw[fill=gray!7,draw=none] (-0.9900000000000001cm,-0.35) rectangle ++(6.6,0.7);
\node [algstep,xshift=-0.81cm,yshift=-0.0cm] {\footnotesize 0};
\draw[fill=gray!22,draw=none] (-0.9900000000000001cm,-1.0499999999999998) rectangle ++(6.6,0.7);
\node [algstep,xshift=-0.81cm,yshift=-0.7cm] {\footnotesize 1};
\draw[fill=gray!7,draw=none] (-0.9900000000000001cm,-1.75) rectangle ++(6.6,0.7);
\node [algstep,xshift=-0.81cm,yshift=-1.4cm] {\footnotesize 2};
\draw[fill=gray!22,draw=none] (-0.9900000000000001cm,-2.4499999999999997) rectangle ++(6.6,0.7);
\node [algstep,xshift=-0.81cm,yshift=-2.0999999999999996cm] {\footnotesize 3};
\draw[fill=gray!7,draw=none] (-0.9900000000000001cm,-3.15) rectangle ++(6.6,0.7);
\node [algstep,xshift=-0.81cm,yshift=-2.8cm] {\footnotesize 4};
\draw[fill=gray!22,draw=none] (-0.9900000000000001cm,-3.85) rectangle ++(6.6,0.7);
\node [algstep,xshift=-0.81cm,yshift=-3.5cm] {\footnotesize 5};
\draw[fill=gray!7,draw=none] (-0.9900000000000001cm,-4.549999999999999) rectangle ++(6.6,0.7);
\node [algstep,xshift=-0.81cm,yshift=-4.199999999999999cm] {\footnotesize 6};
\draw[fill=gray!22,draw=none] (-0.9900000000000001cm,-5.249999999999999) rectangle ++(6.6,0.7);
\node [algstep,xshift=-0.81cm,yshift=-4.8999999999999995cm] {\footnotesize 7};
\draw[fill=gray!7,draw=none] (-0.9900000000000001cm,-5.949999999999999) rectangle ++(6.6,0.7);
\node [algstep,xshift=-0.81cm,yshift=-5.6cm] {\footnotesize 8};
\node [tbox, fill=red!45] (box0_0) {$y_{0}$};
\node [hbox, right of=box0_0, xshift=-4mm] (box0_1) {\phantom{$y$}};
\node [hbox, right of=box0_1, xshift=-4mm] (box0_2) {\phantom{$y$}};
\node [hbox, right of=box0_2, xshift=-2.0mm] (box0_3) {\phantom{$y$}};
\node [hbox, right of=box0_3, xshift=-4mm] (box0_4) {\phantom{$y$}};
\node [hbox, right of=box0_4, xshift=-4mm] (box0_5) {\phantom{$y$}};
\node [hbox, right of=box0_5, xshift=-2.0mm] (box0_6) {\phantom{$y$}};
\node [hbox, right of=box0_6, xshift=-4mm] (box0_7) {\phantom{$y$}};
\node [hbox, right of=box0_7, xshift=-4mm] (box0_8) {\phantom{$y$}};
\node [fbox, yshift=-0.7cm, fill=red!45] (box1_0) {$y_{0}$};
\node [hbox, right of=box1_0, xshift=-4mm] (box1_1) {\phantom{$y$}};
\node [hbox, right of=box1_1, xshift=-4mm] (box1_2) {\phantom{$y$}};
\node [tbox, right of=box1_2, xshift=-2.0mm, fill=red!45] (box1_3) {$y_{3}$};
\node [hbox, right of=box1_3, xshift=-4mm] (box1_4) {\phantom{$y$}};
\node [hbox, right of=box1_4, xshift=-4mm] (box1_5) {\phantom{$y$}};
\node [hbox, right of=box1_5, xshift=-2.0mm] (box1_6) {\phantom{$y$}};
\node [hbox, right of=box1_6, xshift=-4mm] (box1_7) {\phantom{$y$}};
\node [hbox, right of=box1_7, xshift=-4mm] (box1_8) {\phantom{$y$}};
\node [rbox, yshift=-1.4cm, fill=red!15] (box2_0) {$y_{0}$};
\node [hbox, right of=box2_0, xshift=-4mm] (box2_1) {\phantom{$y$}};
\node [hbox, right of=box2_1, xshift=-4mm] (box2_2) {\phantom{$y$}};
\node [fbox, right of=box2_2, xshift=-2.0mm, fill=red!45] (box2_3) {$y_{3}$};
\node [hbox, right of=box2_3, xshift=-4mm] (box2_4) {\phantom{$y$}};
\node [hbox, right of=box2_4, xshift=-4mm] (box2_5) {\phantom{$y$}};
\node [tbox, right of=box2_5, xshift=-2.0mm, fill=red!45] (box2_6) {$y_{6}$};
\node [hbox, right of=box2_6, xshift=-4mm] (box2_7) {\phantom{$y$}};
\node [hbox, right of=box2_7, xshift=-4mm] (box2_8) {\phantom{$y$}};
\node [fbox, yshift=-2.0999999999999996cm, fill=red!45] (box3_0) {$y_{0}$};
\node [tbox, right of=box3_0, xshift=-4mm, fill=green!45] (box3_1) {$y_{1}$};
\node [hbox, right of=box3_1, xshift=-4mm] (box3_2) {\phantom{$y$}};
\node [rbox, right of=box3_2, xshift=-2.0mm, fill=red!15] (box3_3) {$y_{3}$};
\node [hbox, right of=box3_3, xshift=-4mm] (box3_4) {\phantom{$y$}};
\node [hbox, right of=box3_4, xshift=-4mm] (box3_5) {\phantom{$y$}};
\node [rbox, right of=box3_5, xshift=-2.0mm, fill=red!15] (box3_6) {$y_{6}$};
\node [hbox, right of=box3_6, xshift=-4mm] (box3_7) {\phantom{$y$}};
\node [hbox, right of=box3_7, xshift=-4mm] (box3_8) {\phantom{$y$}};
\node [rbox, yshift=-2.8cm, fill=red!15] (box4_0) {$y_{0}$};
\node [fbox, right of=box4_0, xshift=-4mm, fill=green!45] (box4_1) {$y_{1}$};
\node [hbox, right of=box4_1, xshift=-4mm] (box4_2) {\phantom{$y$}};
\node [fbox, right of=box4_2, xshift=-2.0mm, fill=red!45] (box4_3) {$y_{3}$};
\node [tbox, right of=box4_3, xshift=-4mm, fill=green!45] (box4_4) {$y_{4}$};
\node [hbox, right of=box4_4, xshift=-4mm] (box4_5) {\phantom{$y$}};
\node [rbox, right of=box4_5, xshift=-2.0mm, fill=red!15] (box4_6) {$y_{6}$};
\node [hbox, right of=box4_6, xshift=-4mm] (box4_7) {\phantom{$y$}};
\node [hbox, right of=box4_7, xshift=-4mm] (box4_8) {\phantom{$y$}};
\node [rbox, yshift=-3.5cm, fill=red!15] (box5_0) {$y_{0}$};
\node [rbox, right of=box5_0, xshift=-4mm, fill=green!15] (box5_1) {$y_{1}$};
\node [hbox, right of=box5_1, xshift=-4mm] (box5_2) {\phantom{$y$}};
\node [rbox, right of=box5_2, xshift=-2.0mm, fill=red!15] (box5_3) {$y_{3}$};
\node [fbox, right of=box5_3, xshift=-4mm, fill=green!45] (box5_4) {$y_{4}$};
\node [hbox, right of=box5_4, xshift=-4mm] (box5_5) {\phantom{$y$}};
\node [fbox, right of=box5_5, xshift=-2.0mm, fill=red!45] (box5_6) {$y_{6}$};
\node [tbox, right of=box5_6, xshift=-4mm, fill=green!45] (box5_7) {$y_{7}$};
\node [hbox, right of=box5_7, xshift=-4mm] (box5_8) {\phantom{$y$}};
\node [fbox, yshift=-4.199999999999999cm, fill=red!45] (box6_0) {$y_{0}$};
\node [fbox, right of=box6_0, xshift=-4mm, fill=green!45] (box6_1) {$y_{1}$};
\node [tbox, right of=box6_1, xshift=-4mm, fill=orange!45] (box6_2) {$y_{2}$};
\node [rbox, right of=box6_2, xshift=-2.0mm, fill=red!15] (box6_3) {$y_{3}$};
\node [rbox, right of=box6_3, xshift=-4mm, fill=green!15] (box6_4) {$y_{4}$};
\node [hbox, right of=box6_4, xshift=-4mm] (box6_5) {\phantom{$y$}};
\node [rbox, right of=box6_5, xshift=-2.0mm, fill=red!15] (box6_6) {$y_{6}$};
\node [rbox, right of=box6_6, xshift=-4mm, fill=green!15] (box6_7) {$y_{7}$};
\node [hbox, right of=box6_7, xshift=-4mm] (box6_8) {\phantom{$y$}};
\node [rbox, yshift=-4.8999999999999995cm, fill=red!15] (box7_0) {$y_{0}$};
\node [rbox, right of=box7_0, xshift=-4mm, fill=green!15] (box7_1) {$y_{1}$};
\node [fbox, right of=box7_1, xshift=-4mm, fill=orange!45] (box7_2) {$y_{2}$};
\node [fbox, right of=box7_2, xshift=-2.0mm, fill=red!45] (box7_3) {$y_{3}$};
\node [fbox, right of=box7_3, xshift=-4mm, fill=green!45] (box7_4) {$y_{4}$};
\node [tbox, right of=box7_4, xshift=-4mm, fill=orange!45] (box7_5) {$y_{5}$};
\node [rbox, right of=box7_5, xshift=-2.0mm, fill=red!15] (box7_6) {$y_{6}$};
\node [rbox, right of=box7_6, xshift=-4mm, fill=green!15] (box7_7) {$y_{7}$};
\node [hbox, right of=box7_7, xshift=-4mm] (box7_8) {\phantom{$y$}};
\node [rbox, yshift=-5.6cm, fill=red!15] (box8_0) {$y_{0}$};
\node [rbox, right of=box8_0, xshift=-4mm, fill=green!15] (box8_1) {$y_{1}$};
\node [rbox, right of=box8_1, xshift=-4mm, fill=orange!15] (box8_2) {$y_{2}$};
\node [rbox, right of=box8_2, xshift=-2.0mm, fill=red!15] (box8_3) {$y_{3}$};
\node [rbox, right of=box8_3, xshift=-4mm, fill=green!15] (box8_4) {$y_{4}$};
\node [fbox, right of=box8_4, xshift=-4mm, fill=orange!45] (box8_5) {$y_{5}$};
\node [fbox, right of=box8_5, xshift=-2.0mm, fill=red!45] (box8_6) {$y_{6}$};
\node [fbox, right of=box8_6, xshift=-4mm, fill=green!45] (box8_7) {$y_{7}$};
\node [tbox, right of=box8_7, xshift=-4mm, fill=orange!45] (box8_8) {$y_{8}$};
\draw[red, bend left,->, line width=0.9mm,opacity=0.5]  (box0_0) to node [auto] {} (box1_3);
\draw[red, bend left,->, line width=0.9mm,opacity=0.5]  (box1_3) to node [auto] {} (box2_6);
\draw[green, bend left,->, line width=0.9mm,opacity=0.5]  (box3_1) to node [auto] {} (box4_4);
\draw[green, bend left,->, line width=0.9mm,opacity=0.5]  (box4_4) to node [auto] {} (box5_7);
\draw[orange, bend left,->, line width=0.9mm,opacity=0.5]  (box6_2) to node [auto] {} (box7_5);
\draw[orange, bend left,->, line width=0.9mm,opacity=0.5]  (box7_5) to node [auto] {} (box8_8);
\matrix[draw,thick,below left,fill=white,inner sep=1.5pt] at ([xshift=-0.18000000000000002cm,yshift=-3pt]current bounding box.north east) {
  \draw[->,bend left, line width=0.9mm,opacity=0.5,color=red] (0,0) -- ++ (0.35,0); \node[right,xshift=0.3cm]{\textsc{rnn}$_0$};\\
  \draw[->,bend left, line width=0.9mm,opacity=0.5,color=green] (0,0) -- ++ (0.35,0); \node[right,xshift=0.3cm]{\textsc{rnn}$_1$};\\
  \draw[->,bend left, line width=0.9mm,opacity=0.5,color=orange] (0,0) -- ++ (0.35,0); \node[right,xshift=0.3cm]{\textsc{rnn}$_2$};\\
};
\end{tikzpicture}

%% file: tikz_figures/paper_hf_multiv.backfill.wall2wall.3.9.tex
\begin{tikzpicture}
\draw[fill=gray!7,draw=none] (-0.9900000000000001cm,-0.35) rectangle ++(6.6,0.7);
\node [algstep,xshift=-0.81cm,yshift=-0.0cm] {\footnotesize 0};
\draw[fill=gray!22,draw=none] (-0.9900000000000001cm,-1.0499999999999998) rectangle ++(6.6,0.7);
\node [algstep,xshift=-0.81cm,yshift=-0.7cm] {\footnotesize 1};
\draw[fill=gray!7,draw=none] (-0.9900000000000001cm,-1.75) rectangle ++(6.6,0.7);
\node [algstep,xshift=-0.81cm,yshift=-1.4cm] {\footnotesize 2};
\draw[fill=gray!22,draw=none] (-0.9900000000000001cm,-2.4499999999999997) rectangle ++(6.6,0.7);
\node [algstep,xshift=-0.81cm,yshift=-2.0999999999999996cm] {\footnotesize 3};
\draw[fill=gray!7,draw=none] (-0.9900000000000001cm,-3.15) rectangle ++(6.6,0.7);
\node [algstep,xshift=-0.81cm,yshift=-2.8cm] {\footnotesize 4};
\draw[fill=gray!22,draw=none] (-0.9900000000000001cm,-3.85) rectangle ++(6.6,0.7);
\node [algstep,xshift=-0.81cm,yshift=-3.5cm] {\footnotesize 5};
\draw[fill=gray!7,draw=none] (-0.9900000000000001cm,-4.549999999999999) rectangle ++(6.6,0.7);
\node [algstep,xshift=-0.81cm,yshift=-4.199999999999999cm] {\footnotesize 6};
\draw[fill=gray!22,draw=none] (-0.9900000000000001cm,-5.249999999999999) rectangle ++(6.6,0.7);
\node [algstep,xshift=-0.81cm,yshift=-4.8999999999999995cm] {\footnotesize 7};
\draw[fill=gray!7,draw=none] (-0.9900000000000001cm,-5.949999999999999) rectangle ++(6.6,0.7);
\node [algstep,xshift=-0.81cm,yshift=-5.6cm] {\footnotesize 8};
\node [hbox] (box0_0) {\phantom{$y$}};
\node [hbox, right of=box0_0, xshift=-4mm] (box0_1) {\phantom{$y$}};
\node [tbox, right of=box0_1, xshift=-4mm, fill=orange!45] (box0_2) {$y_{2}$};
\node [hbox, right of=box0_2, xshift=-2.0mm] (box0_3) {\phantom{$y$}};
\node [hbox, right of=box0_3, xshift=-4mm] (box0_4) {\phantom{$y$}};
\node [hbox, right of=box0_4, xshift=-4mm] (box0_5) {\phantom{$y$}};
\node [hbox, right of=box0_5, xshift=-2.0mm] (box0_6) {\phantom{$y$}};
\node [hbox, right of=box0_6, xshift=-4mm] (box0_7) {\phantom{$y$}};
\node [hbox, right of=box0_7, xshift=-4mm] (box0_8) {\phantom{$y$}};
\node [hbox, yshift=-0.7cm] (box1_0) {\phantom{$y$}};
\node [hbox, right of=box1_0, xshift=-4mm] (box1_1) {\phantom{$y$}};
\node [fbox, right of=box1_1, xshift=-4mm, fill=orange!45] (box1_2) {$y_{2}$};
\node [hbox, right of=box1_2, xshift=-2.0mm] (box1_3) {\phantom{$y$}};
\node [hbox, right of=box1_3, xshift=-4mm] (box1_4) {\phantom{$y$}};
\node [tbox, right of=box1_4, xshift=-4mm, fill=orange!45] (box1_5) {$y_{5}$};
\node [hbox, right of=box1_5, xshift=-2.0mm] (box1_6) {\phantom{$y$}};
\node [hbox, right of=box1_6, xshift=-4mm] (box1_7) {\phantom{$y$}};
\node [hbox, right of=box1_7, xshift=-4mm] (box1_8) {\phantom{$y$}};
\node [hbox, yshift=-1.4cm] (box2_0) {\phantom{$y$}};
\node [hbox, right of=box2_0, xshift=-4mm] (box2_1) {\phantom{$y$}};
\node [rbox, right of=box2_1, xshift=-4mm, fill=orange!15] (box2_2) {$y_{2}$};
\node [hbox, right of=box2_2, xshift=-2.0mm] (box2_3) {\phantom{$y$}};
\node [hbox, right of=box2_3, xshift=-4mm] (box2_4) {\phantom{$y$}};
\node [fbox, right of=box2_4, xshift=-4mm, fill=orange!45] (box2_5) {$y_{5}$};
\node [hbox, right of=box2_5, xshift=-2.0mm] (box2_6) {\phantom{$y$}};
\node [hbox, right of=box2_6, xshift=-4mm] (box2_7) {\phantom{$y$}};
\node [tbox, right of=box2_7, xshift=-4mm, fill=orange!45] (box2_8) {$y_{8}$};
\node [hbox, yshift=-2.0999999999999996cm] (box3_0) {\phantom{$y$}};
\node [tbox, right of=box3_0, xshift=-4mm, fill=green!45] (box3_1) {$y_{1}$};
\node [fbox, right of=box3_1, xshift=-4mm, fill=orange!45] (box3_2) {$y_{2}$};
\node [hbox, right of=box3_2, xshift=-2.0mm] (box3_3) {\phantom{$y$}};
\node [hbox, right of=box3_3, xshift=-4mm] (box3_4) {\phantom{$y$}};
\node [rbox, right of=box3_4, xshift=-4mm, fill=orange!15] (box3_5) {$y_{5}$};
\node [hbox, right of=box3_5, xshift=-2.0mm] (box3_6) {\phantom{$y$}};
\node [hbox, right of=box3_6, xshift=-4mm] (box3_7) {\phantom{$y$}};
\node [rbox, right of=box3_7, xshift=-4mm, fill=orange!15] (box3_8) {$y_{8}$};
\node [hbox, yshift=-2.8cm] (box4_0) {\phantom{$y$}};
\node [fbox, right of=box4_0, xshift=-4mm, fill=green!45] (box4_1) {$y_{1}$};
\node [rbox, right of=box4_1, xshift=-4mm, fill=orange!15] (box4_2) {$y_{2}$};
\node [hbox, right of=box4_2, xshift=-2.0mm] (box4_3) {\phantom{$y$}};
\node [tbox, right of=box4_3, xshift=-4mm, fill=green!45] (box4_4) {$y_{4}$};
\node [fbox, right of=box4_4, xshift=-4mm, fill=orange!45] (box4_5) {$y_{5}$};
\node [hbox, right of=box4_5, xshift=-2.0mm] (box4_6) {\phantom{$y$}};
\node [hbox, right of=box4_6, xshift=-4mm] (box4_7) {\phantom{$y$}};
\node [rbox, right of=box4_7, xshift=-4mm, fill=orange!15] (box4_8) {$y_{8}$};
\node [hbox, yshift=-3.5cm] (box5_0) {\phantom{$y$}};
\node [rbox, right of=box5_0, xshift=-4mm, fill=green!15] (box5_1) {$y_{1}$};
\node [rbox, right of=box5_1, xshift=-4mm, fill=orange!15] (box5_2) {$y_{2}$};
\node [hbox, right of=box5_2, xshift=-2.0mm] (box5_3) {\phantom{$y$}};
\node [fbox, right of=box5_3, xshift=-4mm, fill=green!45] (box5_4) {$y_{4}$};
\node [rbox, right of=box5_4, xshift=-4mm, fill=orange!15] (box5_5) {$y_{5}$};
\node [hbox, right of=box5_5, xshift=-2.0mm] (box5_6) {\phantom{$y$}};
\node [tbox, right of=box5_6, xshift=-4mm, fill=green!45] (box5_7) {$y_{7}$};
\node [fbox, right of=box5_7, xshift=-4mm, fill=orange!45] (box5_8) {$y_{8}$};
\node [tbox, yshift=-4.199999999999999cm, fill=red!45] (box6_0) {$y_{0}$};
\node [fbox, right of=box6_0, xshift=-4mm, fill=green!45] (box6_1) {$y_{1}$};
\node [fbox, right of=box6_1, xshift=-4mm, fill=orange!45] (box6_2) {$y_{2}$};
\node [hbox, right of=box6_2, xshift=-2.0mm] (box6_3) {\phantom{$y$}};
\node [rbox, right of=box6_3, xshift=-4mm, fill=green!15] (box6_4) {$y_{4}$};
\node [rbox, right of=box6_4, xshift=-4mm, fill=orange!15] (box6_5) {$y_{5}$};
\node [hbox, right of=box6_5, xshift=-2.0mm] (box6_6) {\phantom{$y$}};
\node [rbox, right of=box6_6, xshift=-4mm, fill=green!15] (box6_7) {$y_{7}$};
\node [rbox, right of=box6_7, xshift=-4mm, fill=orange!15] (box6_8) {$y_{8}$};
\node [fbox, yshift=-4.8999999999999995cm, fill=red!45] (box7_0) {$y_{0}$};
\node [rbox, right of=box7_0, xshift=-4mm, fill=green!15] (box7_1) {$y_{1}$};
\node [rbox, right of=box7_1, xshift=-4mm, fill=orange!15] (box7_2) {$y_{2}$};
\node [tbox, right of=box7_2, xshift=-2.0mm, fill=red!45] (box7_3) {$y_{3}$};
\node [fbox, right of=box7_3, xshift=-4mm, fill=green!45] (box7_4) {$y_{4}$};
\node [fbox, right of=box7_4, xshift=-4mm, fill=orange!45] (box7_5) {$y_{5}$};
\node [hbox, right of=box7_5, xshift=-2.0mm] (box7_6) {\phantom{$y$}};
\node [rbox, right of=box7_6, xshift=-4mm, fill=green!15] (box7_7) {$y_{7}$};
\node [rbox, right of=box7_7, xshift=-4mm, fill=orange!15] (box7_8) {$y_{8}$};
\node [rbox, yshift=-5.6cm, fill=red!15] (box8_0) {$y_{0}$};
\node [rbox, right of=box8_0, xshift=-4mm, fill=green!15] (box8_1) {$y_{1}$};
\node [rbox, right of=box8_1, xshift=-4mm, fill=orange!15] (box8_2) {$y_{2}$};
\node [fbox, right of=box8_2, xshift=-2.0mm, fill=red!45] (box8_3) {$y_{3}$};
\node [rbox, right of=box8_3, xshift=-4mm, fill=green!15] (box8_4) {$y_{4}$};
\node [rbox, right of=box8_4, xshift=-4mm, fill=orange!15] (box8_5) {$y_{5}$};
\node [tbox, right of=box8_5, xshift=-2.0mm, fill=red!45] (box8_6) {$y_{6}$};
\node [fbox, right of=box8_6, xshift=-4mm, fill=green!45] (box8_7) {$y_{7}$};
\node [fbox, right of=box8_7, xshift=-4mm, fill=orange!45] (box8_8) {$y_{8}$};
\draw[orange, bend left,->, line width=0.9mm,opacity=0.5]  (box0_2) to node [auto] {} (box1_5);
\draw[orange, bend left,->, line width=0.9mm,opacity=0.5]  (box1_5) to node [auto] {} (box2_8);
\draw[green, bend left,->, line width=0.9mm,opacity=0.5]  (box3_1) to node [auto] {} (box4_4);
\draw[green, bend left,->, line width=0.9mm,opacity=0.5]  (box4_4) to node [auto] {} (box5_7);
\draw[red, bend left,->, line width=0.9mm,opacity=0.5]  (box6_0) to node [auto] {} (box7_3);
\draw[red, bend left,->, line width=0.9mm,opacity=0.5]  (box7_3) to node [auto] {} (box8_6);
\matrix[draw,thick,below left,fill=white,inner sep=1.5pt] at ([xshift=1.845cm,yshift=-3pt]current bounding box.north west) {
  \draw[->,bend left, line width=0.9mm,opacity=0.5,color=orange] (0,0) -- ++ (0.35,0); \node[right,xshift=0.3cm]{\textsc{rnn}$_2$};\\
  \draw[->,bend left, line width=0.9mm,opacity=0.5,color=green] (0,0) -- ++ (0.35,0); \node[right,xshift=0.3cm]{\textsc{rnn}$_1$};\\
  \draw[->,bend left, line width=0.9mm,opacity=0.5,color=red] (0,0) -- ++ (0.35,0); \node[right,xshift=0.3cm]{\textsc{rnn}$_0$};\\
};
\end{tikzpicture}

%% file: tikz_figures/paper_hf_multiv.backfill.interleaved.3.9.tex
\begin{tikzpicture}
\draw[fill=gray!7,draw=none] (-0.9900000000000001cm,-0.35) rectangle ++(6.6,0.7);
\node [algstep,xshift=-0.81cm,yshift=-0.0cm] {\footnotesize 0};
\draw[fill=gray!22,draw=none] (-0.9900000000000001cm,-1.0499999999999998) rectangle ++(6.6,0.7);
\node [algstep,xshift=-0.81cm,yshift=-0.7cm] {\footnotesize 1};
\draw[fill=gray!7,draw=none] (-0.9900000000000001cm,-1.75) rectangle ++(6.6,0.7);
\node [algstep,xshift=-0.81cm,yshift=-1.4cm] {\footnotesize 2};
\draw[fill=gray!22,draw=none] (-0.9900000000000001cm,-2.4499999999999997) rectangle ++(6.6,0.7);
\node [algstep,xshift=-0.81cm,yshift=-2.0999999999999996cm] {\footnotesize 3};
\draw[fill=gray!7,draw=none] (-0.9900000000000001cm,-3.15) rectangle ++(6.6,0.7);
\node [algstep,xshift=-0.81cm,yshift=-2.8cm] {\footnotesize 4};
\draw[fill=gray!22,draw=none] (-0.9900000000000001cm,-3.85) rectangle ++(6.6,0.7);
\node [algstep,xshift=-0.81cm,yshift=-3.5cm] {\footnotesize 5};
\draw[fill=gray!7,draw=none] (-0.9900000000000001cm,-4.549999999999999) rectangle ++(6.6,0.7);
\node [algstep,xshift=-0.81cm,yshift=-4.199999999999999cm] {\footnotesize 6};
\draw[fill=gray!22,draw=none] (-0.9900000000000001cm,-5.249999999999999) rectangle ++(6.6,0.7);
\node [algstep,xshift=-0.81cm,yshift=-4.8999999999999995cm] {\footnotesize 7};
\draw[fill=gray!7,draw=none] (-0.9900000000000001cm,-5.949999999999999) rectangle ++(6.6,0.7);
\node [algstep,xshift=-0.81cm,yshift=-5.6cm] {\footnotesize 8};
\node [hbox] (box0_0) {\phantom{$y$}};
\node [hbox, right of=box0_0, xshift=-4mm] (box0_1) {\phantom{$y$}};
\node [tbox, right of=box0_1, xshift=-4mm, fill=orange!45] (box0_2) {$y_{2}$};
\node [hbox, right of=box0_2, xshift=-2.0mm] (box0_3) {\phantom{$y$}};
\node [hbox, right of=box0_3, xshift=-4mm] (box0_4) {\phantom{$y$}};
\node [hbox, right of=box0_4, xshift=-4mm] (box0_5) {\phantom{$y$}};
\node [hbox, right of=box0_5, xshift=-2.0mm] (box0_6) {\phantom{$y$}};
\node [hbox, right of=box0_6, xshift=-4mm] (box0_7) {\phantom{$y$}};
\node [hbox, right of=box0_7, xshift=-4mm] (box0_8) {\phantom{$y$}};
\node [hbox, yshift=-0.7cm] (box1_0) {\phantom{$y$}};
\node [tbox, right of=box1_0, xshift=-4mm, fill=green!45] (box1_1) {$y_{1}$};
\node [fbox, right of=box1_1, xshift=-4mm, fill=orange!45] (box1_2) {$y_{2}$};
\node [hbox, right of=box1_2, xshift=-2.0mm] (box1_3) {\phantom{$y$}};
\node [hbox, right of=box1_3, xshift=-4mm] (box1_4) {\phantom{$y$}};
\node [hbox, right of=box1_4, xshift=-4mm] (box1_5) {\phantom{$y$}};
\node [hbox, right of=box1_5, xshift=-2.0mm] (box1_6) {\phantom{$y$}};
\node [hbox, right of=box1_6, xshift=-4mm] (box1_7) {\phantom{$y$}};
\node [hbox, right of=box1_7, xshift=-4mm] (box1_8) {\phantom{$y$}};
\node [tbox, yshift=-1.4cm, fill=red!45] (box2_0) {$y_{0}$};
\node [fbox, right of=box2_0, xshift=-4mm, fill=green!45] (box2_1) {$y_{1}$};
\node [fbox, right of=box2_1, xshift=-4mm, fill=orange!45] (box2_2) {$y_{2}$};
\node [hbox, right of=box2_2, xshift=-2.0mm] (box2_3) {\phantom{$y$}};
\node [hbox, right of=box2_3, xshift=-4mm] (box2_4) {\phantom{$y$}};
\node [hbox, right of=box2_4, xshift=-4mm] (box2_5) {\phantom{$y$}};
\node [hbox, right of=box2_5, xshift=-2.0mm] (box2_6) {\phantom{$y$}};
\node [hbox, right of=box2_6, xshift=-4mm] (box2_7) {\phantom{$y$}};
\node [hbox, right of=box2_7, xshift=-4mm] (box2_8) {\phantom{$y$}};
\node [fbox, yshift=-2.0999999999999996cm, fill=red!45] (box3_0) {$y_{0}$};
\node [fbox, right of=box3_0, xshift=-4mm, fill=green!45] (box3_1) {$y_{1}$};
\node [fbox, right of=box3_1, xshift=-4mm, fill=orange!45] (box3_2) {$y_{2}$};
\node [hbox, right of=box3_2, xshift=-2.0mm] (box3_3) {\phantom{$y$}};
\node [hbox, right of=box3_3, xshift=-4mm] (box3_4) {\phantom{$y$}};
\node [tbox, right of=box3_4, xshift=-4mm, fill=orange!45] (box3_5) {$y_{5}$};
\node [hbox, right of=box3_5, xshift=-2.0mm] (box3_6) {\phantom{$y$}};
\node [hbox, right of=box3_6, xshift=-4mm] (box3_7) {\phantom{$y$}};
\node [hbox, right of=box3_7, xshift=-4mm] (box3_8) {\phantom{$y$}};
\node [fbox, yshift=-2.8cm, fill=red!45] (box4_0) {$y_{0}$};
\node [fbox, right of=box4_0, xshift=-4mm, fill=green!45] (box4_1) {$y_{1}$};
\node [rbox, right of=box4_1, xshift=-4mm, fill=orange!15] (box4_2) {$y_{2}$};
\node [hbox, right of=box4_2, xshift=-2.0mm] (box4_3) {\phantom{$y$}};
\node [tbox, right of=box4_3, xshift=-4mm, fill=green!45] (box4_4) {$y_{4}$};
\node [fbox, right of=box4_4, xshift=-4mm, fill=orange!45] (box4_5) {$y_{5}$};
\node [hbox, right of=box4_5, xshift=-2.0mm] (box4_6) {\phantom{$y$}};
\node [hbox, right of=box4_6, xshift=-4mm] (box4_7) {\phantom{$y$}};
\node [hbox, right of=box4_7, xshift=-4mm] (box4_8) {\phantom{$y$}};
\node [fbox, yshift=-3.5cm, fill=red!45] (box5_0) {$y_{0}$};
\node [rbox, right of=box5_0, xshift=-4mm, fill=green!15] (box5_1) {$y_{1}$};
\node [rbox, right of=box5_1, xshift=-4mm, fill=orange!15] (box5_2) {$y_{2}$};
\node [tbox, right of=box5_2, xshift=-2.0mm, fill=red!45] (box5_3) {$y_{3}$};
\node [fbox, right of=box5_3, xshift=-4mm, fill=green!45] (box5_4) {$y_{4}$};
\node [fbox, right of=box5_4, xshift=-4mm, fill=orange!45] (box5_5) {$y_{5}$};
\node [hbox, right of=box5_5, xshift=-2.0mm] (box5_6) {\phantom{$y$}};
\node [hbox, right of=box5_6, xshift=-4mm] (box5_7) {\phantom{$y$}};
\node [hbox, right of=box5_7, xshift=-4mm] (box5_8) {\phantom{$y$}};
\node [rbox, yshift=-4.199999999999999cm, fill=red!15] (box6_0) {$y_{0}$};
\node [rbox, right of=box6_0, xshift=-4mm, fill=green!15] (box6_1) {$y_{1}$};
\node [rbox, right of=box6_1, xshift=-4mm, fill=orange!15] (box6_2) {$y_{2}$};
\node [fbox, right of=box6_2, xshift=-2.0mm, fill=red!45] (box6_3) {$y_{3}$};
\node [fbox, right of=box6_3, xshift=-4mm, fill=green!45] (box6_4) {$y_{4}$};
\node [fbox, right of=box6_4, xshift=-4mm, fill=orange!45] (box6_5) {$y_{5}$};
\node [hbox, right of=box6_5, xshift=-2.0mm] (box6_6) {\phantom{$y$}};
\node [hbox, right of=box6_6, xshift=-4mm] (box6_7) {\phantom{$y$}};
\node [tbox, right of=box6_7, xshift=-4mm, fill=orange!45] (box6_8) {$y_{8}$};
\node [rbox, yshift=-4.8999999999999995cm, fill=red!15] (box7_0) {$y_{0}$};
\node [rbox, right of=box7_0, xshift=-4mm, fill=green!15] (box7_1) {$y_{1}$};
\node [rbox, right of=box7_1, xshift=-4mm, fill=orange!15] (box7_2) {$y_{2}$};
\node [fbox, right of=box7_2, xshift=-2.0mm, fill=red!45] (box7_3) {$y_{3}$};
\node [fbox, right of=box7_3, xshift=-4mm, fill=green!45] (box7_4) {$y_{4}$};
\node [rbox, right of=box7_4, xshift=-4mm, fill=orange!15] (box7_5) {$y_{5}$};
\node [hbox, right of=box7_5, xshift=-2.0mm] (box7_6) {\phantom{$y$}};
\node [tbox, right of=box7_6, xshift=-4mm, fill=green!45] (box7_7) {$y_{7}$};
\node [fbox, right of=box7_7, xshift=-4mm, fill=orange!45] (box7_8) {$y_{8}$};
\node [rbox, yshift=-5.6cm, fill=red!15] (box8_0) {$y_{0}$};
\node [rbox, right of=box8_0, xshift=-4mm, fill=green!15] (box8_1) {$y_{1}$};
\node [rbox, right of=box8_1, xshift=-4mm, fill=orange!15] (box8_2) {$y_{2}$};
\node [fbox, right of=box8_2, xshift=-2.0mm, fill=red!45] (box8_3) {$y_{3}$};
\node [rbox, right of=box8_3, xshift=-4mm, fill=green!15] (box8_4) {$y_{4}$};
\node [rbox, right of=box8_4, xshift=-4mm, fill=orange!15] (box8_5) {$y_{5}$};
\node [tbox, right of=box8_5, xshift=-2.0mm, fill=red!45] (box8_6) {$y_{6}$};
\node [fbox, right of=box8_6, xshift=-4mm, fill=green!45] (box8_7) {$y_{7}$};
\node [fbox, right of=box8_7, xshift=-4mm, fill=orange!45] (box8_8) {$y_{8}$};
\draw[orange, bend left,->, line width=0.9mm,opacity=0.5]  (box0_2) to node [auto] {} (box3_5);
\draw[green, bend left,->, line width=0.9mm,opacity=0.5]  (box1_1) to node [auto] {} (box4_4);
\draw[red, bend left,->, line width=0.9mm,opacity=0.5]  (box2_0) to node [auto] {} (box5_3);
\draw[orange, bend left,->, line width=0.9mm,opacity=0.5]  (box3_5) to node [auto] {} (box6_8);
\draw[green, bend left,->, line width=0.9mm,opacity=0.5]  (box4_4) to node [auto] {} (box7_7);
\draw[red, bend left,->, line width=0.9mm,opacity=0.5]  (box5_3) to node [auto] {} (box8_6);
\matrix[draw,thick,below left,fill=white,inner sep=1.5pt] at ([xshift=-0.18000000000000002cm,yshift=-3pt]current bounding box.north east) {
  \draw[->,bend left, line width=0.9mm,opacity=0.5,color=orange] (0,0) -- ++ (0.35,0); \node[right,xshift=0.3cm]{\textsc{rnn}$_2$};\\
  \draw[->,bend left, line width=0.9mm,opacity=0.5,color=green] (0,0) -- ++ (0.35,0); \node[right,xshift=0.3cm]{\textsc{rnn}$_1$};\\
  \draw[->,bend left, line width=0.9mm,opacity=0.5,color=red] (0,0) -- ++ (0.35,0); \node[right,xshift=0.3cm]{\textsc{rnn}$_0$};\\
};
\end{tikzpicture}

%% file: limitations.tex
$\sutranets$ are more complex than standard sequence models in a few
ways.
First, they require more decisions, such as specifying the number of
sub-series.  While superior accuracy can be obtained when making these
choices heuristically, we evaluate the extent to which tuning can
achieve further gains (\S\ref{sec:experiments}).
Secondly, having $K$ separate RNNs results in roughly $K$ times as
many total parameters. Although, as noted above, this does not
translate into requiring more memory or training/inference time,
$\sutranet$ models are larger to store and transmit.

As a flexible predictor of both long and short horizons (and without
needing \emph{a priori} specification of confidence quantiles),
$\sutranets$ could also serve as the basis for a general-purpose large
forecasting model (LFM), along the lines of LLMs like
GPT3~\cite{brown2020language}.  Like modern LLMs, LFMs could
potentially be used with little task-specific data and no fine-tuning,
and may have similarly-large environmental and financial
costs~\cite{bender2021dangers}.  As with LLMs, new techniques will be
needed to
detect~\cite{rudinger2018gender,sheng2019woman,nadeem2020stereoset},
document~\cite{mitchell2019model,gebru2021datasheets}, and
mitigate~\cite{gonen2019lipstick,ouyang2022training} unsafe usage and
other failure cases.

%% file: experiments.tex
\textbf{Training.} We implemented $\sutranets$ in
PyTorch~\cite{paszke2019pytorch}.  We use Adam~\cite{kingma2014adam}
(default $\beta_1$=$0.9$, $\beta_2$=$0.999$, $\epsilon$=$10^{-8}$),
early stopping, and a decaying learning rate ($\gamma$=$0.99$).  The
default RNN is a 1-layer LSTM with 64 hidden units; when using >1
layers (Table~\ref{tab:depth}), inter-layer
dropout~\cite{melis2017state} with $p_{drop}$=$0.001$ is applied.  For
every model/dataset pair, we tune weight decay and initial learning
rate over a $4{\times}4$ grid.

\textbf{Systems.} We evaluate the following systems.  $\naive$ and
$\seasonalnaive$ are non-probabilistic, untrained baselines, while the
$\ctofar$ variants follow the standard RNN ordering
(Fig.~\ref{fig:rollouts:standard}):
\begin{itemize}[leftmargin=7em]
\item[$\naive$:] At each horizon, repeats last-observed historical value~\cite{hyndman2018forecasting}.
\item[$\seasonalnaive$:] At each horizon, repeat previous value at same \emph{season}
  as horizon~\cite{hyndman2018forecasting}, e.g., same hour-of-day
  ($\seasonalnaived$) or hour-of-week ($\seasonalnaivew$).
\item[$\ctofar$:]  LSTM-based forecasting model~\cite{bergsma2022c2far}, configured as described above~(\S\ref{sec:method}).
\item[$\lags$:] $\ctofar$, but with extra features for previous
  hour-of-day values (C2F-encoded).
\item[$\dropout$:] $\ctofar$, but with $p_{drop}$=$0.2$ (and re-scaling~\cite{srivastava2014dropout}) applied to inputs during training.
\item[$\freqhier$:] High-frequency forecast generated conditional on
low-frequency values~(\S\ref{sec:method}).
\item[$\sutranets$:] Variations $\regularprevs$, $\regularnoprevs$,
$\backfillprevs$, $\backfillnoprevs$~(\S\ref{sec:method}).
\end{itemize}

\textbf{Datasets.} We evaluate on the following datasets, where we
note the number of $\sutranet$ sub-series, $K$, as well as lengths of
conditioning, $C$, and prediction, $P$, ranges:
\begin{itemize}[leftmargin=3.5em]
\item[$\azure$:] $K$=6, $C$=2016, $P$=288: 5-minutely usage of 2085 VM
  flavors, groupings (by tenant, etc.) in Azure cloud. Originally
  from~\cite{cortez2017resource}, aggregated here instead to 5-minute
  intervals.
\item[$\elec$:] $K$=6, $C$=168, $P$=168: Hourly electricity usage of 321
  customers~\cite{dua2017uci}, version from~\cite{lai2018modeling}.
\item[$\traffic$:] $K$=6, $C$=168, $P$=168: Hourly occupancy for 862 car
  lanes~\cite{dua2017uci}, version from~\cite{yu2016temporal}.
\item[$\wiki$:] $K$=7, $C$=91, $P$=91: Daily hit-count for 9535 pages, first used
  in~\cite{gasthaus2019probabilistic}, version from~\cite{gouttes2021probabilistic}.
\item[$\mnist$:] $K$=4, $C$=392, $P$=392: row-wise sequential pixel values
for 70,000 28$\times$28 digits~\cite{lecun1998mnist}.
\item[$\mnists$:] $K$=4, $C$=392, $P$=392: $\mnist$ sequences, but with a fixed, random permutation applied.
\end{itemize}

Sequential (binarized) MNIST has been used for
NLL~\cite{lamb2016professor} and classification~\cite{le2015simple}
evaluations.  \citet{le2015simple} used a fixed permutation ``to make
the task even harder.''  Here, we forecast a distribution over
\emph{real-valued} pixel intensities in the second half of an image, based on
values in the first.  Unlike the other datasets, train/dev/test splits
are not based on time, rather we use the standard test
digits~\cite{lecun1998mnist}.

\textbf{Metrics.}  We evaluate the median
of our forecast distribution by computing \emph{normalized deviation}
(ND) from true values.
We evaluate the full distribution using \emph{weighted quantile loss}
  (wQL),
evaluated at forecast quantiles $\{0.1, 0.2 \ldots 0.9\}$, as
in~\cite{rabanser2020effectiveness,bergsma2022c2far}.  For non-probabilistic $\naive$
and $\seasonalnaive$ baselines, note wQL reduces to ND, similarly to
how CRPS reduces to absolute error~\cite[\S4.2]{gneiting2007strictly}.

%% file: results.tex
\textbf{Results.} We summarize the most important experimental
findings as follows:

\input{tab_main_results.tex}

\result{$\backfillprevs$ and $\backfillnoprevs$ improve over $\ctofar$
  and other baselines on every dataset, with both obtaining an average
  relative reduction in ND of around 15\% over the best
  non-$\sutranet$.}

Table~\ref{tab:main} has the main results; note the differences in
ND\% are very stable: we repeated \emph{grid search tuning and
testing} multiple times with different random seeds and
found \emph{$\sutranets$ superior to standard models across all
repeats} (supplemental \S\ref{supp:subsec:stability}).
$\backfillnoprevs$ does best when the training/inference discrepancy
problem predominates ($\azure$ \& $\mnist$, Table~\ref{tab:main}).
Adverse effects of discrepancy seem to be magnified by high
data \emph{redundancy}; in these cases, non-alternating generation
reduces discrepancy while having sufficient info to forecast
accurately.  We also hypothesize that $\regularnoprevs$ performs
poorer because it generates less-\emph{coherent} output: by
construction, it only conditions on past values, even when subsequent
values have already been generated (e.g., row 7 in
Fig.~\ref{fig:rollouts:regularnoprevs}).  In contrast, when generating
a given point (e.g., row 7 in
Fig.~\ref{fig:rollouts:backfillnoprevs}), $\backfillnoprevs$ has
always seen the previous and future proximal values, which were input
at the previous and current timesteps, respectively.
When signal path problems predominate ($\traffic$ \& $\mnists$),
alternating generation, with access to more info, performs better.
On $\mnists$, differences between regular and backfill depend solely
on the random permutation; the superiority of $\regularprevs$ here
motivates tuning over multiple orderings of sub-series, perhaps
averaging over them, as is done in some prior non-sequential
models~\cite{germain2015made,uria2016neural}.

Accuracy gains seem to derive from solving both discrepancy \emph{and}
signal path problems together.  When discrepancy predominates,
$\regularprevs$ performs poorly, even though signal path is improved.
For the opposite case, where error accumulation alone is improved, we
implemented an RNN that processes values in backfill order, but
otherwise acts as a standard RNN\@.  This model performs much poorer
than SutraNet-based $\backfillprevs$
(supplemental \S\ref{supp:subsec:bs}).  The lack of a holistic
solution may also explain the inconsistency of lags and dropout, which
sometimes help, sometimes hurt compared to vanilla $\ctofar$.
$\freqhier$ works best on datasests where discrepancy predominates,
while performing much poorer when signal path is important.
Note that weekly-seasonal baselines have not previously been evaluated
on these datasets.  Observing in particular the relatively strong
accuracy of $\seasonalnaivew$ on $\traffic$, we suggest they be
included in future evaluations.

\input{forecasts_fig.tex}

\input{images_fig.tex}

\result{$\sutranets$ are able to effectively model both long and
  short-term dependencies.}

On $\elec$ data, where $\seasonalnaived$ is more effective than
longer-term $\seasonalnaivew$ (Table~\ref{tab:main}), modeling
long-term dependencies remains important (Fig.~\ref{fig:forecasts}).
$\backfillprevs$ is able to capture both short-term changes and
long-term seasonality, the latter of which is neglected by $\ctofar$.
In plots of the $\mnist$ forecasts (Fig.~\ref{fig:images}), we see
that compared to $\ctofar$, $\backfillnoprevs$ generates more non-zero
values in the bottom parts of digits 0, 2, 5, etc., e.g., at p50\%,
implying the networks have more certainty or \emph{confidence} (i.e.,
tighter prediction intervals) on the digit being generated.
Also, Table~\ref{tab:main} results indicate that discrepancy is the
predominant problem on $\mnist$; this may help explain why
autoregressive models like PixelRNN~\cite{van2016pixel} score very
high in likelihood-based evaluation (where there is no sampling), but
produce less-realistic samples compared to GANs.  $\sutranets$ may
therefore provide a useful tool for reducing training/inference
discrepancy in a variety of image generators.

\result{$\sutranets$ are more accurate than $\ctofar$ across all
  forecast horizons.}

Autoregressive models are normally thought to accumulate and
compound~\cite{neill2018analysing} errors ``along the generated
sequence''~\cite{bengio2015scheduled}.  We therefore expect the
accuracy gap between $\sutranets$ and $\ctofar$ to grow across
horizons.  Instead, on $\elec$ and $\traffic$, we already see a large
gap over the first 24 hours
(supplemental \S\ref{supp:subsec:horizon}).
Improvements in both informational asymmetry (\S\ref{sec:intro}) and
signal path evidently enable gains across all steps.
$\freqhier$, in contrast, is only accurate at horizons corresponding
to its low-freq forecast, while $\backfillprevs$ is slightly better at
horizons corresponding to its first sub-series.  This suggests errors
do ``accumulate'' to a small extent across sub-series autoregression.

\begin{table}[t]
\RawFloats
\input{tab_depth_results.tex}
\hfill
\input{tab_subseries_results.tex}
\end{table}

\input{resources_fig.tex}

\result{Deeper models enable improved long-sequence forecasting, for
  $\sutranets$ and $\ctofar$.}
In particular, going from 1 to 2 layers greatly improves accuracy on
$\traffic$ (Table~\ref{tab:depth}), a dataset where signal path is the
predominant problem.  Smaller improvements are also seen on $\elec$.
Since both LSTM stacking and use of $\sutranets$ increase complexity,
a natural question is which approach offers the best computational
tradeoff for a desired level of accuracy.  Fig.~\ref{fig:resources}
shows that for a given number of parameters --- and especially for a
given inference time --- $\sutranets$ are able to obtain better ND\%
(plots of memory usage and of training time are similar, see
supplemental).  $\sutranet$ speed and memory usage scale with the
complexity of each RNN rather than the number of sub-series RNNs
(\S\ref{sec:method}); they can consequently make use of $K{\times}$
more parameters without $K{\times}$ the cost.

\result{$\backfillprevs$ is robust across different configurations,
  including when we vary $K$.}

We also experimented with varying the number of sub-series.
For all $K$>$1$, $\sutranets$ work much better than standard $\ctofar$
(Table~\ref{tab:subseries}).
Interestingly, $K$=$12$ led to the highest accuracy on $\traffic$,
where the signal path problem predominates.
When $K$ does not divide evenly into the seasonal period,
non-alternating models have a missing info problem (e.g., they miss
predictive preceding values at the same hour-of-day).  Both
$\backfillprevs$ and $\regularprevs$ can access these values through
covariate features (cf.\ Fig.~\ref{fig:rollouts:backfillprevs}), and
hence perform relatively better when $K$=$7$ on $\elec$.
$\backfillprevs$ is relatively weaker at $K$=$7$ when there is less
redundancy ($\traffic$), although still superior to standard
$\ctofar$.

See the supplemental for further experimental details and results.
Code for SutraNets is available at
\url{https://github.com/huaweicloud/c2far_forecasting/wiki/SutraNets}.

%% file: tab_main_results.tex
\begin{table}[]
  \caption{ND\%, wQL\% across datasets, best results in
    \textbf{bold}. For $\mnist$/$\mnists$, previous \emph{row} value
    used for $\seasonalnaived$.  Poor results for $\ctofar$ relative
    to $\regularprevs$ indicate \emph{signal path problem}
    (\hlc[green!50]{green} cells).  Poor results for both $\ctofar$
    and $\regularprevs$ indicate \emph{discrepancy problem}
    (\hlc[pink]{pink}).  Non-alternating weakness indicates
    \emph{missing info problem} (\hlc[yellow]{yellow}).
    $\backfillprevs$ reduces signal path and discrepancy without
    missing info, therefore offering best or near-best solution on
    each dataset.\label{tab:main}
    \mbox{}
    \vspace{-1mm}
    \mbox{}
  }
  \footnotesize
  \begin{tabular}{@{}lcccccc@{}}
    \toprule
    & $\azure$ & $\elec$    & $\traffic$ & $\wiki$    & $\mnist$     & $\mnists$    \\ \midrule
    $\naive$           & 3.2, 3.2 & 43.0, 43.0 & 75.5, 75.5 & 44.0, 44.0 & 100.0, 100.0 & 117.9, 117.9 \\
    $\seasonalnaived$  & 2.8, 2.8 & 10.5, 10.5 & 31.1, 31.1 & 44.0, 44.0 & 145.2, 145.2 & 163.5, 163.5 \\
    $\seasonalnaivew$  & 5.2, 5.2 & 11.1, 11.1 & 17.5, 17.5 & 41.8, 41.8 & N/A          & N/A            \\[0.6ex]
    $\ctofar$          & \hlc[pink]{3.2, 2.5} & \hlc[green!50]{10.6, 8.4}  & \hlc[green!50]{19.3, 16.0} & \hlc[green!50]{31.1, 27.2}  & \hlc[pink]{67.9, 52.3} & \hlc[green!50]{100.0, 95.7}       \\
    $\lags$            & 2.7, 2.2 & 10.4, 8.3  & 18.0, 15.0 & 31.2, 27.2 &  68.1, 52.8  & 99.3, 89.9 \\
    $\dropout$         & 3.2, 2.6 & 10.9, 8.7  & 18.5, 15.3 & 31.3, 27.0 &  68.6, 53.3  & 99.7, 90.4 \\[0.6ex]
    $\freqhier$        & 2.6, 2.1 & 10.3, 8.2  & 20.2, 16.6 & 30.9, 27.2 &  62.4, 47.6  & 98.1, 88.3 \\[0.6ex]
    $\regularprevs$    & \hlc[pink]{3.3, 2.6} & \hlc[green!50]{9.9, 7.9}   & \hlc[green!50]{15.5, 13.0} &  \hlc[green!50]{30.4, 26.4}  & \hlc[pink]{67.9, 52.4} & \textbf{\hlc[green!50]{48.7, 38.3}}          \\
    $\regularnoprevs$  & 2.6, 2.1 & 9.7, 7.8   & \hlc[yellow]{15.6, 13.1} & \hlc[yellow]{31.2, 27.3} & 59.4, 45.4   & \hlc[yellow]{90.1, 69.7}         \\[0.6ex]
    $\backfillprevs$   & 2.5, 2.0 & \textbf{9.3, 7.4} & \textbf{15.3, 12.8} & \textbf{30.1, 26.1} & 64.4, 49.7 & 72.0, 54.6 \\
    $\backfillnoprevs$ & \textbf{2.3, 1.9} & \textbf{9.3, 7.4} & \hlc[yellow]{15.7, 13.1} & \hlc[yellow]{30.5, 26.8} & \textbf{58.9, 44.9} & \hlc[yellow]{78.2, 58.3} \\ \bottomrule
  \end{tabular}
\end{table}

%% file: forecasts_fig.tex
\begin{figure}
  \centering {\makebox[\textwidth][c]{
      \begin{subfigure}{\shrinkfigthreeb\textwidth}
        \begin{tikzpicture}
          \node (img1) {\includegraphics[width=\textwidth]{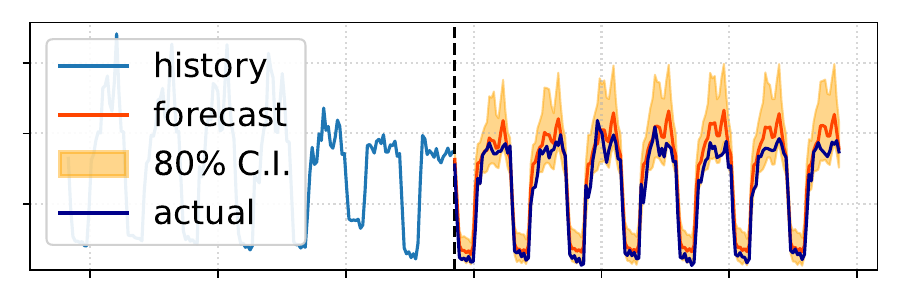}};
          \node[above of=img1, draw=none, align=center, fill=white, yshift=0.0cm, xshift=0.0cm, node distance=0.96cm, anchor=center] {\scriptsize \textbf{$\backfillprevs$}};
        \end{tikzpicture}
      \end{subfigure}
      \hspace{-2mm}
      \begin{subfigure}{\shrinkfigthreeb\textwidth}
        \begin{tikzpicture}
          \node (img1) {\includegraphics[width=\textwidth]{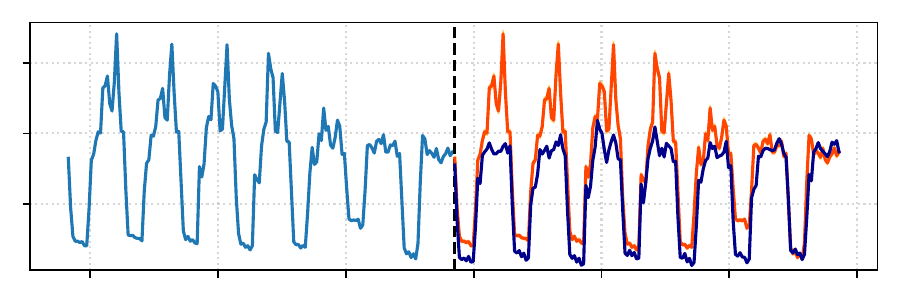}};
          \node[above of=img1, draw=none, align=center, fill=white, yshift=0.0cm, xshift=0.0cm, node distance=0.96cm, anchor=center] {\scriptsize \textbf{$\seasonalnaivew$}};
        \end{tikzpicture}
      \end{subfigure}
      \hspace{-2mm}
      \begin{subfigure}{\shrinkfigthreeb\textwidth}
        \begin{tikzpicture}
          \node (img1) {\includegraphics[width=\textwidth]{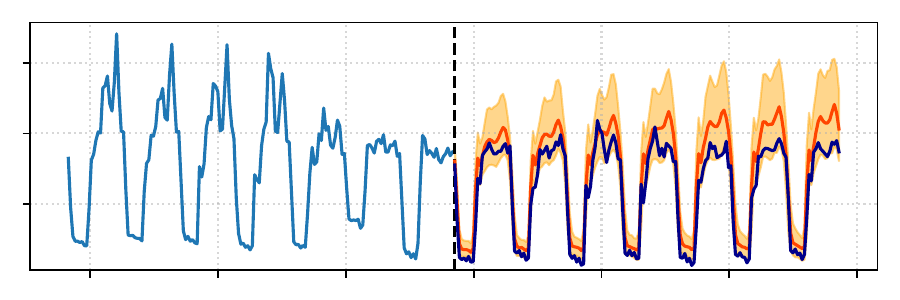}};
          \node[above of=img1, draw=none, align=center, fill=white, yshift=0.0cm, xshift=0.0cm, node distance=0.96cm, anchor=center] {\scriptsize \textbf{$\ctofar$}};
        \end{tikzpicture}
      \end{subfigure}
  }}
  \mbox{}
  \vspace{-3mm}
  \mbox{}
  \centering {\makebox[\textwidth][c]{
      \begin{subfigure}{\shrinkfigthreeb\textwidth}
        \begin{tikzpicture}
          \node (img1) {\includegraphics[width=\textwidth]{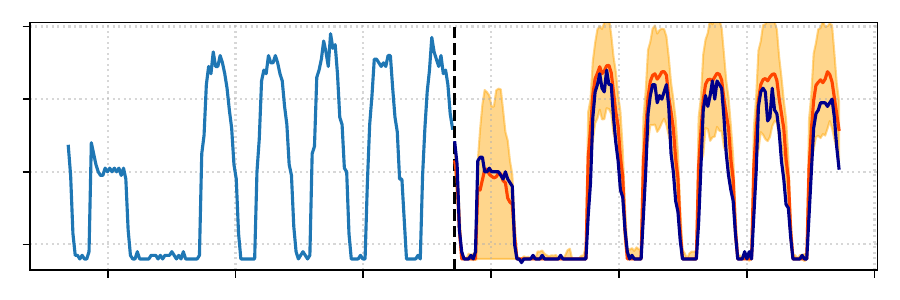}};
        \end{tikzpicture}
      \end{subfigure}
      \hspace{-2mm}
      \begin{subfigure}{\shrinkfigthreeb\textwidth}
        \begin{tikzpicture}
          \node (img1) {\includegraphics[width=\textwidth]{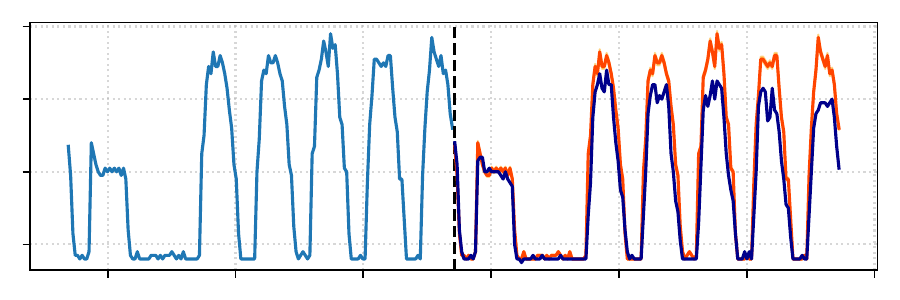}};
        \end{tikzpicture}
      \end{subfigure}
      \hspace{-2mm}
      \begin{subfigure}{\shrinkfigthreeb\textwidth}
        \begin{tikzpicture}
          \node (img1) {\includegraphics[width=\textwidth]{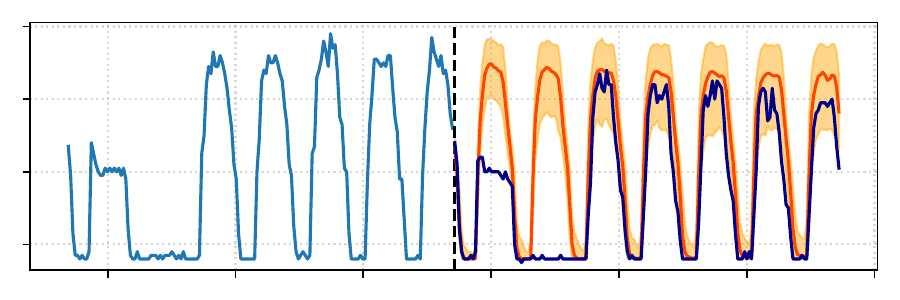}};
        \end{tikzpicture}
      \end{subfigure}
  }}
  \mbox{}
  \vspace{-3mm}
  \mbox{}
  \caption{Forecasts and 80\% intervals for $\elec$ series.
    $\backfillprevs$ performs well with a changepoint (row~1, where
    $\seasonalnaivew$ fails) and with weekly seasonality (row~2, where
    $\ctofar$ fails).\label{fig:forecasts}}
\end{figure}
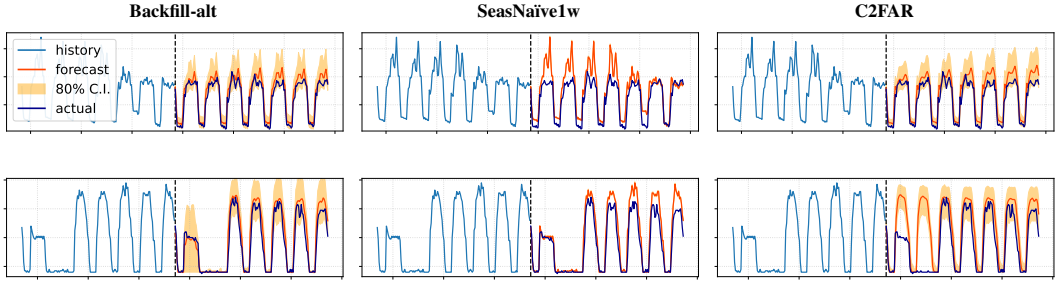

%% file: images_fig.tex
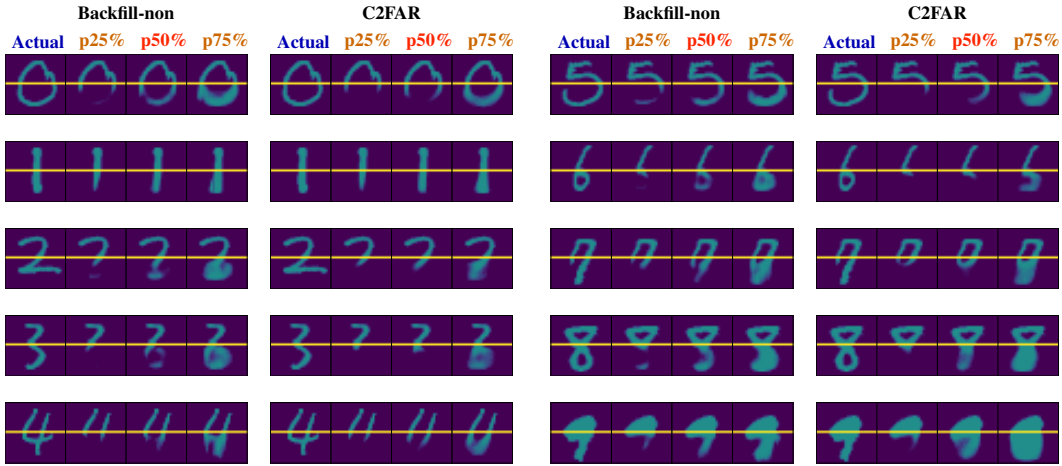
\begin{figure}
  \centering {\makebox[\textwidth][c]{
      \begin{subfigure}{\shrinkfigfour\textwidth}
        \begin{tikzpicture}
          \node (img1) {\includegraphics[width=\shrinkimg\textwidth]{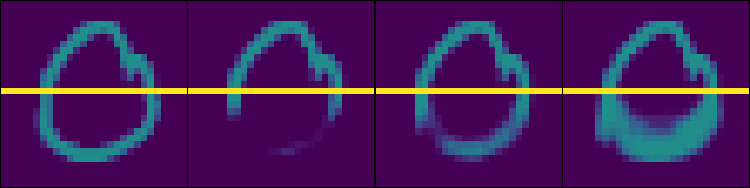}};
          \node[above of=img1, draw=none, align=center, fill=none, yshift=0.0cm, xshift=0.0cm, node distance=0.96cm, anchor=center] {\scriptsize \textbf{$\backfillnoprevs$}};
          \node[above of=img1, text=blue!70!black, draw=none, align=left, fill=none, yshift=0.0cm, xshift=-1.17cm, node distance=0.57cm, anchor=center] (lab1) {\scriptsize \textbf{Actual}};
          \node[right of=lab1, text=orange!80!black, draw=none, align=left, fill=none, yshift=0.0cm, xshift=0.0cm, node distance=0.86cm, anchor=center] (lab2) {\scriptsize \textbf{p25\%}};
          \node[right of=lab2, text=orange!30!red, draw=none, align=left, fill=none, yshift=0.0cm, xshift=0.0cm, node distance=0.82cm, anchor=center] (lab3) {\scriptsize \textbf{p50\%}};
          \node[right of=lab3, text=orange!80!black, draw=none, align=left, fill=none, yshift=0.0cm, xshift=0.0cm, node distance=0.82cm, anchor=center] (lab4) {\scriptsize \textbf{p75\%}};
        \end{tikzpicture}
      \end{subfigure}
      \hspace{0mm}
      \begin{subfigure}{\shrinkfigfour\textwidth}
        \begin{tikzpicture}
          \node (img1) {\includegraphics[width=\shrinkimg\textwidth]{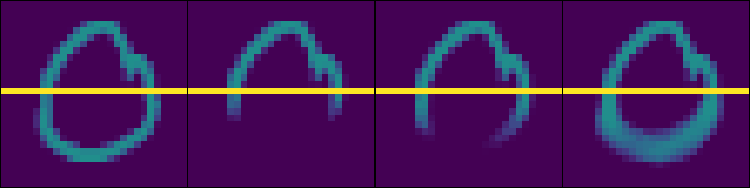}};
          \node[above of=img1, draw=none, align=center, fill=white, yshift=0.0cm, xshift=0.0cm, node distance=0.96cm, anchor=center] {\scriptsize \textbf{$\ctofar$}};
          \node[above of=img1, text=blue!70!black, draw=none, align=left, fill=none, yshift=0.0cm, xshift=-1.17cm, node distance=0.57cm, anchor=center] (lab1) {\scriptsize \textbf{Actual}};
          \node[right of=lab1, text=orange!80!black, draw=none, align=left, fill=none, yshift=0.0cm, xshift=0.0cm, node distance=0.86cm, anchor=center] (lab2) {\scriptsize \textbf{p25\%}};
          \node[right of=lab2, text=orange!30!red, draw=none, align=left, fill=none, yshift=0.0cm, xshift=0.0cm, node distance=0.82cm, anchor=center] (lab3) {\scriptsize \textbf{p50\%}};
          \node[right of=lab3, text=orange!80!black, draw=none, align=left, fill=none, yshift=0.0cm, xshift=0.0cm, node distance=0.82cm, anchor=center] (lab4) {\scriptsize \textbf{p75\%}};
        \end{tikzpicture}
      \end{subfigure}
      \hspace{2mm}
      \begin{subfigure}{\shrinkfigfour\textwidth}
        \begin{tikzpicture}
          \node (img1) {\includegraphics[width=\shrinkimg\textwidth]{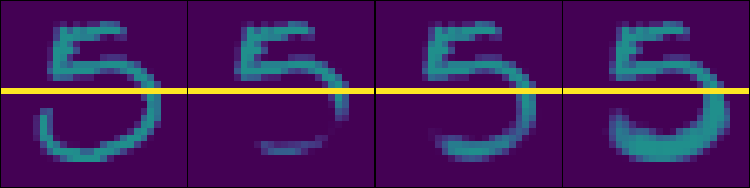}};
          \node[above of=img1, draw=none, align=center, fill=white, yshift=0.0cm, xshift=0.0cm, node distance=0.96cm, anchor=center] {\scriptsize \textbf{$\backfillnoprevs$}};
          \node[above of=img1, text=blue!70!black, draw=none, align=left, fill=none, yshift=0.0cm, xshift=-1.17cm, node distance=0.57cm, anchor=center] (lab1) {\scriptsize \textbf{Actual}};
          \node[right of=lab1, text=orange!80!black, draw=none, align=left, fill=none, yshift=0.0cm, xshift=0.0cm, node distance=0.86cm, anchor=center] (lab2) {\scriptsize \textbf{p25\%}};
          \node[right of=lab2, text=orange!30!red, draw=none, align=left, fill=none, yshift=0.0cm, xshift=0.0cm, node distance=0.82cm, anchor=center] (lab3) {\scriptsize \textbf{p50\%}};
          \node[right of=lab3, text=orange!80!black, draw=none, align=left, fill=none, yshift=0.0cm, xshift=0.0cm, node distance=0.82cm, anchor=center] (lab4) {\scriptsize \textbf{p75\%}};
        \end{tikzpicture}
      \end{subfigure}
      \hspace{0mm}
      \begin{subfigure}{\shrinkfigfour\textwidth}
        \begin{tikzpicture}
          \node (img1) {\includegraphics[width=\shrinkimg\textwidth]{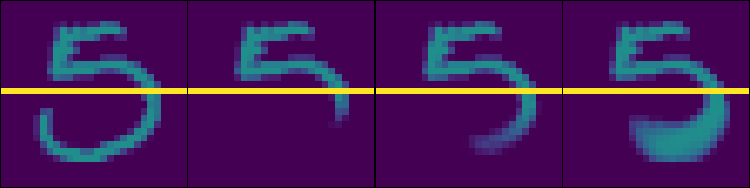}};
          \node[above of=img1, draw=none, align=center, fill=white, yshift=0.0cm, xshift=0.0cm, node distance=0.96cm, anchor=center] {\scriptsize \textbf{$\ctofar$}};
          \node[above of=img1, text=blue!70!black, draw=none, align=left, fill=none, yshift=0.0cm, xshift=-1.17cm, node distance=0.57cm, anchor=center] (lab1) {\scriptsize \textbf{Actual}};
          \node[right of=lab1, text=orange!80!black, draw=none, align=left, fill=none, yshift=0.0cm, xshift=0.0cm, node distance=0.86cm, anchor=center] (lab2) {\scriptsize \textbf{p25\%}};
          \node[right of=lab2, text=orange!30!red, draw=none, align=left, fill=none, yshift=0.0cm, xshift=0.0cm, node distance=0.82cm, anchor=center] (lab3) {\scriptsize \textbf{p50\%}};
          \node[right of=lab3, text=orange!80!black, draw=none, align=left, fill=none, yshift=0.0cm, xshift=0.0cm, node distance=0.82cm, anchor=center] (lab4) {\scriptsize \textbf{p75\%}};
        \end{tikzpicture}
      \end{subfigure}
  }}
  \mbox{}
  \vspace{-3mm}
  \mbox{}
  \centering {\makebox[\textwidth][c]{
      \begin{subfigure}{\shrinkfigfour\textwidth}
        \begin{tikzpicture}
          \node (img1) {\includegraphics[width=\shrinkimg\textwidth]{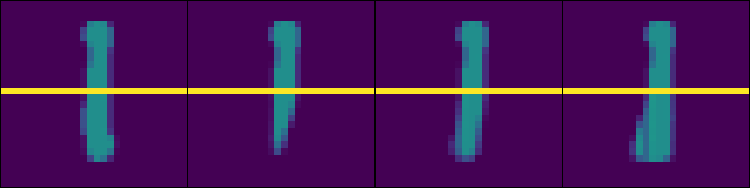}};
        \end{tikzpicture}
      \end{subfigure}
      \hspace{0mm}
      \begin{subfigure}{\shrinkfigfour\textwidth}
        \begin{tikzpicture}
          \node (img1) {\includegraphics[width=\shrinkimg\textwidth]{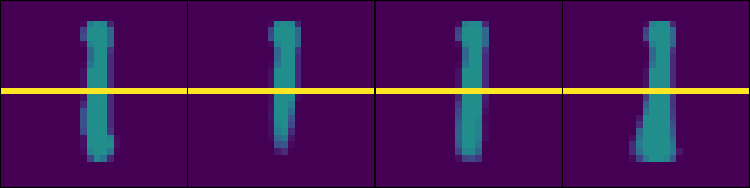}};
        \end{tikzpicture}
      \end{subfigure}
      \hspace{2mm}
      \begin{subfigure}{\shrinkfigfour\textwidth}
        \begin{tikzpicture}
          \node (img1) {\includegraphics[width=\shrinkimg\textwidth]{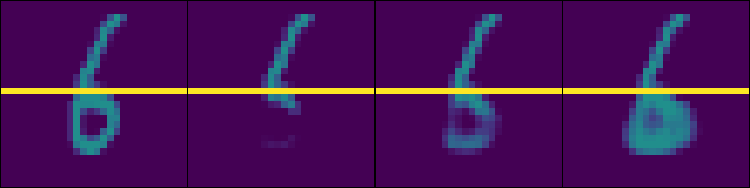}};
        \end{tikzpicture}
      \end{subfigure}
      \hspace{0mm}
      \begin{subfigure}{\shrinkfigfour\textwidth}
        \begin{tikzpicture}
          \node (img1) {\includegraphics[width=\shrinkimg\textwidth]{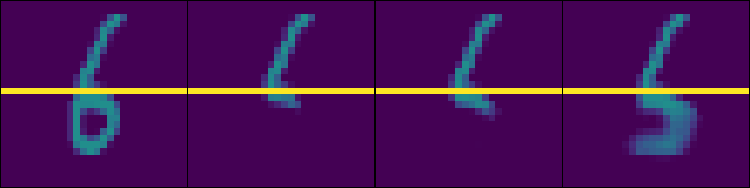}};
        \end{tikzpicture}
      \end{subfigure}
  }}
  \mbox{}
  \vspace{-3mm}
  \mbox{}
  \centering {\makebox[\textwidth][c]{
      \begin{subfigure}{\shrinkfigfour\textwidth}
        \begin{tikzpicture}
          \node (img1) {\includegraphics[width=\shrinkimg\textwidth]{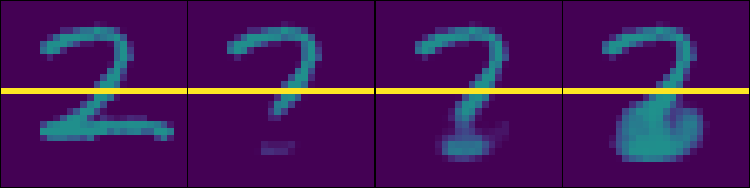}};
        \end{tikzpicture}
      \end{subfigure}
      \hspace{0mm}
      \begin{subfigure}{\shrinkfigfour\textwidth}
        \begin{tikzpicture}
          \node (img1) {\includegraphics[width=\shrinkimg\textwidth]{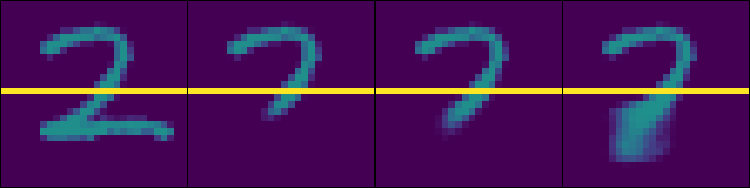}};
        \end{tikzpicture}
      \end{subfigure}
      \hspace{2mm}
      \begin{subfigure}{\shrinkfigfour\textwidth}
        \begin{tikzpicture}
          \node (img1) {\includegraphics[width=\shrinkimg\textwidth]{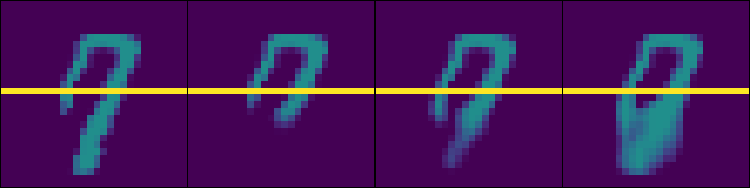}};
        \end{tikzpicture}
      \end{subfigure}
      \hspace{0mm}
      \begin{subfigure}{\shrinkfigfour\textwidth}
        \begin{tikzpicture}
          \node (img1) {\includegraphics[width=\shrinkimg\textwidth]{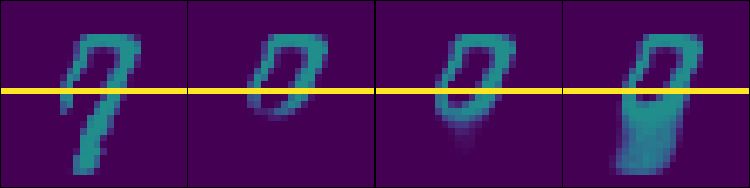}};
        \end{tikzpicture}
      \end{subfigure}
  }}
  \mbox{}
  \vspace{-3mm}
  \mbox{}
  \centering {\makebox[\textwidth][c]{
      \begin{subfigure}{\shrinkfigfour\textwidth}
        \begin{tikzpicture}
          \node (img1) {\includegraphics[width=\shrinkimg\textwidth]{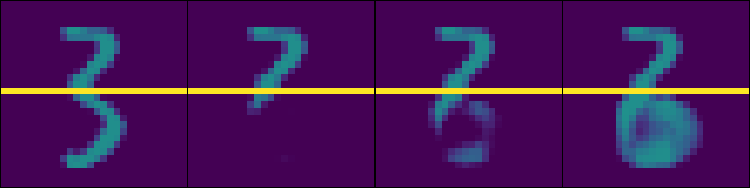}};
        \end{tikzpicture}
      \end{subfigure}
      \hspace{0mm}
      \begin{subfigure}{\shrinkfigfour\textwidth}
        \begin{tikzpicture}
          \node (img1) {\includegraphics[width=\shrinkimg\textwidth]{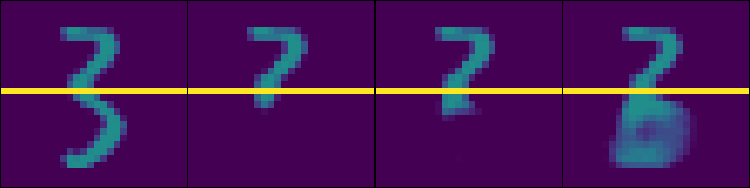}};
        \end{tikzpicture}
      \end{subfigure}
      \hspace{2mm}
      \begin{subfigure}{\shrinkfigfour\textwidth}
        \begin{tikzpicture}
          \node (img1) {\includegraphics[width=\shrinkimg\textwidth]{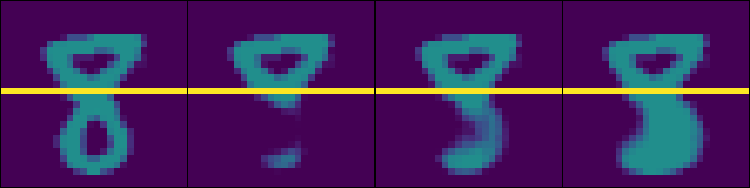}};
        \end{tikzpicture}
      \end{subfigure}
      \hspace{0mm}
      \begin{subfigure}{\shrinkfigfour\textwidth}
        \begin{tikzpicture}
          \node (img1) {\includegraphics[width=\shrinkimg\textwidth]{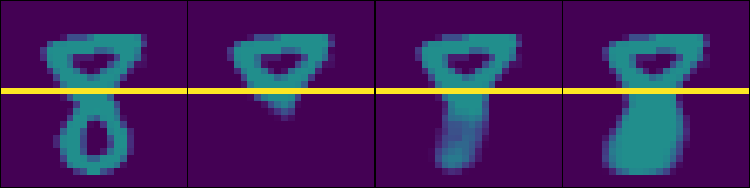}};
        \end{tikzpicture}
      \end{subfigure}
  }}
  \mbox{}
  \vspace{-3mm}
  \mbox{}
  \centering {\makebox[\textwidth][c]{
      \begin{subfigure}{\shrinkfigfour\textwidth}
        \begin{tikzpicture}
          \node (img1) {\includegraphics[width=\shrinkimg\textwidth]{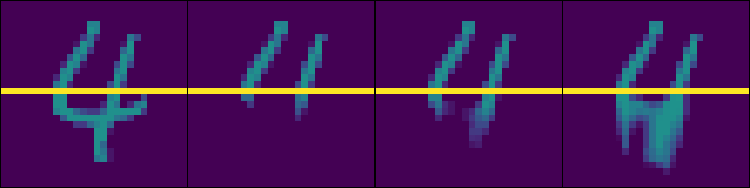}};
        \end{tikzpicture}
      \end{subfigure}
      \hspace{0mm}
      \begin{subfigure}{\shrinkfigfour\textwidth}
        \begin{tikzpicture}
          \node (img1) {\includegraphics[width=\shrinkimg\textwidth]{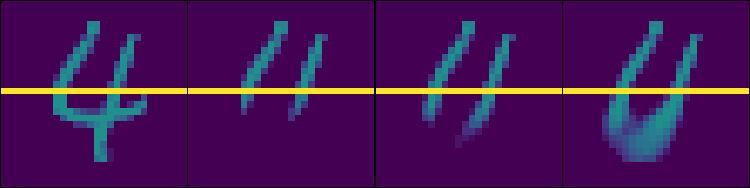}};
        \end{tikzpicture}
      \end{subfigure}
      \hspace{2mm}
      \begin{subfigure}{\shrinkfigfour\textwidth}
        \begin{tikzpicture}
          \node (img1) {\includegraphics[width=\shrinkimg\textwidth]{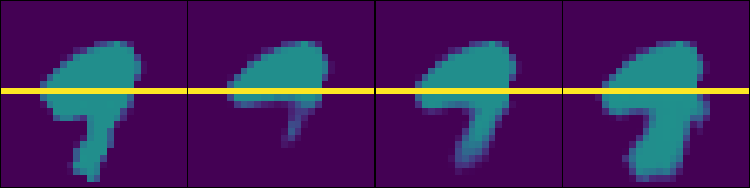}};
        \end{tikzpicture}
      \end{subfigure}
      \hspace{0mm}
      \begin{subfigure}{\shrinkfigfour\textwidth}
        \begin{tikzpicture}
          \node (img1) {\includegraphics[width=\shrinkimg\textwidth]{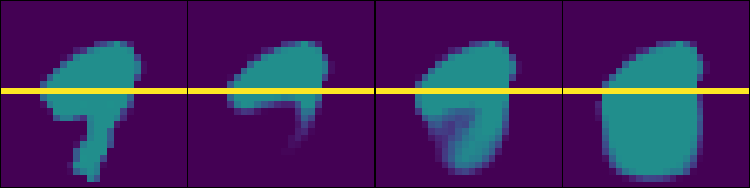}};
        \end{tikzpicture}
      \end{subfigure}
  }}
  \mbox{}
  \vspace{-1mm}
  \mbox{}
  \caption{Percentiles of forecast distributions for first occurrence
    of 0 \ldots 9 in $\mnist$ test data.  Pixels before yellow line
    (in row-order) are history.  $\backfillnoprevs$ more confident in
    bottom part of digits.\label{fig:images}}
\end{figure}

%% file: tab_depth_results.tex
%\begin{table}[]
\begin{minipage}{.545\linewidth}
  \centering
  \begin{minipage}{.96\linewidth}
    \centering
    \caption{ND\% decreases for $\sutranets$ \& $\ctofar$ as number of
      RNN layers/hidden units increases.\label{tab:depth}} \footnotesize
    \begin{tabular}{@{}l@{\hspace{2mm}}c@{\hspace{1.4mm}}c@{\hspace{1.4mm}}c@{\hspace{1.4mm}}cc@{\hspace{1.4mm}}c@{\hspace{1.4mm}}c@{\hspace{1.4mm}}c@{}}
      \toprule
      Dataset            & \multicolumn{4}{c}{$\elec$} & \multicolumn{4}{c}{$\traffic$}   \\
      \#layers           & 1     & 2            & 3   &   4 &  1     &   2  &    3 &   4 \\
      \#hidden           & 64    & 128          & 256 & 256 & 64     & 128  &  256 & 256 \\ \midrule
      $\ctofar$          & 10.6  & 10.2         & 9.9 & 9.7 & 19.3   & 14.7 & 14.8 &  14.4 \\
      $\regularprevs$    &  9.9  & 9.3 & \textbf{9.0} & 9.2 & 15.5   & 14.3 & 14.3 &  14.2 \\
      $\backfillprevs$   & {9.3} & 9.3 & \textbf{9.0} & 9.3 & 15.3   & 14.2 & 14.1 & \textbf{14.0} \\
      $\backfillnoprevs$ & {9.3} & 9.1 &         9.5  & 9.3 & 15.7   & 14.4 & 14.4 &  14.4 \\ \bottomrule
    \end{tabular}
  \end{minipage}
\end{minipage}
%\end{table}

%% file: tab_subseries_results.tex
%\begin{table}[]
\begin{minipage}{.455\linewidth}
  \centering
  \begin{minipage}{.96\linewidth}
    \centering
    \mbox{}
    \vspace{-1mm}
    \mbox{}
    \caption{ND\% lowest when \#sub-series divides into seasonal
      period (24), but alternating models less sensitive on $\elec$
      (cf.  $\ctofar$ with 10.6\% on $\elec$, 19.3\% on
      $\traffic$.)\label{tab:subseries}} \footnotesize
    \begin{tabular}{@{}l@{\hspace{2mm}}c@{\hspace{1.4mm}}c@{\hspace{1.4mm}}cc@{\hspace{1.4mm}}c@{\hspace{1.4mm}}c@{}}
      \toprule
      Dataset       & \multicolumn{3}{c}{$\elec$} & \multicolumn{3}{c}{$\traffic$} \\
      \#subseries   & 6      & 7       & 12    & 6       & 7       & 12      \\ \midrule
      $\regularprevs$  & 9.9    & 10.0     & 10.0   & 15.5    & 17.2    & 15.0    \\
      $\backfillprevs$  & \textbf{9.3}    & 9.4     & 9.4   & 15.3    & 16.9    & \textbf{14.9}    \\
      $\backfillnoprevs$  & \textbf{9.3}    & 10.0    & 9.5   & 15.7    & 17.3    & 15.4    \\ \bottomrule
    \end{tabular}
  \end{minipage}
\end{minipage}
%\end{table}

%% file: resources_fig.tex
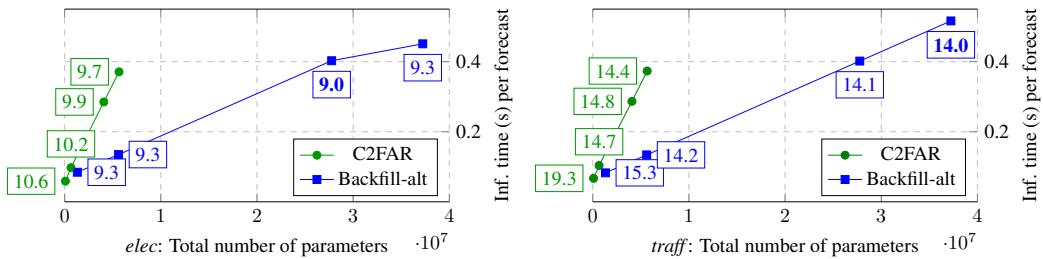
\begin{figure}
  \centering
      {\makebox[\textwidth][c]{
          \begin{subfigure}{\shrinkfigtwo\textwidth}
            \centering
            \scalebox{\resourcesshrink}{
              {\input{tikz_figures/resources_tikz_elec.tex}}
            }
          \end{subfigure}
      \hspace{0mm}
          \begin{subfigure}{\shrinkfigtwo\textwidth}
            \centering
            \scalebox{\resourcesshrink}{
              {\input{tikz_figures/resources_tikz_traffic.tex}}
            }
          \end{subfigure}
      }}
      \mbox{}
      \vspace{-1mm}
      \mbox{}
      \caption{Total parameters (x-axis) and average inference time
        per forecast (y-axis, lower better) for $\ctofar$ and
        $\backfillprevs$ systems of Table~\ref{tab:depth}, for $\elec$
        (left) and $\traffic$ (right).  ND\% labeled at each point in
        \boxed{$\mbox{boxes}$}.  $\backfillprevs$ enables more
        parameters and better ND\% at equivalent
        speeds.\label{fig:resources}}
\end{figure}

%% file: tikz_figures/resources_tikz_elec.tex
\begin{tikzpicture}
\begin{axis}[
    xlabel={$\elec$: Total number of parameters},
    grid=major,grid style={dashed},
    ylabel={Inf. time (s) per forecast},
    xmin=0, xmax=40000000,
    ymin=0, ymax=55,    
    xtick={0,10000000,20000000,30000000,40000000},    
    ytick={20,40,60},    
    % Note, we just changed the LABELS to make this time per forecast,
    % the data is still time per *100* forecasts
    yticklabels={0.2,0.4,0.6},
    separate axis lines,
    clip mode=individual,    
    width=8.4cm,    
    height=5cm,
    axis y line*=right,
    y axis line style= { draw opacity=0 },    
    yticklabel pos=right,
    ylabel near ticks,
    legend pos=south east
    ]

\addplot[black!40!green, mark=*] coordinates {
    (81830,5.890625002)
    (658214,9.75)
    (4075046,28.49999999)
    (5654054,37.09375)
};

\addplot[blue, mark=square*] coordinates {
    (1320420,8.437499998)
    (5608164,13.48437501)
    (27768036,40.265625)
    (37242084,45.09375)
};

\legend{C2FAR,Backfill-alt}

\newcommand{\fboxgreen}{\fcolorbox{black!40!green}{white}}
\newcommand{\fboxblue}{\fcolorbox{blue}{white}}

\node[circle,draw=black!40!green,inner sep=1pt,label={[black!40!green]left:\fboxgreen{10.6}}] at (axis cs:81830,5.890625002) {};
\node[circle,draw=black!40!green,inner sep=1pt,label={[black!40!green]above:\fboxgreen{10.2}}] at (axis cs:658214,9.75) {};
\node[circle,draw=black!40!green,inner sep=1pt,label={[black!40!green]left:\fboxgreen{9.9}}] at (axis cs:4075046,28.49999999) {};
\node[circle,draw=black!40!green,inner sep=1pt,label={[black!40!green]left:\fboxgreen{9.7}}] at (axis cs:5654054,37.09375) {};

\node[circle,draw=blue,inner sep=1pt,label={[blue]right:\fboxblue{9.3}}] at (axis cs:1320420,8.437499998) {};
\node[circle,draw=blue,inner sep=1pt,label={[blue]right:\fboxblue{9.3}}] at (axis cs:5608164,13.48437501) {};
\node[circle,draw=blue,inner sep=1pt,label={[blue]below:\fboxblue{\textbf{9.0}}}] at (axis cs:27768036,40.265625) {};
\node[circle,draw=blue,inner sep=1pt,label={[blue]below:\fboxblue{9.3}}] at (axis cs:37242084,45.09375) {};

\end{axis}
\end{tikzpicture}

%% file: tikz_figures/resources_tikz_traffic.tex
\begin{tikzpicture}
\begin{axis}[
    xlabel={$\traffic$: Total number of parameters},
    grid=major,grid style={dashed},
    ylabel={Inf. time (s) per forecast},
    xmin=0, xmax=40000000,
    ymin=0, ymax=55,    
    xtick={0,10000000,20000000,30000000,40000000},    
    ytick={20,40,60},    
    % Note, we just changed the LABELS to make this time per forecast,
    % the data is still time per *100* forecasts
    yticklabels={0.2,0.4,0.6},
    separate axis lines,
    clip mode=individual,    
    width=8.4cm,    
    height=5cm,
    axis y line*=right,
    y axis line style= { draw opacity=0 },    
    yticklabel pos=right,
    ylabel near ticks,
    legend pos=south east
    ]

\addplot[black!50!green, mark=*] coordinates {
    (81830,6.703125005)
    (658214,10.359375)
    (4075046,28.671875)
    (5654054,37.32812499)
};

\addplot[blue, mark=square*] coordinates {
    (1320420,8.265625007)
    (5608164,13.390625)
    (27768036,40.171875)
    (37242084,51.62500001)
};

\legend{C2FAR,Backfill-alt}

\newcommand{\fboxgreen}{\fcolorbox{black!40!green}{white}}
\newcommand{\fboxblue}{\fcolorbox{blue}{white}}

\node[circle,draw=black!40!green,inner sep=1pt,label={[black!40!green]left:\fboxgreen{19.3}}] at (axis cs:81830,6.703125005) {};
\node[circle,draw=black!40!green,inner sep=1pt,label={[black!40!green]above:\fboxgreen{14.7}}] at (axis cs:658214,10.359375) {};
\node[circle,draw=black!40!green,inner sep=1pt,label={[black!40!green]left:\fboxgreen{14.8}}] at (axis cs:4075046,28.671875) {};
\node[circle,draw=black!40!green,inner sep=1pt,label={[black!40!green]left:\fboxgreen{14.4}}] at (axis cs:5654054,37.32812499) {};

\node[circle,draw=blue,inner sep=1pt,label={[blue]right:\fboxblue{15.3}}] at (axis cs:1320420,8.265625007) {};
\node[circle,draw=blue,inner sep=1pt,label={[blue]right:\fboxblue{14.2}}] at (axis cs:5608164,13.390625) {};
\node[circle,draw=blue,inner sep=1pt,label={[blue]below:\fboxblue{14.1}}] at (axis cs:27768036,40.171875) {};
\node[circle,draw=blue,inner sep=1pt,label={[blue]below:\fboxblue{\textbf{14.0}}}] at (axis cs:37242084,51.62500001) {};

\end{axis}
\end{tikzpicture}

%% file: conclusion.tex
Probabilistic forecasting of long time series is an under-studied
problem, with most recent work focusing on point prediction.
We presented $\sutranets$, a novel autoregressive approach to
probabilistic forecasting of long-sequence time series.
$\sutranets$ convert a univariate series into a $K$-dimensional
multivariate series, each dimension comprising a distinct
every-$K$th-value sub-series of the original sequence.
The $\sutranet$ autoregressive model generates each sub-series
conditional on both its own prior values, and on other sub-series,
ensuring coherent output samples.  From these samples, an estimate of
the full joint probability of future values is made.
$\sutranets$ leverage $\ctofar$ under-the-hood, allowing for efficient
representation of time series amplitudes, and low-overhead input
encoding of covariate sub-series.
$\sutranets$ provide a holistic solution to challenges in
long-sequence sampling, reducing training/inference discrepancy (by
generating conditional on longer-term information) and signal path
distances (by a factor of $K$).
Experimentally, $\sutranets$ achieve state-of-the-art results on a
variety of long-sequence forecasting data sets, while demonstrating
similar speed and memory requirements as standard sequence models,
even when using $K{\times}$ more parameters.

%% file: supplemental.tex
\appendix

\section{Experimental details}

\subsection{Architecture}

As mentioned in \S\ref{sec:method} of the main paper, we use
$\ctofar$-LSTMs~\cite{bergsma2022c2far} for our sub-series RNNs.  The
detailed architecture of the LSTMs follows the description
in~\cite[supplement]{bergsma2022c2far}.  In particular, network
layers include \emph{bias weights}, and the same number of hidden
units are used in each LSTM layer when multi-layer LSTMs are used.
Like $\ctofar$, we also follow DeepAR~\cite{salinas2020deepar} in
using the same network to encode (i.e., process the conditioning
range) and decode (i.e., generate values in the prediction range).
Like $\ctofar$, during training we only compute loss over the
prediction range.

\subsection{Form of output distribution and input encoding}

As mentioned in \S\ref{sec:method} of the main paper, when
training DeepAR-style models, values in the conditioning and
prediction ranges are normalized based on the amplitudes in the
conditioning range.  Likewise, during inference, conditioning values
are normalized based on the conditioning range; forecasts are
subsequently made in the normalized space before they are ultimately
unnormalized in order to create the final output sample.
In $\sutranets$, we use min-max
scaling~\cite{rabanser2020effectiveness} to normalize values for a
target sub-series, based on the min and max of the conditiong range
\emph{of that sub-series}.  In other words, each sub-series is
forecast in its own normalized space. Theoretically, this could be
advantageous if the sub-series have very different amplitudes as it
would allow each sub-series to make use of the full range of bins in
the coarse-to-fine discretization, ultimately increasing the precision
of the forecasts.

However, also recall that for each sub-series RNN, at each step we
encode and provide as inputs both previous values of that sub-series
(the \emph{target} sub-series), as well as covariate features from
\emph{other} sub-series (autoregressively).  We therefore have two
options for normalizing the covariate values from the other
sub-series: (1)~normalize these values according to the dynamic range
in the conditioning range of the \emph{target} sub-series, or
(2)~normalize these values according to the dynamic range of
\emph{their own} conditioning range.  In preliminary experiments on
validation data, we found the former approach to be slightly more
effective, so adopt this approach with $\sutranets$.
The advantage of target-specific normalization is that covariate
sub-series values are always normalized \emph{consistently} with the
target sub-series.  The disadvantage is that any covariate sub-series
that has very different amplitudes than the target could become
normalized to very high or very low values; this could result in the
discretization of all such values to only a few bins, and therefore
only very coarse information from the covariate series would be
conveyed by the covariate features.
In future work, we plan to investigate this issue further and ascertain
whether other normalization strategies could prove more effective in
certain cases.

\subsection{Training}

{\input{supplemental_files/fixed.tex}}

\subsubsection{Slicing of training windows}\label{subsubsec:slicing}

As mentioned in \S\ref{sec:method} of the main paper, recall that
$\sutranets$, like other autoregressive forecasting models, are
trained by slicing many training series into many \emph{windows},
i.e., conditioning+prediction ranges at different start points.
During training, we randomly select windows for training batches
\emph{without replacement}, until all such windows have been
exhausted, at which point we repeat the random slicing process.

These windows are clock-aligned in the sense that, for example, the
$i$th hourly value in a window is assumed to correspond to some hour
of the day (e.g., 2pm-3pm).  However, as noted in
Footnote~\ref{footnote:offsets} of the main paper, these windows
do not begin and end at \emph{fixed} hours of the day; one window may
begin at 2pm (and span two weeks) and the next may begin at 7am (and
again span two weeks).  This means each sub-series sequence model does
not learn hour-of-day-specific (or, more generally, season-specific)
patterns.

\subsection{Training parallelism}

We vectorize across multiple conditioning+prediction windows during
training.  The number of windows that we parallelize over is referred
to as the \emph{n\_train\_batch\_size} (currently set to 128, see
Table~\ref{tab:fixed}).  We evaluate 8192 windows during each training
checkpoint (\emph{n\_train\_ranges\_per\_checkpoint}), and train for a
maximum of 750 checkpoints (\emph{n\_max\_checkpoints}).

\subsection{Tuning}

{\input{supplemental_files/tuning.tex}}

As mentioned in \S\ref{sec:experiments} of the main paper, we
tune the hyperparameters weight decay and initial learning rate over a
$4{\times}4$ grid.  Tuning via grid search is commonly performed in
forecasting~\cite{zhou2021informer,zhu2021mixseq,rangapuram2021end,salinas2019high}.
The specific values used in our grid are given in
Table~\ref{tab:tuning}.
We tune directly for normalized deviation (ND) on validation data,
evaluating \emph{after every training checkpoint}.
ND evaluation requires running the Monte Carlo sampling procedure in
order to generate a forecast distribution (\S\ref{sec:supp_metrics});
we use the median of this forecast distribution as the point forecast
for evaluation.  We use \emph{n\_rollouts}=25 samples in the Monte
Carlo estimate, over a fixed validation set of
\emph{n\_validation\_set}=8192 prediction ranges.
We stop a tuning trial early if we see
\emph{n\_stop\_evals\_no\_improve} evaluations without a new top score
(currently set to 37, see Table~\ref{tab:fixed}).
As training times are roughly comparable for $\ctofar$ and
$\sutranets$ (Fig.~\ref{fig:supp_trainingtime}), total tuning cost
is similar for both approaches, as well as for $\lags$, $\dropout$,
and $\freqhier$.

\subsection{Evaluation}

We use 500 separate rollouts during the forecasting process on
held-out data (\emph{n\_rollouts}=500).
We compute rolling evaluations with a stride of 1, i.e., we forecast
and evaluate over overlapping prediction ranges, as
in~\cite{gouttes2021probabilistic}.

\subsection{Computational resources}

$\sutranets$ are implemented in PyTorch~\cite{paszke2019pytorch},
version \texttt{1.9.1+cu102}.
We use GPUs from Nvidia: four Tesla P100 GPUs with 16280MiB and two
Tesla K80 GPUs with 11441MiB.

\subsection{Datasets}

{\input{supplemental_files/dataset_raw.tex}}

{\input{supplemental_files/dataset_config.tex}}

\subsubsection{Azure VM demand dataset}\label{subsec:azure}

The $\azure$ dataset was first used for forecasting
in~\cite{bergsma2022c2far}; this work leveraged the publicly-available
\emph{Azure Public
  Dataset}\footnote{\url{https://github.com/Azure/AzurePublicDataset/blob/master/AzurePublicDatasetV1.md}},
originally released in~\cite{cortez2017resource} under a Creative
Commons Attribution 4.0 International Public License.
We converted the event stream in the Azure Public Dataset into time
series by exactly following the approach
in~\cite[supplement]{bergsma2022c2far}, except rather than aggregating
the data over a 1-hour period, we aggregated the data over a 5-minute
period.  This is actually the most precise aggregation possible given
the original dataset quantizes all timestamps using 5-minute
precision.
We also used the same experimental splits as
in~\cite{bergsma2022c2far}, using 20 days as training, 3 days for
validation, and 3 final days for testing.
Compared to hourly granularity, with 5-minute intervals, there are
$12{\times}$ as many rolling windows to evaluate, and each has
$12{\times}$ as many steps.  To alleviate the computational burden for
this dataset, we evaluate all systems on a random, fixed 53\% subset
of the 4.1M+ windows in the test period.

\subsubsection{Other datasets}

The $\elec$, $\traffic$, and $\wiki$ datasets were obtained using
scripts in GluonTS~\cite{alexandrov2020gluonts}.  Compared to the
training/validation/test splits used in prior
work~\cite{salinas2019high,rabanser2020effectiveness,gouttes2021probabilistic},
here we use more validation/test values (Table~\ref{tab:dataset_raw}),
reflecting the longer forecast horizons that we evaluate on.

The $\mnist$ dataset was obtained from~\cite{lecun1998mnist}.  The
standard 10,000 test images were used as our test set, while a random
10,000-element subset of the 60,000 training images were used as a
validation set.  The $\mnists$ dataset was obtained by applying a
fixed random permutation to every element of these same datasets.  The
pixel values were used in their original floating-point format, i.e.,
they were not binarized or modified in any way beyond the
\texttt{ToTensor()} transform in \texttt{torchvision}.
Also, note that even though $\mnist$ and $\traffic$ dynamic ranges are
actually bounded (between 0 and 1), we do not use this information
explicitly; that is, we read the conditioning ranges and normalize in
the same way as we do on all datasets.

Table~\ref{tab:dataset_raw} provides the details of these datasets and
$\azure$.  In Table~\ref{tab:dataset_config}, we note some
system-specific configurations for each dataset.  To help explain this
table, we now provide some background information.  First of all, note
that a $\ctofar$ binning always has a fixed extent from a low to a
high cutoff, over the min-max normalized
values~\cite{bergsma2022c2far}.
The binning extent is selected in order to cover from roughly the 1\%
to the 99\% percentiles of normalized values in the conditioning and
prediction ranges of training data for each series.  These values are
normalized, as noted above, using min-max values from the
\emph{conditioning} range; the prediction range can go below the min
and above the max.  Also as noted above, recall that in $\sutranets$,
each sub-series is normalized using the conditioning min and max
values \emph{of that sub-series}.  We see in
Table~\ref{tab:dataset_config} that this generally corresponds to a
wider binning range than with standard $\ctofar$, likely because there
is greater variance between conditioning and prediction ranges when
using the shorter sub-series sequences.
Table~\ref{tab:dataset_config} also gives the lag period used in
$\lags$.

\subsection{Metrics}\label{sec:supp_metrics}

Let $y_{i,t}$ be the $t$th value of the $i$th time series, that is,
$i$ indexes over time series and $t$ indexes over time steps.
Recall that autoregressive forecasting approaches create a Monte Carlo
estimate of the forecast distribution by repeatedly sequentially
sampling the model in the prediction range.
Forecast quantiles at a given horizon are then estimated by
calculating quantiles of the sampled forecasts at that horizon.
Let $\alpha$ represent the quantile of interest, e.g., $\alpha$=$0.5$
indicates we want quantile $0.5$, i.e., the $50$th percentile, while
$\alpha$=$0.9$ indicates the 90th percentile, etc.
Let $\hat{y}_{i,t}^{(q_\alpha)}$ be the actual estimated $\alpha$
quantile of the forecast distribution for time series $i$ at point
$t$, e.g. $\hat{y}_{i,t}^{(q_{0.9})}$ is the value such that 90\% of
possible values for point $y_{i,t}$ are expected to be below this
value (and, as noted above, we obtain this estimate from quantiles of
our Monte Carlo samples).
Our evaluation metrics always involve comparing an estimated quantile
of the forecast distribution, $\hat{y}_{i,t}^{(q_{\alpha})}$, at some
horizon of the forecast, to an observed true value at that horizon
$y_{i,t}$.
For normalized deviation (ND), we compare the $50$th percentile of the
forecast distribution to the observed true value, i.e., we use the
$50$th percentile as a \emph{point estimate}.  For wQL,
following~\cite{rabanser2020effectiveness,bergsma2022c2far}, we
compare multiple estimated quantiles (at $\alpha$=$\{0.1, 0.2 \ldots
0.9\}$), to the same observed true value.

More formally, let $\mathcal{I}(\cdot)$ denote the indicator function.
We define \emph{pinball loss} and \emph{quantile loss} as part of the
derivation of \emph{weighted quantile loss}.  \emph{Weighted quantile
  loss} and \emph{normalized deviation} are reported in the main
paper, as percentages.

\paragraph{Pinball loss:}

{\input{supplemental_files/equations/pinball_loss_eqn.tex}}

\paragraph{Quantile loss:}

{\input{supplemental_files/equations/QL_eqn.tex}}

\paragraph{Weighted quantile loss:}

{\input{supplemental_files/equations/wQL_eqn.tex}}

\paragraph{Normalized deviation:}

{\input{supplemental_files/equations/ND_eqn.tex}}

\section{Experimental results}

\subsection{Computational performance and resource requirements}

{\input{supplemental_files/memory_fig.tex}}

See Fig.~\ref{fig:resources} in the main paper for inference
\emph{time}, and Fig.~\ref{fig:supp_memory} here for inference
\emph{memory usage}, for the systems of Table~\ref{tab:depth} in
the main paper.  All measurements of speed and memory were made on
NVIDIA Tesla P100 GPUs, with a common test batch size of 32, and 500
Monte Carlo samples for the forecast distribution estimation.

{\input{supplemental_files/trainingtime_fig.tex}}

Fig.~\ref{fig:supp_trainingtime} has the training times for the
systems of Table~\ref{tab:depth} in the main paper.  Training
time naturally reflects both the speed of convergence in learning
(number of training epochs) and the speed of operating the specific
architecture.

{\input{supplemental_files/nparameters_table.tex}}

Exact numbers of parameters for the different models (and others that
have not been evaluated), are given in Table~\ref{tab:nparameters}.

Looking holistically at all of the performance figures, we can
conclude two things:
\begin{enumerate}
\item For the same inference time, memory usage, or training time,
  $\sutranets$ are typically more accurate than standard $\ctofar$
  (comparing along horizontal lines of each plot).
\item For the same number of parameters, $\sutranets$ are typically
  more accurate than standard $\ctofar$ (comparing along vertical
  lines of \emph{any} plot).
\end{enumerate}

The only exception to this rule is the two-level $\ctofar$ model on
$\traffic$ (scoring 14.7\%), which is superior to the 1-level
$\sutranet$ model (15.3\%) while using fewer parameters.  This model
is also arguably competitive with $\sutranets$ in terms of training
and inference time (but uses more memory than a more accurate
$\sutranet$ model at 14.2\%).  As noted in the main paper
(\S\ref{sec:experiments}), the signal path problem predominates
on the $\traffic$ dataset.  For this problem, it seems depth and
$\sutranets$ both offer effective strategies for improving accuracy
with minor increases in computational overhead.  With depth and
$\sutranets$ together, the 2-level $\sutranet$ already dominates the
deepest vanilla $\ctofar$ model along all performance dimensions,
while using fewer parameters.

It is also worth noting here that the $\regularprevs$ is a unique
$\sutranet$ in that the same sequence model parameters could
theoretically be used for each sub-series model.  That is, each
sub-series LSTM could continue to predict a distinct every-$K$th-value
of the full sequence, and each such LSTM could continue to evolve its
own distinct hidden state, while conditioning on its own unique
covariates.  However, each of these LSTMs could use the same trained
LSTM \emph{parameters}.  This is a consequence of the unique ordering
of $\regularprevs$ (and the fact that the starting points of
conditioning+prediction windows are not clock-aligned in training, as
noted in \S\ref{subsubsec:slicing}).
As such, a $\regularprevs$ LSTM that shares model parameters among its
different sub-series models could use only $\nicefrac{1}{K}$ of the
parameters compared to other $\sutranets$.

\subsection{Stability of empirical results}\label{subsec:stability}

In this section, we investigate the stability of our empirical
results.
Random seeds are used in both our training/tuning process (via random
sampling of windows for training batches, \S\ref{subsubsec:slicing})
and our testing process (via Monte Carlo sampling of predicted future
values).  It is important to quantify the stability of these sources
of randomness separately~\cite{clark2011better}.
Regarding our training/tuning process, ideally, for each system on
each dataset, we would repeat our entire grid search tuning procedure
multiple times with different random seeds, allowing us to determine
the end-to-end stability of our approach to model fitting.
While such repetition is not practical to perform over all datasets
and over all depth/subseries variations, given the total time required, we
elected to perform this procedure on $\elec$ and $\traffic$, in order
to get definitive quantification of stability on these two datasets,
and through these findings obtain a sense of the overall stability of
our experimental results.

{\input{supplemental_files/tuning_stability.tex}}

Fig.~\ref{fig:supp_tuning_stability} provides the tuning stability
results.  $\backfillprevs$ is remarkably stable across tuning runs on
both $\elec$ and $\traffic$, while $\ctofar$ shows much greater
variation.  In general, we find our tuning results to be very stable:
$\sutranets$ are superior to $\ctofar$ across all repeats.

{\input{supplemental_files/testing_stability.tex}}

Meanwhile, Fig.~\ref{fig:supp_testing_stability} provides the testing
stability results.  In evaluation, both $\backfillprevs$ and $\ctofar$
are extremely stable across different random seeds.

\subsection{Evaluation by forecast horizon}\label{subsec:horizon}

\input{supplemental_files/all_horizons_fig.tex}

Fig.~\ref{fig:supp_horizons} shows the forecast error of the systems
as a function of the forecast horizon, for all datasets.
These plots provide an interesting perspective on the
training/inference discrepancy versus signal path problems.  On
datasets where discrepancy predominates ($\azure$ and especially
$\mnist$, main Table~\ref{tab:main}), differences between
$\ctofar$ and $\sutranets$ do seem to start small and grow over time.
Meanwhile, when signal path problems predominate ($\traffic$ and
especially $\mnists$), differences between $\ctofar$ and $\sutranets$
are immediately large.
These observations are further evidence that $\sutranets$ provide both
a useful diagnostic for errors in long-sequence generation, and a
useful solution to these errors.

\subsection{Evaluation of a Backfill-Standard Model}\label{subsec:bs}

Experimental results in the main paper (\S\ref{sec:experiments})
clearly demonstrate improvements in forecasting accuracy when
SutraNets are applied in backfill order.  For example,
$\backfillprevs$ improves over $\regularprevs$ on each of $\azure$,
$\elec$, $\traffic$ and $\mnist$ datasets.
This raises an interesting question: could backfill ordering alone ---
i.e., used without SutraNets --- lead to improvements over standard
RNNs that process the values in regular order?

{\input{supplemental_files/backfill_standard_rollout.tex}}

To investigate this question, we implemented a normal RNN model, with
a single set of parameters and a single evolving RNN hidden state, but
where we step through the time series in backfill order in segments of
$K$ consecutive values.  That is, within blocks of $K$ values, we
visit the values in reverse order, and then move to the next block,
akin to reading a document downwards but from right-to-left on each
line.  The resulting state transitions and feature dependencies are
pictured in Fig.~\ref{fig:bs_rollouts}.  This is essentially the same
generative order as the $\backfillprevs$ model in
Fig.~\ref{fig:rollouts:backfillprevs} of the main paper, but
where a single state is updated at every value.  We call this the
$\backfillstandard$ model.

Comparing the diagrams of $\backfillstandard$ and $\backfillprevs$, it
is clear that signal path will not be improved by $\backfillstandard$.
However, because they both take $K$-step maximum \emph{generative
  strides}, both approaches may have similar improvements in error
accumulation.  We therefore hypothesize that $\backfillstandard$ may
improve accuracy over the standard RNN, but remain less accurate than
the $\backfillprevs$ SutraNet.
Such an improvement would be of significant practical importance,
since $\backfillstandard$ is very straightforward to implement, simply
requiring a kind of shuffling of the values in the conditioning and
generation windows (i.e., a simple pre-processing step), which could
be applied before using any standard sequence model.

{\input{supplemental_files/tab_bs_results.tex}}

To evaluate this question, we trained/tuned and tested
$\backfillstandard$ on three datasets, following the same experimental
setup as we used for the main paper evaluations.
Unfortunately,
$\backfillstandard$ performed quite poorly, worse than all SutraNets
and worse than the standard $\ctofar$ model
(Table~\ref{tab:bs_results}).
One key point about $\backfillstandard$ is that it not only does not
\emph{improve} signal path, but actually hurts it in many cases.  For
example, when generating value $y_3$ in row 5 of
Fig.~\ref{fig:bs_rollouts}, the key information about the value of
$y_2$ was provided many steps in the past (back in row 1, on the order
of $2K$ steps in the past).  The standard RNN, meanwhile, is always
provided the $(i-1)$ input directly when generating the $i$th output,
while $\backfillprevs$ has access to $y_2$ at only one step in the
past (as an input for the value generated at the previous step, see
Fig.~\ref{fig:rollouts:backfillprevs} in the main paper).
Results for standard $\ctofar$ and $\backfillstandard$ are closer on
$\elec$, where signal path is less of an issue, but $\ctofar$ still
performs better.
It seems the potential benefits in reducing error accumulation cannot
be realized without also improving signal path; if the model cannot
rely on longer-range information, it struggles to take accurate,
longer predictive strides.
These results provide further evidence that SutraNets are a useful
approach because they solve both error accumulation and signal path
together, and their benefits cannot be achieved with simple
preprocessing tricks.

\subsection{Experiments with $\deeparbinned$}\label{subsec:deeparbinned}

Experimental results in the main paper (\S\ref{sec:experiments})
show that SutraNets provide strong gains over the standard $\ctofar$
model.  In this section, we investigate whether a different
distributional estimator,
$\deeparbinned$~\cite{rabanser2020effectiveness}, also sees gains when
used with SutraNets.
To evaluate this question, we trained/tuned and tested both
$\deeparbinned$ alone, and $\deeparbinned$+$\backfillprevs$ SutraNets.
In both cases we use 1024 bins.  We use the same three datasets as in
\S\ref{subsec:bs} and again follow the same experimental setup as was
used in the main paper evaluations.

{\input{supplemental_files/tab_binned_results.tex}}

Results are in Table~\ref{tab:binned_results}.
First of all, note that we replicate the findings
of~\cite{bergsma2022c2far}, but in the long-sequence setting:
$\ctofar$ alone performs better than $\deeparbinned$ alone in all
cases (although ND\% is quite close on $\elec$).
Secondly, we observe that applying SutraNets via the $\backfillprevs$
ordering substantially improves $\deeparbinned$ on all datasets,
demonstrating the broad applicability of the SutraNet approach.
Finally, we note that SutraNets and the $\ctofar$ enhancement are
synergistic: applying both together results in the most accurate
forecasting models.
We also note that $\deeparbinned$ requires significantly more memory
than the $\ctofar$ system; $\deeparbinned$ uses 1024 output bins,
while $\ctofar$ uses only $12{\times}3$ output bins total via its
efficient hierarchical factorization.
Note the main computational bottleneck in tuning is sampling the
output rollouts in order to compute ND\% on the validation set.  Since
$\deeparbinned$ requires more memory, we must perform this sampling
over smaller batch sizes.  The net result is $\deeparbinned$ tuning
taking roughly 3X-4X more total GPU time to tune.
$\ctofar$ is therefore doubly synergistic with SutraNets: the
combination allows for accurate, coherent forecasts, and also reduces
the computational cost of these forecasts.

\subsection{Effects of shorter context windows}

In the main paper, we mentioned that many recent systems for
``long-term'' forecasting use conditionining windows with a maximum
length of 96 inputs, which amounts to \emph{4 days} of history at
1-hour granularity.  We explained that such restricted context is a
serious limitation given that values from \emph{7 days} in the past
are highly predictive on datasets with strong weekly seasonality, as
reflected in the good results for $\seasonalnaivew$ on $\elec$ and
$\traffic$ in main Table~\ref{tab:main}.

{\input{supplemental_files/tab_scaleformer_results.tex}}

Although Scaleformer with Transformers may be limited to short
conditioning windows, the core idea of Scaleformer can be applied with
RNNs; indeed, we implemented and evaluated a similar method to
Scaleformer, but for RNNs (and using C2FAR as the output
distribution), which we called $\freqhier$, as noted in main paper
\S\ref{sec:background}.
To illustrate the impact of decreased context windows, we compare the
accuracy of $\freqhier$ and SutraNets in our work with those in
Table~9 of \citet{shabani2022scaleformer}, where Informer and
Scaleformer were used to generate \emph{probabilistic} forecasts.
We provide this comparison in Table~\ref{tab:scaleformer_results}.

Note that SutraNets predict 168 steps ahead vs. 96 for Scaleformer and
Informer (which should disadvantage SutraNet results).  Also note
\citet{shabani2022scaleformer} evaluate using CRPS, while we evaluate
using wQL, a 10-point approximation to CRPS (for point predictors like
seasonal-naive-1week, note both CRPS and wQL reduce to normalized
deviation).
It is problematic to compare these systems directly, as they have been
trained and tuned in different ways, on different data splits, and
with different output distributions.  However, these results are
nevertheless suggestive of the large potential drawbacks of using
limited historical context.
We hope that prior systems may find ways to use larger amounts of
context and dramatically improve their forecasting accuracy.

%% file: supplemental_files/fixed.tex
\begin{table}
  \caption{Fixed hyperparameters.\label{tab:fixed}}
  \footnotesize
\begin{tabular}{@{}lcl@{}}
\toprule
Hyperparameter                                       & Value     & Note \\
\midrule
n\_train\_batch\_size                                & 128       & Total num\@. prediction ranges per training batch                 \\
n\_train\_ranges\_per\_checkpoint                    & 8192      & Total num\@. prediction ranges in one \emph{checkpoint} (train loss reported)        \\
n\_max\_checkpoints                                  & 750       & Maximum num\@. checkpoints (n\_train\_ranges\_per\_checkpoint sets)       \\
n\_rollouts (validation)                             & 25        & Num\@. samples for sampling when evaluating on validation set                          \\
n\_validation\_set                                   & 8192      & Total num\@. prediction ranges per validation evaluation \\
n\_stop\_evals\_no\_improve                          & 37        & Num\@. validation evals without improvement before early stop \\
n\_rollouts (test)                                   & 500       & Num\@. samples for sampling when evaluating on test set                                  \\
\bottomrule
\end{tabular}
\end{table}

%% file: supplemental_files/tuning.tex
\begin{table}
\caption{Tuning ranges for hyperparameter grid-search optimization.\label{tab:tuning}} \footnotesize
\begin{tabular}{@{}cc@{}}
\toprule
Hyperparameter                                    & Range            \\ \midrule
learning\_rate                                    & {[}1e-4, 1e-3, 1e-2, 1e-1{]}    \\
weight\_decay                                     & {[}1e-7, 1e-6, 1e-5, 1e-4{]}    \\
\bottomrule
\end{tabular}
\end{table}

%% file: supplemental_files/dataset_raw.tex
\begin{table}[]
\caption{Dataset details\label{tab:dataset_raw}.  $^{*}$Note: while
  other datasets use disjoint temporal periods for Dev and Test data,
  totally disjoint \emph{series} (MNIST images) are used for the Dev
  and Test sets of $\mnist$ and $\mnists$.}  \footnotesize
\begin{tabular}{@{}cccccccccc@{}}
\toprule
{Dataset} & {Domain} & {Freq} & {Num.}        & {Vals}   & {Dev}      & {Test}        & {Condit.} & {Pred.} \\
                 &                 &               & {series}      & {per}    & {vals per} & {vals per}    & {range}   & {range} \\
                 &                 &               &             & {series} & {series}   & {series}      & {size}    & {size}  \\ \midrule
$\elec$          & Discrete        & Hourly        & 321                  & 21212           & 504               & 504                  & 168              & 168            \\
$\traffic$       & Real            & Hourly        & 862                  & 14204           & 504               & 504                  & 168              & 168            \\
$\wiki$          & Discrete        & Daily         & 9535                 & 912             & 91                & 182                  & 91               & 91             \\
$\azure$         & Discrete        & 5-minute      & 4048                 & 8628            & 864               & 864                  & 2016             & 288            \\
$\mnist$         & Real            & 1 pixel       & 70000                & 784             & 784$^{*}$         & 784$^{*}$            & 392              & 392            \\
$\mnists$        & Real            & 1 pixel       & 70000                & 784             & 784$^{*}$         & 784$^{*}$            & 392              & 392            \\ \bottomrule
\end{tabular}
\end{table}

%% file: supplemental_files/dataset_config.tex
\begin{table}[]
\caption{Dataset-specific parameter settings for different systems.\label{tab:dataset_config}}
\footnotesize
\begin{tabular}{@{}cccccc@{}}
\toprule
{Dataset} & {$\ctofar$} & {$\ctofar$} & {$\sutranets$} & {$\sutranets$} & {Lag}    \\
                 & {Binning}   & {Binning}   & {Binning}      & {Binning}      & {Period} \\
                 & {Low}       & {High}      & {Low}          & {High}         &        \\ \midrule
$\elec$          & -0.01                      & 1.06                        & -0.06                          & 1.20                            & 24 (one day)                  \\
$\traffic$       & -0.01                      & 1.01                        & -0.02                          & 1.23                            & 24 (one day)                  \\
$\wiki$          & -0.16                      & 2.34                        & -0.79                          & 5.13                            & 7 (one week)                   \\
$\azure$         & -0.05                      & 1.15                        & -0.08                          & 1.20                            & 288 (one day)                 \\
$\mnist$         & 0.00                       & 1.00                        & 0.00                           & 1.00                            & 28 (one row)                  \\
$\mnists$        & 0.00                       & 1.00                        & 0.00                           & 1.00                            & 28 (one row)                  \\ \bottomrule
\end{tabular}
\end{table}

%% file: supplemental_files/equations/pinball_loss_eqn.tex
\begin{equation*}
\Lambda_\alpha(\hat{y}_{i,t}^{(q_\alpha)}, y_{i,t})=(\alpha - \mathcal{I}(y_{i,t} < \hat{y}_{i,t}^{(q_\alpha)}))(y_{i,t}-\hat{y}_{i,t}^{(q_\alpha)})
\end{equation*}

%% file: supplemental_files/equations/QL_eqn.tex
\begin{equation*}
\text{QL$_\alpha$} = \frac{\sum_{i,t} 2\Lambda_{\alpha}(\hat{y}_{i,t}^{(q_\alpha)}, y_{i,t})}{\sum_{i,t}|y_{i,t}|}
\end{equation*}

%% file: supplemental_files/equations/wQL_eqn.tex
\begin{equation*}
\text{wQL} = \frac{1}{9}(\text{QL}_{0.1} + \text{QL}_{0.2} + \dots + \text{QL}_{0.9})
\end{equation*}

%% file: supplemental_files/equations/ND_eqn.tex
\begin{equation*}
\text{ND} = \frac{\sum_{i,t}|y_{i,t} - \hat{y}_{i,t}^{(q_{0.5})}|}{\sum_{i,t}|y_{i,t}|}
\end{equation*}

%% file: supplemental_files/memory_fig.tex
\begin{figure}
  \centering
      {\makebox[\textwidth][c]{
          \begin{subfigure}{\shrinkfigtwo\textwidth}
            \centering
            \scalebox{\resourcesshrink}{
              {\input{supplemental_files/memory_tikz_elec.tex}}
            }
%            \subcaption{$\elec$\label{fig:resources:elec}}
          \end{subfigure}
      \hspace{0mm}
          \begin{subfigure}{\shrinkfigtwo\textwidth}
            \centering
            \scalebox{\resourcesshrink}{
              {\input{supplemental_files/memory_tikz_traffic.tex}}
            }
%            \subcaption{$\traffic$\label{fig:resources:traffic}}
          \end{subfigure}
      }}
      \mbox{}
      \vspace{-1mm}
      \mbox{}
      \caption{Total number of parameters (x-axis) and total \emph{memory
        consumed} during inference (y-axis, lower is better), measured
        via \texttt{nvidia-smi} on NVIDIA Tesla P100, in MiB, for
        $\ctofar$ and $\backfillprevs$ systems of
        Table~\ref{tab:depth} in the main paper, for $\elec$
        (left) and $\traffic$ (right).  Accuracy labeled at each point
        in \boxed{$\mbox{boxes}$}, with best result in
        \textbf{bold}.\label{fig:supp_memory}}
\end{figure}
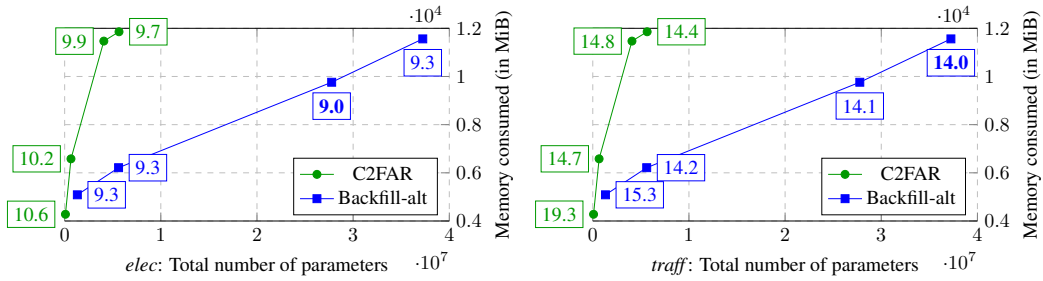

%% file: supplemental_files/memory_tikz_elec.tex
\begin{tikzpicture}
\begin{axis}[
    xlabel={$\elec$: Total number of parameters},
    grid=major,grid style={dashed},
    ylabel={Memory consumed (in MiB)},
    xmin=0, xmax=40000000,
    ymin=4000, ymax=12000,
    xtick={0,10000000,20000000,30000000,40000000},    
    ytick={4000,6000,8000,10000,12000},    
    separate axis lines,
    clip mode=individual,    
    width=8.4cm,    
    height=5cm,
    axis y line*=right,
    y axis line style= { draw opacity=0 },    
    yticklabel pos=right,
    ylabel near ticks,
    legend pos=south east
    ]

\addplot[black!40!green, mark=*] coordinates {
    (81830,4279)
    (658214,6581)
    (4075046,11469)
    (5654054,11857)
};

\addplot[blue, mark=square*] coordinates {
    (1320420,5087)
    (5608164,6213)
    (27768036,9761)
    (37242084,11561)
};

\legend{C2FAR,Backfill-alt}

\newcommand{\fboxgreen}{\fcolorbox{black!40!green}{white}}
\newcommand{\fboxblue}{\fcolorbox{blue}{white}}
\node[circle,draw=black!40!green,inner sep=1pt,label={[black!40!green]left:\fboxgreen{10.6}}] at (axis cs:81830,4279) {};
\node[circle,draw=black!40!green,inner sep=1pt,label={[black!40!green]left:\fboxgreen{10.2}}] at (axis cs:658214,6581) {};
\node[circle,draw=black!40!green,inner sep=1pt,label={[black!40!green]left:\fboxgreen{9.9}}] at (axis cs:4075046,11469) {};
\node[circle,draw=black!40!green,inner sep=1pt,label={[black!40!green]right:\fboxgreen{9.7}}] at (axis cs:5654054,11857) {};

\node[circle,draw=blue,inner sep=1pt,label={[blue]right:\fboxblue{9.3}}] at (axis cs:1320420,5087) {};
\node[circle,draw=blue,inner sep=1pt,label={[blue]right:\fboxblue{9.3}}] at (axis cs:5608164,6213) {};
\node[circle,draw=blue,inner sep=1pt,label={[blue]below:\fboxblue{\textbf{9.0}}}] at (axis cs:27768036,9761) {};
\node[circle,draw=blue,inner sep=1pt,label={[blue]below:\fboxblue{9.3}}] at (axis cs:37242084,11561) {};

\end{axis}
\end{tikzpicture}

%% file: supplemental_files/memory_tikz_traffic.tex
\begin{tikzpicture}
\begin{axis}[
    xlabel={$\traffic$: Total number of parameters},
    grid=major,grid style={dashed},
    ylabel={Memory consumed (in MiB)},
    xmin=0, xmax=40000000,
    ymin=4000, ymax=12000,
    xtick={0,10000000,20000000,30000000,40000000},    
    ytick={4000,6000,8000,10000,12000},    
    separate axis lines,
    clip mode=individual,    
    width=8.4cm,    
    height=5cm,
    axis y line*=right,
    y axis line style= { draw opacity=0 },    
    yticklabel pos=right,
    ylabel near ticks,
    legend pos=south east
    ]

\addplot[black!40!green, mark=*] coordinates {
    (81830,4279)
    (658214,6581)
    (4075046,11469)
    (5654054,11857)
};

\addplot[blue, mark=square*] coordinates {
    (1320420,5087)
    (5608164,6213)
    (27768036,9761)
    (37242084,11561)
};

\legend{C2FAR,Backfill-alt}

\newcommand{\fboxgreen}{\fcolorbox{black!40!green}{white}}
\newcommand{\fboxblue}{\fcolorbox{blue}{white}}
\node[circle,draw=black!40!green,inner sep=1pt,label={[black!40!green]left:\fboxgreen{19.3}}] at (axis cs:81830,4279) {};
\node[circle,draw=black!40!green,inner sep=1pt,label={[black!40!green]left:\fboxgreen{14.7}}] at (axis cs:658214,6581) {};
\node[circle,draw=black!40!green,inner sep=1pt,label={[black!40!green]left:\fboxgreen{14.8}}] at (axis cs:4075046,11469) {};
\node[circle,draw=black!40!green,inner sep=1pt,label={[black!40!green]right:\fboxgreen{14.4}}] at (axis cs:5654054,11857) {};

\node[circle,draw=blue,inner sep=1pt,label={[blue]right:\fboxblue{15.3}}] at (axis cs:1320420,5087) {};
\node[circle,draw=blue,inner sep=1pt,label={[blue]right:\fboxblue{14.2}}] at (axis cs:5608164,6213) {};
\node[circle,draw=blue,inner sep=1pt,label={[blue]below:\fboxblue{14.1}}] at (axis cs:27768036,9761) {};
\node[circle,draw=blue,inner sep=1pt,label={[blue]below:\fboxblue{\textbf{14.0}}}] at (axis cs:37242084,11561) {};

\end{axis}
\end{tikzpicture}

%% file: supplemental_files/trainingtime_fig.tex
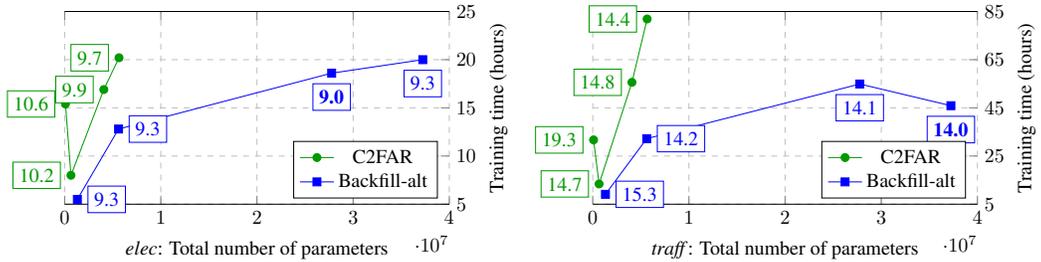
\begin{figure}
  \centering
      {\makebox[\textwidth][c]{
          \begin{subfigure}{\shrinkfigtwo\textwidth}
            \centering
            \scalebox{\resourcesshrink}{
              {\input{supplemental_files/trainingtime_tikz_elec.tex}}
            }
%            \subcaption{$\elec$\label{fig:resources:elec}}
          \end{subfigure}
      \hspace{0mm}
          \begin{subfigure}{\shrinkfigtwo\textwidth}
            \centering
            \scalebox{\resourcesshrink}{
              {\input{supplemental_files/trainingtime_tikz_traffic.tex}}
            }
%            \subcaption{$\traffic$\label{fig:resources:traffic}}
          \end{subfigure}
      }}
      \mbox{}
      \vspace{-1mm}
      \mbox{}
      \caption{Total number of parameters (x-axis) and \emph{training time}
        in hours (y-axis, lower is better) for $\ctofar$ and
        $\backfillprevs$ systems of Table~\ref{tab:depth} in the
        main paper, for $\elec$ (left) and $\traffic$ (right).
        Accuracy labeled at each point in \boxed{$\mbox{boxes}$}, with
        best result in \textbf{bold}.\label{fig:supp_trainingtime}}
\end{figure}

%% file: supplemental_files/trainingtime_tikz_elec.tex
\begin{tikzpicture}
\begin{axis}[
    xlabel={$\elec$: Total number of parameters},
    grid=major,grid style={dashed},
    ylabel={Training time (hours)},
    xmin=0, xmax=40000000,
    ymin=5, ymax=25,    
    xtick={0,10000000,20000000,30000000,40000000},    
    ytick={5,10,15,20,25},    
    separate axis lines,
    clip mode=individual,    
    width=8.4cm,    
    height=5cm,
    axis y line*=right,
    y axis line style= { draw opacity=0 },    
    yticklabel pos=right,
    ylabel near ticks,
    legend pos=south east
    ]

\addplot[black!40!green, mark=*] coordinates {
    (81830,15.4)
    (658214,8.0)
    (4075046,16.9)
    (5654054,20.2)
};

\addplot[blue, mark=square*] coordinates {
    (1320420,5.5)
    (5608164,12.8)
    (27768036,18.6)
    (37242084,20.0)
};

\legend{C2FAR,Backfill-alt}

\newcommand{\fboxgreen}{\fcolorbox{black!40!green}{white}}
\newcommand{\fboxblue}{\fcolorbox{blue}{white}}
\node[circle,draw=black!40!green,inner sep=1pt,label={[black!40!green]left:\fboxgreen{10.6}}] at (axis cs:81830,15.4) {};
\node[circle,draw=black!40!green,inner sep=1pt,label={[black!40!green]left:\fboxgreen{10.2}}] at (axis cs:658214,8.0) {};
\node[circle,draw=black!40!green,inner sep=1pt,label={[black!40!green]left:\fboxgreen{9.9}}] at (axis cs:4075046,16.9) {};
\node[circle,draw=black!40!green,inner sep=1pt,label={[black!40!green]left:\fboxgreen{9.7}}] at (axis cs:5654054,20.2) {};

\node[circle,draw=blue,inner sep=1pt,label={[blue]right:\fboxblue{9.3}}] at (axis cs:1320420,5.5) {};
\node[circle,draw=blue,inner sep=1pt,label={[blue]right:\fboxblue{9.3}}] at (axis cs:5608164,12.8) {};
\node[circle,draw=blue,inner sep=1pt,label={[blue]below:\fboxblue{\textbf{9.0}}}] at (axis cs:27768036,18.6) {};
\node[circle,draw=blue,inner sep=1pt,label={[blue]below:\fboxblue{9.3}}] at (axis cs:37242084,20.0) {};

\end{axis}
\end{tikzpicture}

%% file: supplemental_files/trainingtime_tikz_traffic.tex
\begin{tikzpicture}
\begin{axis}[
    xlabel={$\traffic$: Total number of parameters},
    grid=major,grid style={dashed},
    ylabel={Training time (hours)},
    xmin=0, xmax=40000000,
    ymin=5, ymax=85,
    xtick={0,10000000,20000000,30000000,40000000},    
    ytick={5,25,45,65,85},    
    separate axis lines,
    clip mode=individual,    
    width=8.4cm,    
    height=5cm,
    axis y line*=right,
    y axis line style= { draw opacity=0 },    
    yticklabel pos=right,
    ylabel near ticks,
    legend pos=south east
    ]

\addplot[black!40!green, mark=*] coordinates {
    (81830,31.7)
    (658214,13.4)
    (4075046,55.6)
    (5654054,81.9)
};

\addplot[blue, mark=square*] coordinates {
    (1320420,9.1)
    (5608164,32.2)
    (27768036,54.9)
    (37242084,45.9)
};

\legend{C2FAR,Backfill-alt}

\newcommand{\fboxgreen}{\fcolorbox{black!40!green}{white}}
\newcommand{\fboxblue}{\fcolorbox{blue}{white}}
\node[circle,draw=black!40!green,inner sep=1pt,label={[black!40!green]left:\fboxgreen{19.3}}] at (axis cs:81830,31.7) {};
\node[circle,draw=black!40!green,inner sep=1pt,label={[black!40!green]left:\fboxgreen{14.7}}] at (axis cs:658214,13.4) {};
\node[circle,draw=black!40!green,inner sep=1pt,label={[black!40!green]left:\fboxgreen{14.8}}] at (axis cs:4075046,55.6) {};
\node[circle,draw=black!40!green,inner sep=1pt,label={[black!40!green]left:\fboxgreen{14.4}}] at (axis cs:5654054,81.9) {};

\node[circle,draw=blue,inner sep=1pt,label={[blue]right:\fboxblue{15.3}}] at (axis cs:1320420,9.1) {};
\node[circle,draw=blue,inner sep=1pt,label={[blue]right:\fboxblue{14.2}}] at (axis cs:5608164,32.2) {};
\node[circle,draw=blue,inner sep=1pt,label={[blue]below:\fboxblue{14.1}}] at (axis cs:27768036,54.9) {};
\node[circle,draw=blue,inner sep=1pt,label={[blue]below:\fboxblue{\textbf{14.0}}}] at (axis cs:37242084,45.9) {};

\end{axis}
\end{tikzpicture}

%% file: supplemental_files/nparameters_table.tex
\begin{table}[]
  \caption{Number of parameters for $\ctofar$ and different
    $\sutranet$ variations, as the number of layers and hidden units
    in the LSTMs vary.  Note the number of $\sutranet$ parameters is
    not affected by the sub-series ordering (backfill versus
    regular).\label{tab:nparameters}}
  \scriptsize
  \begin{tabular}{@{}ccccccccc@{}}
    \toprule
    nlayer & nhidden & $\ctofar$     & 6-alt      & 6-non-alt  & 7-alt      & 7-non-alt  & 12-alt     & 12-non-alt \\ \midrule
    1      & 64      & 81,830    & 1,320,420  & 905,700    & 1,734,026  & 1,153,418  & 4,631,496  & 2,806,728  \\
    1      & 128     & 261,926   & 3,230,436  & 2,400,996  & 4,155,914  & 2,994,698  & 10,442,184 & 6,792,648  \\
    1      & 256     & 917,030   & 8,819,940  & 7,161,060  & 11,064,074 & 8,741,642  & 25,602,504 & 18,303,432 \\
    2      & 64      & 181,670   & 1,919,460  & 1,504,740  & 2,432,906  & 1,852,298  & 5,829,576  & 4,004,808  \\
    2      & 128     & 658,214   & 5,608,164  & 4,778,724  & 6,929,930  & 5,768,714  & 15,197,640 & 11,548,104 \\
    2      & 256     & 2,496,038 & 18,293,988 & 16,635,108 & 22,117,130 & 19,794,698 & 44,550,600 & 37,251,528 \\
    3      & 64      & 281,510   & 2,518,500  & 2,103,780  & 3,131,786  & 2,551,178  & 7,027,656  & 5,202,888  \\
    3      & 128     & 1,054,502 & 7,985,892  & 7,156,452  & 9,703,946  & 8,542,730  & 19,953,096 & 16,303,560 \\
    3      & 256     & 4,075,046 & 27,768,036 & 26,109,156 & 33,170,186 & 30,847,754 & 63,498,696 & 56,199,624 \\
    4      & 64      & 381,350   & 3,117,540  & 2,702,820  & 3,830,666  & 3,250,058  & 8,225,736  & 6,400,968  \\
    4      & 128     & 1,450,790 & 10,363,620 & 9,534,180  & 12,477,962 & 11,316,746 & 24,708,552 & 21,059,016 \\
    4      & 256     & 5,654,054 & 37,242,084 & 35,583,204 & 44,223,242 & 41,900,810 & 82,446,792 & 75,147,720 \\ \bottomrule
  \end{tabular}
\end{table}

%% file: supplemental_files/tuning_stability.tex
\begin{figure}
  \centering {\makebox[\textwidth][c]{
      \begin{subfigure}{\shrinkfigtwo\textwidth}
        \input{supplemental_files/tuning_stability_elec.tex}
        \mbox{}
        \vspace{0mm}
        \mbox{}
        \caption{$\elec$}
      \end{subfigure}
      \hspace{0mm}
      \begin{subfigure}{\shrinkfigtwo\textwidth}
        \input{supplemental_files/tuning_stability_traffic.tex}
        \mbox{}
        \vspace{0mm}
        \mbox{}
        \caption{$\traffic$}
      \end{subfigure}
  }}
  \caption{Tuning stability: ND\% for the original \emph{tuning} run
    (first marker) and four subsequent repeats with different random
    seeds, for both $\ctofar$ and $\backfillprevs$, on $\elec$ (left)
    and $\traffic$ (right).  $\sutranets$ have less variation across
    seeds than $\ctofar$.\label{fig:supp_tuning_stability}}
\end{figure}
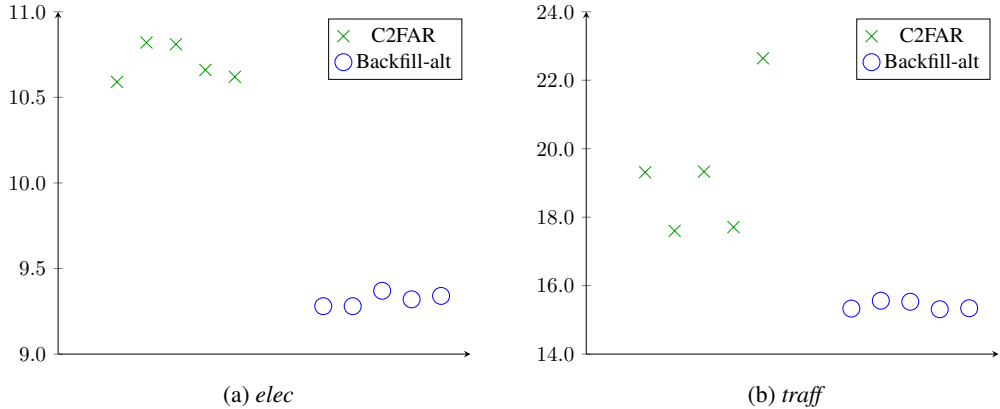

%% file: supplemental_files/tuning_stability_elec.tex
\begin{tikzpicture}[scale=0.8]
  \begin{axis}[
      y tick label style={/pgf/number format/.cd, fixed, fixed zerofill, precision=1},
      axis x line=center,
      axis y line=left,
      xmin=-1,xmax=13,
      ymin=9, ymax=11,
      xticklabels={,,},
      xmajorticks=false,
      scatter/classes={%
        a={mark=x,mark size=4pt,draw=black!40!green},
        b={mark=o,mark size=4pt,draw=blue}}]
    \addplot[scatter,only marks,scatter src=explicit symbolic]%
    table[meta=label]{
      x   y   label
      1 10.59 a
      2 10.82 a
      3 10.81 a
      4 10.66 a
      5 10.62 a
      8 9.28 b
      9 9.28 b
      10 9.37 b
      11 9.32 b
      12 9.34 b
    };
    \legend{C2FAR,Backfill-alt}
  \end{axis}
\end{tikzpicture}

%% file: supplemental_files/tuning_stability_traffic.tex
\begin{tikzpicture}[scale=0.8]
  \begin{axis}[
      y tick label style={/pgf/number format/.cd, fixed, fixed zerofill, precision=1},
      axis x line=center,
      axis y line=left,
      xmin=-1,xmax=13,
      ymin=14, ymax=24,
      xticklabels={,,},
      xmajorticks=false,
      scatter/classes={%
        a={mark=x,mark size=4pt,draw=black!40!green},
        b={mark=o,mark size=4pt,draw=blue}}]
    \addplot[scatter,only marks,scatter src=explicit symbolic]%
    table[meta=label]{
      x   y   label
      1 19.31 a
      2 17.60 a
      3 19.33 a
      4 17.71 a
      5 22.64 a
      8 15.33 b
      9 15.56 b
      10 15.53 b
      11 15.31 b
      12 15.34 b
    };
    \legend{C2FAR,Backfill-alt}
  \end{axis}
\end{tikzpicture}

%% file: supplemental_files/testing_stability.tex
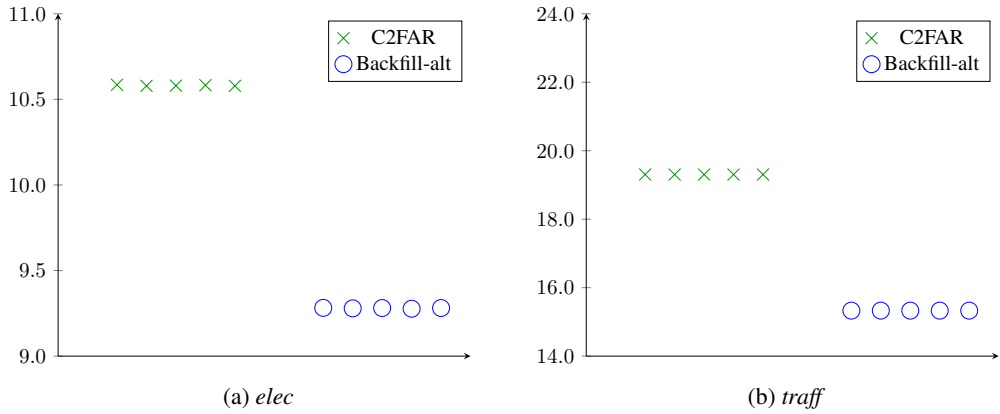
\begin{figure}
  \centering {\makebox[\textwidth][c]{
      \begin{subfigure}{\shrinkfigtwo\textwidth}
        \input{supplemental_files/testing_stability_elec.tex}
        \mbox{}
        \vspace{0mm}
        \mbox{}
        \caption{$\elec$}
      \end{subfigure}
      \hspace{0mm}
      \begin{subfigure}{\shrinkfigtwo\textwidth}
        \input{supplemental_files/testing_stability_traffic.tex}
        \mbox{}
        \vspace{0mm}
        \mbox{}
        \caption{$\traffic$}
      \end{subfigure}
  }}
  \caption{Testing stability: ND\% for the original \emph{evaluation}
    run (first marker) and four subsequent repeats with different
    random seeds, for both $\ctofar$ and $\backfillprevs$, on $\elec$
    (left) and $\traffic$ (right).  The variation across seeds is
    difficult to detect visually in all
    cases.\label{fig:supp_testing_stability}}
\end{figure}

%% file: supplemental_files/testing_stability_elec.tex
\begin{tikzpicture}[scale=0.8]
  \begin{axis}[
      y tick label style={/pgf/number format/.cd, fixed, fixed zerofill, precision=1},
      axis x line=center,
      axis y line=left,
      xmin=-1,xmax=13,
      ymin=9, ymax=11,
      xticklabels={,,},
      xmajorticks=false,
      scatter/classes={%
        a={mark=x,mark size=4pt,draw=black!40!green},
        b={mark=o,mark size=4pt,draw=blue}}]
    \addplot[scatter,only marks,scatter src=explicit symbolic]%
    table[meta=label]{
      x   y   label
      1 10.585 a
      2 10.579 a
      3 10.580 a
      4 10.583 a
      5 10.579 a
      8 9.282 b
      9 9.279 b
      10 9.281 b
      11 9.277 b
      12 9.281 b
    };
    \legend{C2FAR,Backfill-alt}
  \end{axis}
\end{tikzpicture}

%% file: supplemental_files/testing_stability_traffic.tex
\begin{tikzpicture}[scale=0.8]
  \begin{axis}[
      y tick label style={/pgf/number format/.cd, fixed, fixed zerofill, precision=1},
      axis x line=center,
      axis y line=left,
      xmin=-1,xmax=13,
      ymin=14, ymax=24,
      xticklabels={,,},
      xmajorticks=false,
      scatter/classes={%
        a={mark=x,mark size=4pt,draw=black!40!green},
        b={mark=o,mark size=4pt,draw=blue}}]
    \addplot[scatter,only marks,scatter src=explicit symbolic]%
    table[meta=label]{
      x   y   label
      1 19.306 a
      2 19.304 a
      3 19.305 a
      4 19.304 a
      5 19.305 a
      8 15.329 b
      9 15.329 b
      10 15.329 b
      11 15.329 b
      12 15.329 b
    };
    \legend{C2FAR,Backfill-alt}
  \end{axis}
\end{tikzpicture}

%% file: supplemental_files/all_horizons_fig.tex
\begin{figure}
  \centering {\makebox[\textwidth][c]{
      \begin{subfigure}{\shrinkfigthree\textwidth}
        \begin{tikzpicture}
          \node (img1) {\includegraphics[width=\textwidth]{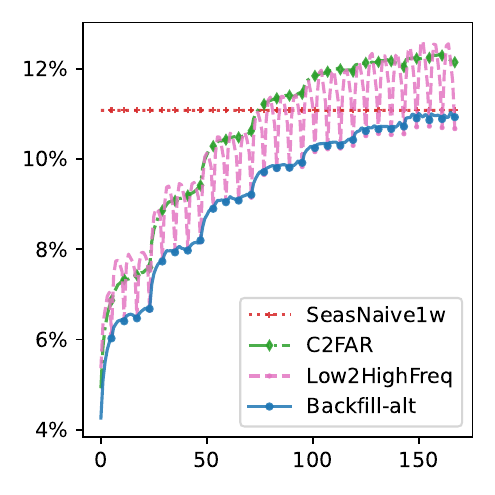}};
        \end{tikzpicture}
        \mbox{}
        \vspace{-8mm}
        \mbox{}
        \caption{$\elec$}
      \end{subfigure}
      \hspace{-3mm}
      \begin{subfigure}{\shrinkfigthree\textwidth}
        \begin{tikzpicture}
          \node (img1) {\includegraphics[width=\textwidth]{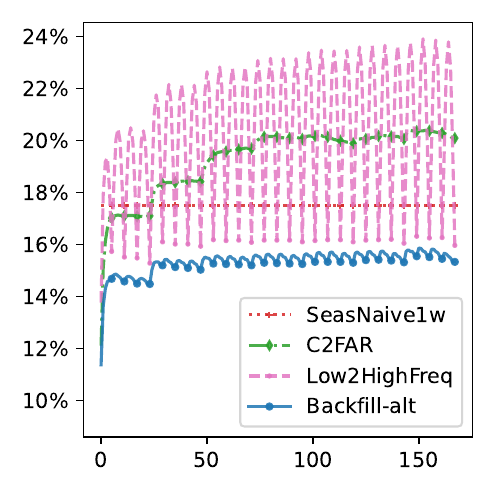}};
        \end{tikzpicture}
        \mbox{}
        \vspace{-8mm}
        \mbox{}
        \caption{$\traffic$}
      \end{subfigure}
      \hspace{-3mm}
      \begin{subfigure}{\shrinkfigthree\textwidth}
        \begin{tikzpicture}
          \node (img1) {\includegraphics[width=\textwidth]{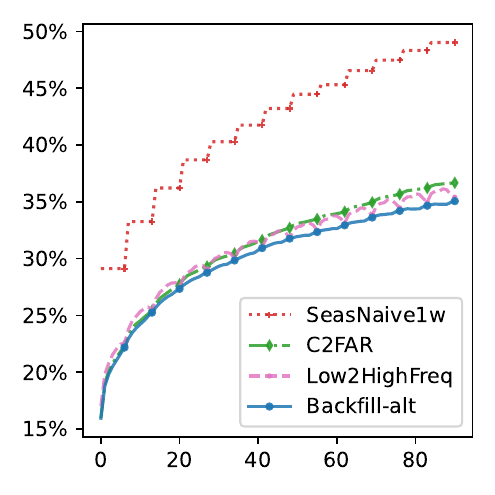}};
        \end{tikzpicture}
        \mbox{}
        \vspace{-8mm}
        \mbox{}
        \caption{$\wiki$}
      \end{subfigure}
  }}
  \mbox{}
  \vspace{-1mm}
  \mbox{}
  \centering {\makebox[\textwidth][c]{
      \begin{subfigure}{\shrinkfigthree\textwidth}
        \begin{tikzpicture}
          \node (img1) {\includegraphics[width=\textwidth]{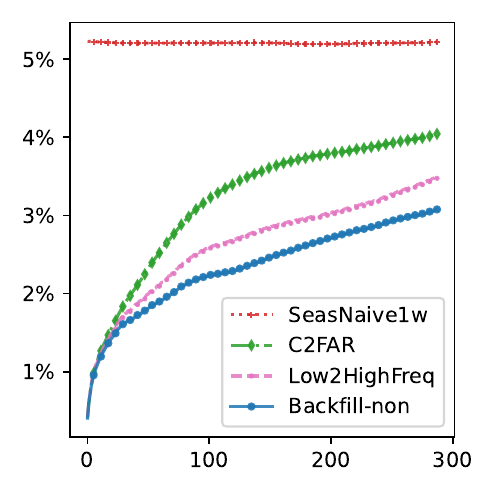}};
        \end{tikzpicture}
        \mbox{}
        \vspace{-8mm}
        \mbox{}
        \caption{$\azure$}
      \end{subfigure}
      \hspace{-3mm}
      \begin{subfigure}{\shrinkfigthree\textwidth}
        \begin{tikzpicture}
          \node (img1) {\includegraphics[width=\textwidth]{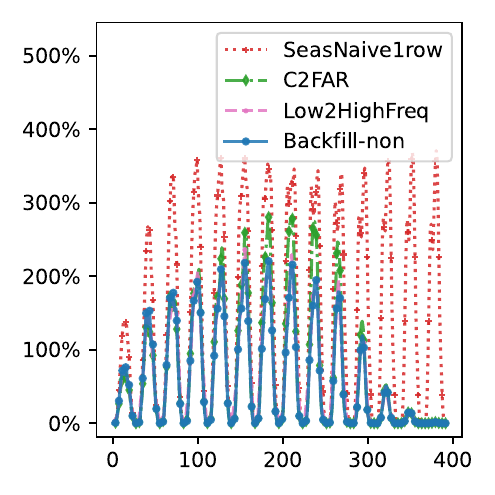}};
        \end{tikzpicture}
        \mbox{}
        \vspace{-8mm}
        \mbox{}
        \caption{$\mnist$}
      \end{subfigure}
      \hspace{-3mm}
      \begin{subfigure}{\shrinkfigthree\textwidth}
        \begin{tikzpicture}
          \node (img1) {\includegraphics[width=\textwidth]{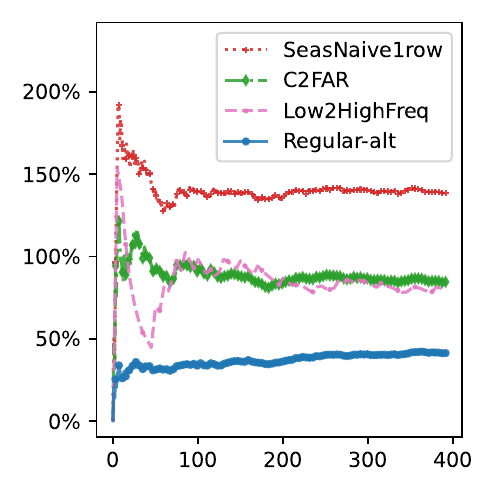}};
        \end{tikzpicture}
        \mbox{}
        \vspace{-8mm}
        \mbox{}
        \caption{$\mnists$ (running average)}
      \end{subfigure}
  }}
  \caption{Normalized deviation (ND\%) at different forecast horizons
    for all the datasets.  Note the very cyclical nature on $\mnist$,
    with a period of 28 --- i.e., one row of the image --- is due to
    the images being more predictable on the left/right edges of each
    row (where they are usually zero), but harder toward the center
    (where they are usually non-zero).  Note also that we use the
    running average in $\mnists$ since the original errors fluctuate
    very wildly by horizon.\label{fig:supp_horizons}}
\end{figure}

%% file: supplemental_files/backfill_standard_rollout.tex
\setul{}{1.6pt}
\begin{figure}
  \centering
      {\makebox[\textwidth][c]{
          \begin{subfigure}{\shrinkfigtwo\textwidth}
            \centering
            \scalebox{\rolloutshrink}{
              {\input{tikz_figures/paper_hf_multiv.uni_backfill.3.9.tex}}
            }
          \end{subfigure}
      }}
      \mbox{}
      \vspace{-1mm}
      \mbox{}
      \caption{Ordering of generative steps in the $\backfillstandard$
        model, following same visual representations as
        Figure~\ref{fig:rollouts} in the main paper. That is, one
        output value (\ul{bold border}) is generated in each row,
        conditional on (1)~feature nodes (\dashuline{dashed border})
        \emph{in that row}, and on (2)~state from previous steps
        ($\Longrightarrow$ arrow connections).  The
        $\backfillstandard$ model does not use SutraNets; instead, it
        flips blocks of $K$ consecutive values into reverse (backfill)
        order, and then processes the values using a standard
        RNN.\label{fig:bs_rollouts}}
\end{figure}
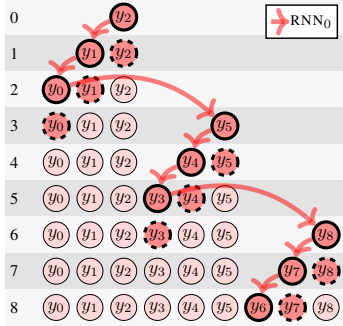

%% file: tikz_figures/paper_hf_multiv.uni_backfill.3.9.tex
\begin{tikzpicture}
\draw[fill=gray!7,draw=none] (-0.9900000000000001cm,-0.35) rectangle ++(6.6,0.7);
\node [algstep,xshift=-0.81cm,yshift=-0.0cm] {\footnotesize 0};
\draw[fill=gray!22,draw=none] (-0.9900000000000001cm,-1.0499999999999998) rectangle ++(6.6,0.7);
\node [algstep,xshift=-0.81cm,yshift=-0.7cm] {\footnotesize 1};
\draw[fill=gray!7,draw=none] (-0.9900000000000001cm,-1.75) rectangle ++(6.6,0.7);
\node [algstep,xshift=-0.81cm,yshift=-1.4cm] {\footnotesize 2};
\draw[fill=gray!22,draw=none] (-0.9900000000000001cm,-2.4499999999999997) rectangle ++(6.6,0.7);
\node [algstep,xshift=-0.81cm,yshift=-2.0999999999999996cm] {\footnotesize 3};
\draw[fill=gray!7,draw=none] (-0.9900000000000001cm,-3.15) rectangle ++(6.6,0.7);
\node [algstep,xshift=-0.81cm,yshift=-2.8cm] {\footnotesize 4};
\draw[fill=gray!22,draw=none] (-0.9900000000000001cm,-3.85) rectangle ++(6.6,0.7);
\node [algstep,xshift=-0.81cm,yshift=-3.5cm] {\footnotesize 5};
\draw[fill=gray!7,draw=none] (-0.9900000000000001cm,-4.549999999999999) rectangle ++(6.6,0.7);
\node [algstep,xshift=-0.81cm,yshift=-4.199999999999999cm] {\footnotesize 6};
\draw[fill=gray!22,draw=none] (-0.9900000000000001cm,-5.249999999999999) rectangle ++(6.6,0.7);
\node [algstep,xshift=-0.81cm,yshift=-4.8999999999999995cm] {\footnotesize 7};
\draw[fill=gray!7,draw=none] (-0.9900000000000001cm,-5.949999999999999) rectangle ++(6.6,0.7);
\node [algstep,xshift=-0.81cm,yshift=-5.6cm] {\footnotesize 8};
\node [hbox] (box0_0) {\phantom{$y$}};
\node [hbox, right of=box0_0, xshift=-3.5mm] (box0_1) {\phantom{$y$}};
\node [tbox, right of=box0_1, xshift=-3.5mm, fill=red!45] (box0_2) {$y_{2}$};
\node [hbox, right of=box0_2, xshift=-3.5mm] (box0_3) {\phantom{$y$}};
\node [hbox, right of=box0_3, xshift=-3.5mm] (box0_4) {\phantom{$y$}};
\node [hbox, right of=box0_4, xshift=-3.5mm] (box0_5) {\phantom{$y$}};
\node [hbox, right of=box0_5, xshift=-3.5mm] (box0_6) {\phantom{$y$}};
\node [hbox, right of=box0_6, xshift=-3.5mm] (box0_7) {\phantom{$y$}};
\node [hbox, right of=box0_7, xshift=-3.5mm] (box0_8) {\phantom{$y$}};
\node [hbox, yshift=-0.7cm] (box1_0) {\phantom{$y$}};
\node [tbox, right of=box1_0, xshift=-3.5mm, fill=red!45] (box1_1) {$y_{1}$};
\node [fbox, right of=box1_1, xshift=-3.5mm, fill=red!45] (box1_2) {$y_{2}$};
\node [hbox, right of=box1_2, xshift=-3.5mm] (box1_3) {\phantom{$y$}};
\node [hbox, right of=box1_3, xshift=-3.5mm] (box1_4) {\phantom{$y$}};
\node [hbox, right of=box1_4, xshift=-3.5mm] (box1_5) {\phantom{$y$}};
\node [hbox, right of=box1_5, xshift=-3.5mm] (box1_6) {\phantom{$y$}};
\node [hbox, right of=box1_6, xshift=-3.5mm] (box1_7) {\phantom{$y$}};
\node [hbox, right of=box1_7, xshift=-3.5mm] (box1_8) {\phantom{$y$}};
\node [tbox, yshift=-1.4cm, fill=red!45] (box2_0) {$y_{0}$};
\node [fbox, right of=box2_0, xshift=-3.5mm, fill=red!45] (box2_1) {$y_{1}$};
\node [rbox, right of=box2_1, xshift=-3.5mm, fill=red!15] (box2_2) {$y_{2}$};
\node [hbox, right of=box2_2, xshift=-3.5mm] (box2_3) {\phantom{$y$}};
\node [hbox, right of=box2_3, xshift=-3.5mm] (box2_4) {\phantom{$y$}};
\node [hbox, right of=box2_4, xshift=-3.5mm] (box2_5) {\phantom{$y$}};
\node [hbox, right of=box2_5, xshift=-3.5mm] (box2_6) {\phantom{$y$}};
\node [hbox, right of=box2_6, xshift=-3.5mm] (box2_7) {\phantom{$y$}};
\node [hbox, right of=box2_7, xshift=-3.5mm] (box2_8) {\phantom{$y$}};
\node [fbox, yshift=-2.0999999999999996cm, fill=red!45] (box3_0) {$y_{0}$};
\node [rbox, right of=box3_0, xshift=-3.5mm, fill=red!15] (box3_1) {$y_{1}$};
\node [rbox, right of=box3_1, xshift=-3.5mm, fill=red!15] (box3_2) {$y_{2}$};
\node [hbox, right of=box3_2, xshift=-3.5mm] (box3_3) {\phantom{$y$}};
\node [hbox, right of=box3_3, xshift=-3.5mm] (box3_4) {\phantom{$y$}};
\node [tbox, right of=box3_4, xshift=-3.5mm, fill=red!45] (box3_5) {$y_{5}$};
\node [hbox, right of=box3_5, xshift=-3.5mm] (box3_6) {\phantom{$y$}};
\node [hbox, right of=box3_6, xshift=-3.5mm] (box3_7) {\phantom{$y$}};
\node [hbox, right of=box3_7, xshift=-3.5mm] (box3_8) {\phantom{$y$}};
\node [rbox, yshift=-2.8cm, fill=red!15] (box4_0) {$y_{0}$};
\node [rbox, right of=box4_0, xshift=-3.5mm, fill=red!15] (box4_1) {$y_{1}$};
\node [rbox, right of=box4_1, xshift=-3.5mm, fill=red!15] (box4_2) {$y_{2}$};
\node [hbox, right of=box4_2, xshift=-3.5mm] (box4_3) {\phantom{$y$}};
\node [tbox, right of=box4_3, xshift=-3.5mm, fill=red!45] (box4_4) {$y_{4}$};
\node [fbox, right of=box4_4, xshift=-3.5mm, fill=red!45] (box4_5) {$y_{5}$};
\node [hbox, right of=box4_5, xshift=-3.5mm] (box4_6) {\phantom{$y$}};
\node [hbox, right of=box4_6, xshift=-3.5mm] (box4_7) {\phantom{$y$}};
\node [hbox, right of=box4_7, xshift=-3.5mm] (box4_8) {\phantom{$y$}};
\node [rbox, yshift=-3.5cm, fill=red!15] (box5_0) {$y_{0}$};
\node [rbox, right of=box5_0, xshift=-3.5mm, fill=red!15] (box5_1) {$y_{1}$};
\node [rbox, right of=box5_1, xshift=-3.5mm, fill=red!15] (box5_2) {$y_{2}$};
\node [tbox, right of=box5_2, xshift=-3.5mm, fill=red!45] (box5_3) {$y_{3}$};
\node [fbox, right of=box5_3, xshift=-3.5mm, fill=red!45] (box5_4) {$y_{4}$};
\node [rbox, right of=box5_4, xshift=-3.5mm, fill=red!15] (box5_5) {$y_{5}$};
\node [hbox, right of=box5_5, xshift=-3.5mm] (box5_6) {\phantom{$y$}};
\node [hbox, right of=box5_6, xshift=-3.5mm] (box5_7) {\phantom{$y$}};
\node [hbox, right of=box5_7, xshift=-3.5mm] (box5_8) {\phantom{$y$}};
\node [rbox, yshift=-4.199999999999999cm, fill=red!15] (box6_0) {$y_{0}$};
\node [rbox, right of=box6_0, xshift=-3.5mm, fill=red!15] (box6_1) {$y_{1}$};
\node [rbox, right of=box6_1, xshift=-3.5mm, fill=red!15] (box6_2) {$y_{2}$};
\node [fbox, right of=box6_2, xshift=-3.5mm, fill=red!45] (box6_3) {$y_{3}$};
\node [rbox, right of=box6_3, xshift=-3.5mm, fill=red!15] (box6_4) {$y_{4}$};
\node [rbox, right of=box6_4, xshift=-3.5mm, fill=red!15] (box6_5) {$y_{5}$};
\node [hbox, right of=box6_5, xshift=-3.5mm] (box6_6) {\phantom{$y$}};
\node [hbox, right of=box6_6, xshift=-3.5mm] (box6_7) {\phantom{$y$}};
\node [tbox, right of=box6_7, xshift=-3.5mm, fill=red!45] (box6_8) {$y_{8}$};
\node [rbox, yshift=-4.8999999999999995cm, fill=red!15] (box7_0) {$y_{0}$};
\node [rbox, right of=box7_0, xshift=-3.5mm, fill=red!15] (box7_1) {$y_{1}$};
\node [rbox, right of=box7_1, xshift=-3.5mm, fill=red!15] (box7_2) {$y_{2}$};
\node [rbox, right of=box7_2, xshift=-3.5mm, fill=red!15] (box7_3) {$y_{3}$};
\node [rbox, right of=box7_3, xshift=-3.5mm, fill=red!15] (box7_4) {$y_{4}$};
\node [rbox, right of=box7_4, xshift=-3.5mm, fill=red!15] (box7_5) {$y_{5}$};
\node [hbox, right of=box7_5, xshift=-3.5mm] (box7_6) {\phantom{$y$}};
\node [tbox, right of=box7_6, xshift=-3.5mm, fill=red!45] (box7_7) {$y_{7}$};
\node [fbox, right of=box7_7, xshift=-3.5mm, fill=red!45] (box7_8) {$y_{8}$};
\node [rbox, yshift=-5.6cm, fill=red!15] (box8_0) {$y_{0}$};
\node [rbox, right of=box8_0, xshift=-3.5mm, fill=red!15] (box8_1) {$y_{1}$};
\node [rbox, right of=box8_1, xshift=-3.5mm, fill=red!15] (box8_2) {$y_{2}$};
\node [rbox, right of=box8_2, xshift=-3.5mm, fill=red!15] (box8_3) {$y_{3}$};
\node [rbox, right of=box8_3, xshift=-3.5mm, fill=red!15] (box8_4) {$y_{4}$};
\node [rbox, right of=box8_4, xshift=-3.5mm, fill=red!15] (box8_5) {$y_{5}$};
\node [tbox, right of=box8_5, xshift=-3.5mm, fill=red!45] (box8_6) {$y_{6}$};
\node [fbox, right of=box8_6, xshift=-3.5mm, fill=red!45] (box8_7) {$y_{7}$};
\node [rbox, right of=box8_7, xshift=-3.5mm, fill=red!15] (box8_8) {$y_{8}$};
\draw[red, bend right,->, line width=0.9mm,opacity=0.5]  (box0_2) to node [auto] {} (box1_1);
\draw[red, bend right,->, line width=0.9mm,opacity=0.5]  (box1_1) to node [auto] {} (box2_0);
\draw[red, bend left,->, line width=0.9mm,opacity=0.5]  (box2_0) to node [auto] {} (box3_5);
\draw[red, bend right,->, line width=0.9mm,opacity=0.5]  (box3_5) to node [auto] {} (box4_4);
\draw[red, bend right,->, line width=0.9mm,opacity=0.5]  (box4_4) to node [auto] {} (box5_3);
\draw[red, bend left,->, line width=0.9mm,opacity=0.5]  (box5_3) to node [auto] {} (box6_8);
\draw[red, bend right,->, line width=0.9mm,opacity=0.5]  (box6_8) to node [auto] {} (box7_7);
\draw[red, bend right,->, line width=0.9mm,opacity=0.5]  (box7_7) to node [auto] {} (box8_6);
\matrix[draw,thick,below left,fill=white,inner sep=1.5pt] at ([xshift=-0.18000000000000002cm,yshift=-3pt]current bounding box.north east) {
  \draw[->,bend left, line width=0.9mm,opacity=0.5,color=red] (0,0) -- ++ (0.35,0); \node[right,xshift=0.3cm]{\textsc{rnn}$_0$};\\
};
\end{tikzpicture}

%% file: supplemental_files/tab_bs_results.tex
\begin{table}[]
  \caption{ND\%, wQL\% across three datasets for different RNNs. The
    $\backfillstandard$ model performs worse than vanilla $\ctofar$
    and worse than all SutraNet variations across all
    datasets.\label{tab:bs_results} \mbox{}
    \vspace{-1mm}
    \mbox{}
  }
  \footnotesize
  \begin{tabular}{@{}cccc@{}}
    \toprule
    & $\elec$ & $\traffic$ & $\mnist$ \\ \midrule
    $\ctofar$                    & 10.6, 8.4                   & 19.3, 16.0                     & 67.9, 52.3                    \\
    $\regularprevs$              & 9.9, 7.9                    & 15.5, 13.0                     & 67.9, 52.4                    \\
    $\regularnoprevs$            & 9.7, 7.8                    & 15.6, 13.1                     & 59.4, 45.4                    \\
    $\backfillprevs$             & 9.3, 7.4                    & 15.3, 12.8                     & 64.4, 49.7                    \\
    $\backfillnoprevs$           & 9.3, 7.4                    & 15.7, 13.1                     & 58.9, 44.9                    \\
    \textbf{$\backfillstandard$} & 11.0, 8.8          & 22.4, 18.1            & 83.2, 67.3           \\
    \bottomrule
\end{tabular}
\end{table}

%% file: supplemental_files/tab_binned_results.tex
\begin{table}[]
  \caption{ND\%, wQL\% across three datasets, \emph{best} results in
    \textbf{bold}.  Applying SutraNets via $\backfillprevs$ improves
    both $\deeparbinned$ and $\ctofar$.  $\ctofar$ also improves over
    $\deeparbinned$, and works synergistically with
    SutraNets.\label{tab:binned_results} \mbox{}
    \vspace{-1mm}
    \mbox{}
  }
  \footnotesize
  \begin{tabular}{@{}cccc@{}}
    \toprule
    & $\elec$ & $\traffic$ & $\mnists$ \\
    \midrule
    $\deeparbinned$                  & 10.6, 8.5               & 20.7, 17.1                     & 82.8, 67.3                    \\
    $\deeparbinned$+$\backfillprevs$ & 9.6, 7.7                & 15.9, 13.3                     & 76.4, 59.1                    \\
    $\ctofar$                    & 10.6, 8.4                   & 19.3, 16.0                     & 67.9, 52.3                    \\
    $\ctofar$+$\backfillprevs$   & \textbf{9.3, 7.4}           & \textbf{15.3, 12.8}            & \textbf{64.4, 49.7}           \\
    \bottomrule
\end{tabular}
\end{table}

%% file: supplemental_files/tab_scaleformer_results.tex
\begin{table}[]
  \caption{Comparison of CRPS for Informer and Scaleformer and wQL (a
    CRPS approximation) for other systems, using results from our main
    Table~\ref{tab:main} and from Table~9 in
    \citet{shabani2022scaleformer}. Since the implemented Scaleformer
    and Informer predictors cannot see values from one week (168
    hours) ago, it makes sense that they perform much worse than both
    SutraNets, and the $\seasonalnaivew$ baseline, on datasets with
    strong weekly seasonality.  Practitioners and researchers should
    be aware of the potential cost of restricting the look-back
    context.\label{tab:scaleformer_results} \mbox{}
    \vspace{-1mm}
    \mbox{}
  }
  \footnotesize
\begin{tabular}{@{}ccccc@{}}
\toprule
System                                      & Length of & Length of & CRPS/wQL & CRPS/wQL \\
                                      & conditioning & prediction & on $\elec$ & on $\traffic$ \\ \midrule
Informer~\cite{zhou2021informer} via~\cite[Table 9]{shabani2022scaleformer}    & 96                     & 96                   & 0.330                   & 0.372               \\
Scaleformer~\cite{shabani2022scaleformer} via~\cite[Table 9]{shabani2022scaleformer} & 96                     & 96                   & 0.238                   & 0.288               \\
$\seasonalnaivew$                        & 168                    & 168                  & 0.111                   & 0.175               \\
$\freqhier$ (Scaleformer for RNNs)         & 168                    & 168                  & 0.082                   & 0.166               \\
SutraNets ($\backfillprevs$)                    & 168                    & 168                  & 0.074                   & 0.128               \\ \bottomrule
\end{tabular}
\end{table}